\documentclass[10pt,twocolumn,letterpaper]{article}

\usepackage{cvpr}              

\makeatletter
\def\blfootnote{\xdef\@thefnmark{}\@footnotetext}
\makeatother

\usepackage[accsupp]{axessibility} 
\usepackage[most,skins]{tcolorbox}
\usepackage{cuted}
\usepackage{afterpage}
\usepackage{lipsum}
\usepackage{capt-of}
\usepackage{multirow}
\usepackage[normalem]{ulem}
\useunder{\uline}{\ul}{}
\usepackage[T1]{fontenc}
\usepackage{nicematrix}
\usepackage{graphicx}
\usepackage{mathtools}

\definecolor{box_section_title_back}{RGB}{166, 195, 226}
\definecolor{box_section_back}{RGB}{225, 235, 246}
\definecolor{box_subsection_title_back}{RGB}{247, 226, 209}
\definecolor{box_subsection_back}{RGB}{250, 240, 232}
\definecolor{answer_right}{RGB}{226, 235, 214}
\definecolor{answer_wrong}{RGB}{255, 207, 205}

\DeclareDocumentCommand{\toolitem}{ m m m }{
\begin{tcolorbox}[
  width=\linewidth,
  colback=box_subsection_back,
  colbacktitle=box_subsection_title_back,
  coltitle=black,
  left=4pt,
  right=4pt,
  top=4pt,
  bottom=4pt,
  title={#3},
  arc=1mm,
  boxrule=0pt,
  halign title=center,
  breakable
  ]
  \begin{center}
    \begin{minipage}{0.29\linewidth}
      \centering
      \includegraphics[height=6em]{#1}
    \end{minipage}
    \hfill
    \begin{minipage}{0.68\linewidth}
      \small{#2}
    \end{minipage}
  \end{center}
\end{tcolorbox}
}

\ExplSyntaxOn
\NewDocumentCommand{\toolsection}{ m +m }
{
  \begin{tcolorbox}[
    width=\linewidth,
    colback=box_section_back,
    colbacktitle=box_section_title_back,
    coltitle=black,
    arc=1mm,
    boxrule=0pt,
    title={
      \parbox{\linewidth}{
        \subsection*{#1}
      }
    },
    fonttitle=\large\bfseries,
    left=4pt,
    right=4pt,
    breakable,
  ]
  #2
  \end{tcolorbox}
}
\ExplSyntaxOff

\definecolor{cvprblue}{rgb}{0.21,0.49,0.74}
\usepackage[pagebackref,breaklinks,colorlinks,allcolors=cvprblue]{hyperref}

\def\paperID{36756} 
\def\confName{CVPR}
\def\confYear{2026}

\title{\textsc{\textbf{ViRC}}: Enhancing Visual Interleaved Mathematical CoT with Reason Chunking}

\author{
\makebox[\textwidth]{
Lihong Wang$^{1,2\dagger}$ \quad
Liangqi Li$^{2}$ \quad
Weiwei Feng$^{2}$ \quad
Jiamin Wu$^{3}$ \quad
Changtao Miao$^{2\ast}$
} \\
\makebox[\textwidth]{
Tieru Wu$^{1,4}$ \quad
Rui Ma$^{1,4\ast}$ \quad
Bo Zhang$^{2}$ \quad
Zhe Li$^{2}$
}\vspace{0.125cm} \\
\makebox[\textwidth]{
$^1$Jilin University \quad
$^2$Ant Digital Technologies, Ant Group \quad
$^3$The Chinese University of Hong Kong
} \\
\makebox[\textwidth]{
$^4$Engineering Research Center of Knowledge-Driven Human-Machine Intelligence, MOE, China
}
\vspace{-1em}
}

\makeatletter
\renewcommand{\maketitlesupplementary}{%
  \newpage
  \twocolumn[%
    \centering
    \Large\textbf{\thetitle}\par
    \vspace{0.5em}
    \Large Supplementary Material\par
    \vspace{1.0em}

    {\large
      \lineskip .5em
      \begin{tabular}[t]{c}
        Lihong Wang$^{1,2\dagger}$ \quad
        Liangqi Li$^{2}$ \quad
        Weiwei Feng$^{2}$ \quad
        Jiamin Wu$^{3}$ \quad
        Changtao Miao$^{2\ast}$\\
        Tieru Wu$^{1,4}$ \quad
        Rui Ma$^{1,4\ast}$ \quad
        Bo Zhang$^{2}$ \quad
        Zhe Li$^{2}$\\[0.125cm]
        $^1$Jilin University \quad
        $^2$Ant Digital Technologies, Ant Group \quad
        $^3$The Chinese University of Hong Kong\\
        $^4$Engineering Research Center of Knowledge-Driven Human-Machine Intelligence, MOE, China
      \end{tabular}\par
    }

    \vskip .5em
    \vspace*{12pt}
  ]%
}
\makeatother

\begin{document}

\maketitle
\begin{abstract}
CoT has significantly enhanced the reasoning ability of LLMs while it faces challenges when extended to multimodal domains, particularly in mathematical tasks.
Existing MLLMs typically perform textual reasoning solely from a single static mathematical image, overlooking dynamic visual acquisition during reasoning. 
In contrast, humans repeatedly examine visual image and employ step-by-step reasoning to prove intermediate propositions. 
This strategy of decomposing the problem-solving process into key logical nodes adheres to Miller's Law in cognitive science.
Inspired by this insight, we propose a \textsc{\textbf{ViRC}} framework for multimodal mathematical tasks, introducing a \textit{Reason Chunking} mechanism that structures multimodal mathematical CoT into consecutive \textbf{Critical Reasoning Units (CRUs)} to simulate human expert problem-solving patterns.
CRUs ensure intra-unit textual coherence for intermediate proposition verification while integrating visual information across units to generate subsequent propositions and support structured reasoning. 
To this end, we present \textbf{CRUX} dataset by using three visual tools and four reasoning patterns to provide explicitly annotated CRUs across multiple reasoning paths for each mathematical problem.
Leveraging the CRUX dataset, we propose a progressive training strategy inspired by human cognitive learning, which includes Instructional SFT, Practice SFT, and Strategic RL, aimed at further strengthening the Reason Chunking ability of the model.
The resulting \textsc{\textbf{ViRC}}\textbf{-7B} model achieves a 18.8\% average improvement over baselines across multiple mathematical benchmarks.
Code is available at \href{https://github.com/Leon-LihongWang/ViRC}{https://github.com/Leon-LihongWang/ViRC}.
\end{abstract}

\blfootnote{$^{\dagger}$Work done during Lihong Wang's internship at Ant Digital Technologies, Ant Group.}
\blfootnote{$^{\ast}$Corresponding authors.}

\vspace{-1em}
\begin{figure}
  \centering
  \includegraphics[width=0.99\linewidth]{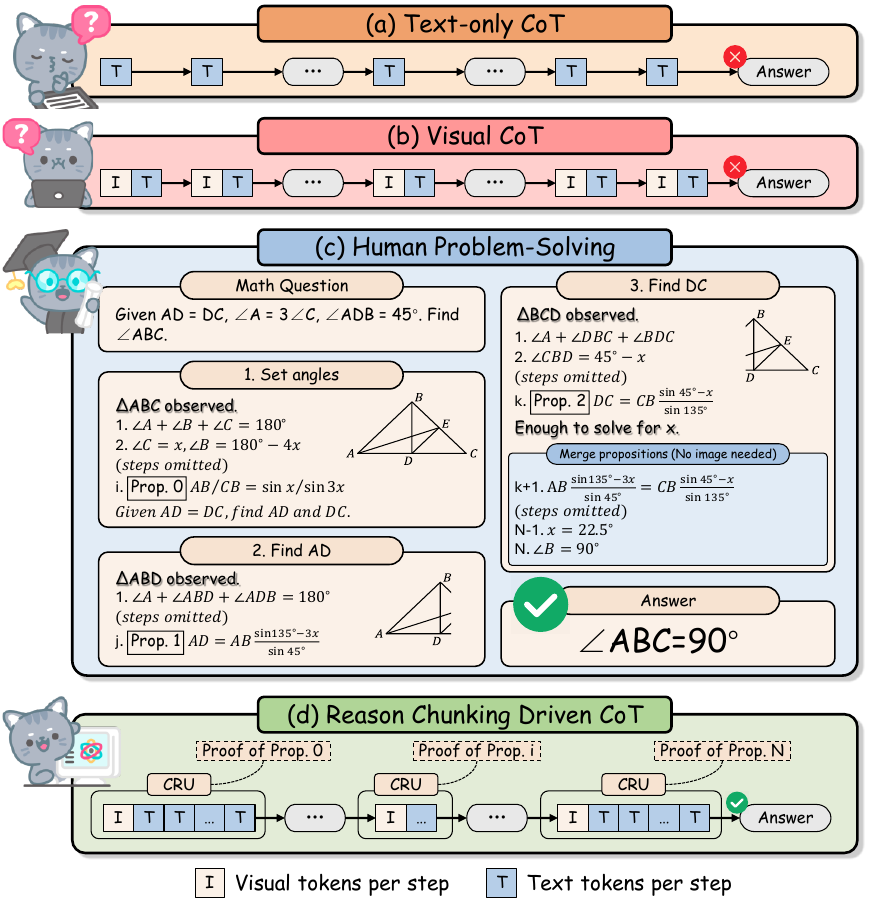}
    \caption{
Comparison of reasoning paradigms. 
(a) \textbf{Text-only CoT} lacks visual grounding. 
(b) \textbf{Visual CoT} injects visual signals at every step, causing redundancy. 
(c) \textbf{Human problem-solving} selectively attends to visual cues and organizes reasoning into chunks. 
(d) \textbf{Reason Chunking Driven CoT} follows this selective, structured strategy to produce image-grounded reasoning.
}
  \label{fig:teaser}
\end{figure}

\section{Introduction}

Large language models (LLMs)~\cite{llama3,qwen3,gpt4,deepseekr1,deepseekv3} have significantly improved their reasoning capabilities by introducing the Chain-of-Thought (CoT)~\cite{kojima2022zeroshotreasoners,wei2022cotprompting} mechanism, which is inspired by the deliberate reasoning process of humans.
In recent works, it is believed that extending CoT into multimodal large language models (MLLMs)~\cite{claude35,gpt4v,openai2024gpt4ocard,wu2024deepseekvl2,qwen25VL,qwen3vl} for visual–language tasks is worth studying, such as image recognition~\cite{bao2025mllmi2w,heng2025rodmllm,liang2025improvingmllmocr} and video understanding~\cite{min2025morevqa,wu2024zeroshotlongformvideounderstanding,kalarani2024unveilinginvisiblecaptioningvideos}.
However, for more complex reasoning in math problems~\cite{chen2021geoqa,lu2024mathvista,chen2024mmstar}, directly applying CoT to the reasoning process with visual context remains challenging.
%

Existing MLLMs~\cite{li2024llavaov,chen2025intervl25,wu2024deepseekvl2,qwen25VL} typically generate text-only reasoning steps from a single static image when handling multimodal mathematical problems, as shown in~\cref{fig:teaser}~(a). 
Due to their limited visual perception capabilities, these models struggle to dynamically capture fine-grained information from mathematical images during the reasoning process, often leading to errors.
Recently, several approaches~\cite{zheng2025deepeyes,su2025pixelreasoner,zhang2025cof} propose a novel paradigm called \textit{“Thinking with Images”}, mimicking natural human visual cognitive processes by invoking visual tools to naturally interleave visual content with textual CoT, as shown in \cref{fig:teaser}~(b). 
However, these methods~\cite{lai2025minio3,chen2025mint} indiscriminately insert visual tokens at every reasoning step, potentially introducing redundant or irrelevant information. 
Meanwhile, existing visual CoT frameworks~\cite{su2025openthinkimg,wu2025reinforcingspatialreasoning,hu2024visualsketchpad,liu2025visualagenticreinforcementfinetuning} remain constrained to a single reasoning path, which deviates from the inherent problem-solving patterns of mathematical propositions.
Specifically, they lack an explicit mechanism for hierarchical decomposition of the reasoning process,  a strategy commonly employed by humans.


To this end, we explore the fundamental mechanism of how humans solve multimodal mathematical problems.
As shown in~\cref{fig:teaser}~(c), humans solve the geometry problem through a hierarchical step-by-step strategy involving three image-related proof steps (i.e., “Set angles”, “Find AD”, and “Find DC”) and one intermediate conclusion integration step (i.e., “Merge propositions”), deriving the solution $\angle ABC = 90^\circ$. 
This strategy of decomposing complex problems into steps not only aligns with the “scoring points” design in mathematical assessment systems, but also conforms to Miller’s Law\footnote{According to Miller’s Law (Miller, 1956), the capacity of human short-term memory is limited to approximately seven (±2) chunks of information, where a chunk refers to a meaningful unit formed by grouping individual items based on prior knowledge or perceptual patterns.
More details in the supplementary material
\cref{supp:millerslaw}.
} 
in cognitive science. 
Miller’s Law~\cite{miller1956miller} proposes the concept of “information chunking”, which integrates multiple related information elements into a single cognitive unit to enhance human cognitive capabilities.
In comparison, existing MLLMs~\cite{jiang2025vlmr3,chen2025mint,zhong2025omnir1,su2025pixelreasoner} predominantly adopt single, direct reasoning pathways, neither adopting the human step-by-step problem-solving strategy nor following the principles of Miller’s Law.


Inspired by the aforementioned discussion, we propose a \textsc{\textbf{ViRC}} framework for multimodal mathematical tasks, introducing a \textbf{Reason Chunking} mechanism that mimics human mathematical problem-solving processes to enhance reasoning capability. 
Specifically, Reason Chunking is grounded in Miller’s information chunking theory, restructuring the visual reasoning chain into a series of \textbf{Critical Reasoning Units (CRUs)} to establish clear logical hierarchies.
As shown in \cref{fig:teaser}~(d), each CRU encapsulates a self-contained intermediate  proposition, directly conforming to the chunk-based cognitive principles of Miller’s Law. 
Within a CRU, the model maintains textual logical coherence to validate the intermediate proposition, enabling hierarchical decomposition of complex problems. 
Across CRUs, visual information is dynamically injected to support subsequent propositions and realize adaptive visual attention.
To ensure the \textsc{\textbf{ViRC}} model follows the Reason Chunking mechanism during training, we construct the \textbf{CRUX} dataset, a multimodal mathematical reasoning dataset containing 100K samples.
Each sample consists of one correct reasoning path and two plausible incorrect paths, covering four human-aligned cognitive modes: planning, verifying, backtracking, and reflecting.
We employ a three-stage annotation pipeline for reasoning path generation of mathematical problems: (1) sampling diverse reasoning paths, (2) mapping steps to CRUs, and (3) grounding CRUs. 
This process generates a visually interleaved reasoning dataset containing multiple CRUs, providing token-level supervision for training models to mimic human hierarchical problem-solving processes, as shown in \cref{fig:teaser}~(c-d).
To simulate human experts' selective review behavior, each reasoning path CRU incorporates three visual tools: \texttt{crop} (region selection), \texttt{scale} (resolution adjustment), and \texttt{display} (visual context recall).

Furthermore, we design a three-stage training strategy based on the CRUX dataset: (1) Instructional SFT, (2) Practice SFT, and (3) Strategic RL. 
This progressive training framework is inspired by the developmental pattern of human problem-solving skills: first acquiring declarative knowledge and foundational solving habits, then enhancing these abilities through practice in realistic scenarios, and finally achieving mastery by concentrating on challenging problems and error correction.
Following this training strategy, we train the \textsc{\textbf{ViRC}}\textbf{-7B} model on the mathematical CRUX dataset.
In multimodal mathematical reasoning benchmarks (MathVista~\cite{lu2024mathvista}, MMStar~\cite{chen2024mmstar}, and GeoQA~\cite{chen2021geoqa}), our model achieves an average performance gain of 18.8\% over the baselines.
In cross-domain high-resolution image benchmarks (VisualProbe~\cite{lai2025minio3}, V*~\cite{wu2024vstar}, and HR-Bench~\cite{wang2025hrbench}), it exceeds baselines by an average of 9\%, demonstrating strong generalization capability.

The main contributions of this work are as follows:
\begin{itemize}
    \item We propose the \textsc{\textbf{ViRC}} framework, incorporating \textit{\textbf{Reason Chunking}} to emulate human experts’ problem-solving patterns and enhance multimodal mathematical reasoning through structured CRUs.
    \item We construct the \textbf{CRUX} dataset through developing a three-stage pipeline to generate the first visually interleaved reasoning dataset with multiple CRUs.
    \item We design a progressive training strategy aligned with human cognitive learning processes to boost reasoning capabilities of \textsc{\textbf{ViRC}}, validated by extensive experiments.
\end{itemize}

\begin{figure*}[t]
  \centering
  \includegraphics[width=\linewidth]{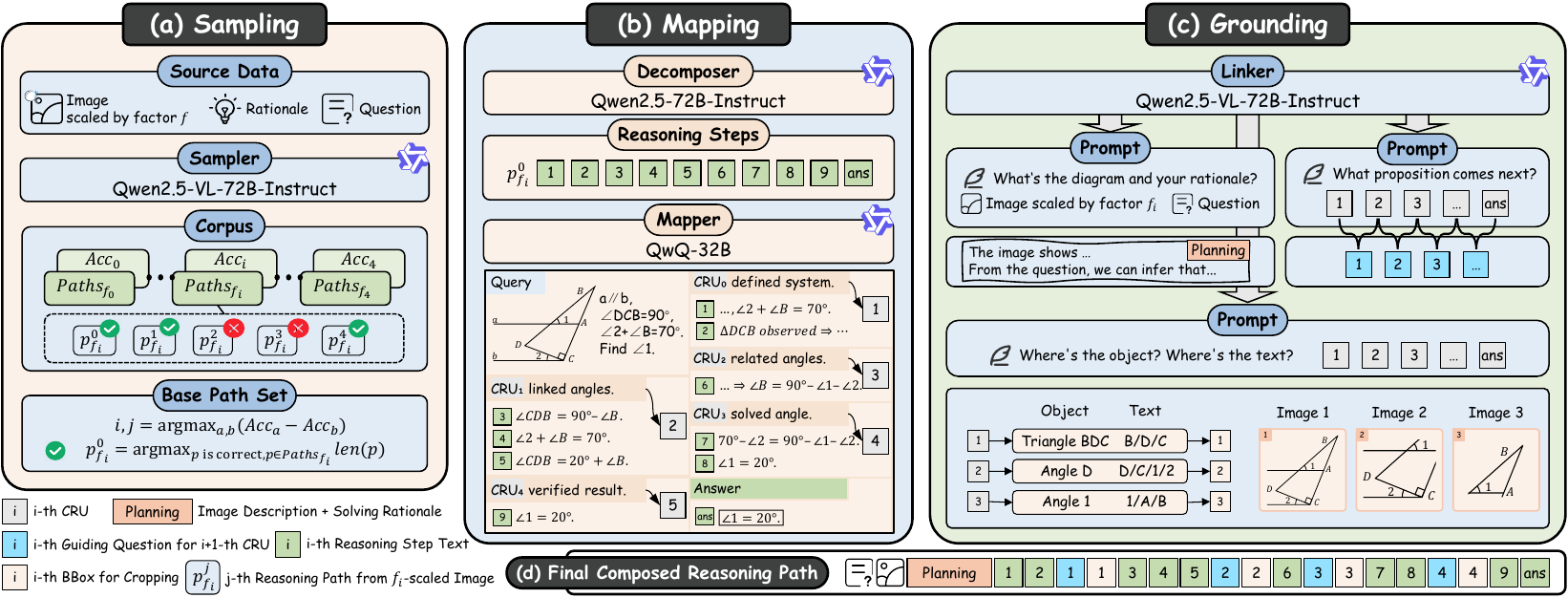}
  \caption{
\textbf{CRUX data curation pipeline.} 
(a) Sampling: generating diverse reasoning paths. 
(b) Mapping: grouping fine-grained steps into CRUs. 
(c) Grounding: assigning each CRU its supporting visual region and auxiliary texts. 
(d) Final Composed Reasoning Path: assembling all components into a coherent CRU-aligned reasoning sequence.
}
  \label{fig:data_pipeline}
\end{figure*}
\section{Related Work}

\noindent\textbf{Multimodal Mathematical Reasoning.}
Recent Multimodal Large Language Models (MLLMs), such as GPT-4V~\cite{gpt4v} and Qwen2.5-VL~\cite{qwen25VL}, exhibit strong general vision-language capabilities but still underperform on visual mathematical reasoning benchmarks~\cite{chen2021geoqa,chen2024mmstar,lu2024mathvista}.
To address this, existing efforts fall into two categories.
The first direction focuses on improving model training. For instance, Hint-GRPO~\cite{huang2025hintgrpo} enhances data efficiency in reinforcement learning to reduce textual bias, while R1-VL~\cite{zhang2025r1vl} employs fine-grained reward signals to strengthen step-wise reasoning.
The second direction emphasizes high-quality data curation. Specifically, Mulberry~\cite{yao2024mulberry} and MM-Eureka~\cite{meng2025mmeureka} provide diverse mathematical problems with detailed solutions that mimic human problem-solving processes.
Despite their progress, these approaches share a fundamental limitation in that they assume a single, static encoding of the input image suffices for the entire reasoning process.
This design ignores the iterative nature of human visual inspection.
Consequently, models often misinterpret complex diagrams or lose track of intermediate goals.

\noindent\textbf{Visual Chain-of-Thought.}
To overcome the limitations of static visual perception, recent works propose Visual Chain-of-Thought (VCoT), which interleaves visual information throughout textual reasoning.
It is realized via agentic tool use~\cite{zheng2025deepeyes,su2025pixelreasoner,lai2025minio3,wang2025simpleo3}, code-executed visual operations~\cite{su2025openthinkimg,shen2025zoomeye}, or imagined visual feedback~\cite{xu2025visualplanning,chern2025thinkingwithgeneratedimages}.
Although implemented differently, these approaches all aim to inject denser and more diverse visual signals to enhance fine-grained image understanding during reasoning.
Within the agentic tool paradigm, methods such as DeepEyes~\cite{zheng2025deepeyes} use reinforcement learning to invoke zoom-in tools for obtaining informative subregions, while Mini-o3~\cite{lai2025minio3} extends this idea by enabling longer tool-use trajectories to access denser visual cues.
In contrast, approaches like OpenThinkImg~\cite{su2025openthinkimg} and Visual Planning~\cite{xu2025visualplanning} enhance perception through broader image manipulation, including fine-grained segmentation or generated visual sequences.
Despite these advances, most VCoT methods still interleave visual tokens at every reasoning step, regardless of the actual necessity of additional visual evidence.
Such indiscriminate integration increases computational overhead and injects redundant or weakly relevant visual signals, preventing the model from behaving in a human-like manner, where visual information is consulted only when necessary.
\begin{figure*}[t]
  \centering
  \includegraphics[width=\linewidth]{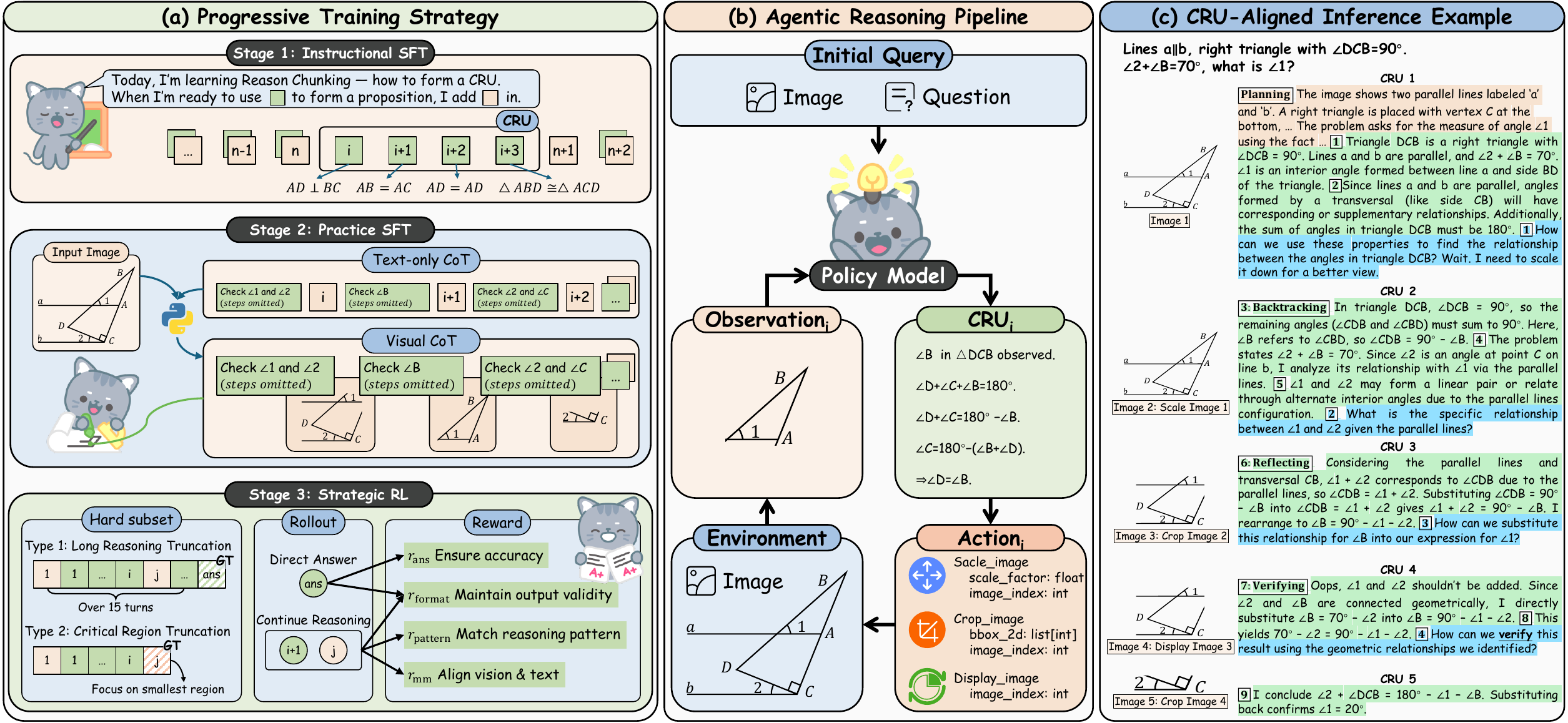}
\caption{
Overview of the \textsc{\textbf{ViRC}} framework. 
(a) Progressive training strategy: Instructional SFT learns CRU-level structure from text-only CRUX, Practice SFT trains grounded multimodal reasoning with executed tool calls, and Strategic RL refines reasoning and tool selection on a curated hard subset derived from CRUX. 
(b) Agentic reasoning pipeline: the model alternates between tool invocation and CRU-level reasoning, using the observation returned by each tool call as the visual context for generating the next proposition. 
(c) CRU-aligned inference example from \textsc{\textbf{ViRC}}-7B: a test-time trajectory that visualizes, for each CRU, the consulted image view, tool calls, and textual reasoning, where multiple textual steps may belong to one CRU and different CRUs are tagged with planning, verifying, backtracking, or reflecting.
}

  \label{fig:train_stage}
\end{figure*}

\section{\textsc{\textbf{ViRC}}}

To address the challenges of multimodal CoT in mathematical reasoning, we propose the \textsc{\textbf{ViRC}} framework. 
In this section, we first introduce basic notations 
in \cref{sec:notations}. 
In \cref{sec:reason_chunking}, we illustrate the Reason Chunking mechanism in \textsc{\textbf{ViRC}}. 
Then, we present the CRUX dataset and describe its construction pipeline in \cref{sec:dataset_curation}. 
Finally, we detail our three-stage progressive training strategy in \cref{sec:training_strategy}.

\subsection{Notations and Preliminaries} \label{sec:notations}
Given an input image $I$ and a mathematical question $T$, an MLLM generates a response $\mathcal{C}$ comprising reasoning steps and a final answer $a$. Let ${V} = \text{VisionEncoder}(I)$ denote the sequence of visual tokens derived from the image, and ${X} = \text{TextEncoder}(T)$ the embedded representation of the textual prompt. The generation process is formalized as:
\begin{equation}
    \mathcal{C} = \text{MLLM}({V}, {X}).
    \label{eq:general_cot}
\end{equation}

Current reasoning approaches instantiate $\mathcal{C}$ in two predominant forms, \textit{Chain-of-Thought} (CoT) and \textit{Visual Chain-of-Thought} (VCoT).
CoT fuses the global visual context only at the beginning and subsequently produces a purely textual reasoning sequence:
\begin{equation}
    \mathcal{C}_{\text{CoT}} = \big[ {s}^{(1)},\, {s}^{(2)},\, \dots,\, {s}^{(K)},\, {a} \big],
    \label{eq:text_cot}
\end{equation}
where ${s}^{(k)}$ denotes the $k$-th reasoning step and $K$ is the total number of reasoning steps.
This approach lacks dynamic access to visual details, often missing critical cues during inference.
To compensate for this deficiency, VCoT is proposed to interleave visual tokens ${v}^{(i)}$ before every reasoning step, yielding a dense alternating structure:
\begin{equation}
  \resizebox{0.9\columnwidth}{!}{$
  \mathcal{C}_{\text{VCoT}} = \big[ ( {v}^{(1)},\, {s}^{(1)}) ,\, ({v}^{(2)},\, {s}^{(2)}) ,\, \dots,\, ({v}^{(K)},\, {s}^{(K)}),\, {a} \big].
  $}
  \label{eq:v_cot}
\end{equation}

\subsection{Reason Chunking}
\label{sec:reason_chunking}
Although VCoT enhances step-wise visual grounding, it indiscriminately injects visual signals regardless of necessity, introducing redundancy and violating the principle of selective attention observed in human cognition.
To address these limitations, we propose \textbf{Reason Chunking}, a structured reasoning framework that decomposes multimodal CoT into a sequence of \textbf{Critical Reasoning Units (CRUs)}, where each CRU represents a self-contained proposition validated through on-demand visual grounding.
Formally, the fundamental unit, the CRU, can be defined as:
\begin{equation}
    \text{CRU}^{(i)} \coloneqq \big( {v}^{(i)},\ \{{s}^{(i,1)},\, {s}^{(i,2)},\, \dots,\, {s}^{(i,m_i)}\} \big).
    \label{eq:cru}
\end{equation}
Here, ${v}^{(i)}$ is a sequence of visual tokens dynamically introduced at the $i$-th reasoning stage.
The set $\{{s}^{(i,\ell)}\}_{\ell=1}^{m_i}$ consists of $m_i \geq 1$ coherent textual steps that jointly verify a semantically complete proposition.
The full reasoning chain is constructed by sequentially concatenating $N$ such CRUs, which we compactly express as:
\begin{equation}
    \mathcal{C}_{\text{Ours}} = \big[ \text{CRU}^{(1)},\ \text{CRU}^{(2)},\ \dots,\ \text{CRU}^{(N)},\ {a} \big].
    \label{eq:reason_chunking}
\end{equation}
The total number of reasoning steps is $K = \sum_{i=1}^N m_i$, with $1 < N < K$.
This mechanism enables on-demand visual invocation and hierarchical problem decomposition, thereby improving both accuracy and cognitive efficiency in complex multimodal mathematical tasks.

Notably, although existing approaches do not explicitly structure reasoning into semantic units, their behavior can be naturally interpreted within the Reason Chunking framework.
CoT corresponds to a degenerate case with a single CRU:
\begin{equation}
    \mathcal{C}_{\text{CoT}} = \big[ \text{CRU}^{(1)},\ {a} \big],
    \label{eq:text_as_cru}
\end{equation}
where the CRU utilizes the full visual context ${V}$.
In contrast, visual CoT represents an extreme instantiation in which each reasoning step forms its own CRU:
\begin{equation}
    \mathcal{C}_{\text{VCoT}} = \big[ \text{CRU}^{(1)}_{m_1=1}, \text{CRU}^{(2)}_{m_2=1}, \dots, \text{CRU}^{(K)}_{m_K=1}
,\ {a} \big],
    \label{eq:vcot_as_cru}
\end{equation}
where each $m_k$ = 1, meaning each $\text{CRU}^{(k)}$ contains exactly one textual step paired with its preceding visual tokens.

\subsection{Dataset Curation}
\label{sec:dataset_curation}
To support structured multimodal reasoning grounded in Critical Reasoning Units (CRUs), we construct the \textbf{CRUX} dataset, which contains 100K multimodal mathematical reasoning examples.
We start from the same set of 54K problems as MINT-CoT~\cite{chen2025mint}, but instead of reusing their annotations, we regenerate all reasoning paths.
As illustrated in \cref{fig:data_pipeline}, the CRUX dataset is created through a three-stage annotation pipeline that expands each original problem into one or more CRU-aligned reasoning paths, resulting in 100K annotated instances.

\noindent\textbf{Sampling Diverse Reasoning Paths.}
As shown in \cref{fig:data_pipeline} (a), we generate multiple reasoning paths by prompting a vision–language model across different image scales.
We then identify one high-accuracy and one low-accuracy scale to obtain both correct and incorrect reasoning paths.
From these, we extract the longest correct and incorrect paths to form the base path set for subsequent processing.
Since the following stages treat correct and incorrect paths in an identical manner, we present the pipeline using only the correct path for clarity.
Details on incorporating incorrect paths are provided in the supplementary material
\cref{supp:dataset_curation}. 

\noindent\textbf{Mapping Steps to CRUs.}
As illustrated in \cref{fig:data_pipeline}~(b), we decompose the reasoning path into fine-grained steps and map them to their corresponding intermediate propositions.
Steps associated with the same proposition are grouped into a semantically coherent CRU.

\noindent\textbf{Grounding CRUs.}
\cref{fig:data_pipeline}~(c) shows how each CRU is grounded with its supporting visual evidence.
We detect the focal object and the corresponding reference text, and take the union of their bounding boxes as the CRU’s associated image region.
We also generate an image description, a brief solution rationale, and a guiding question between consecutive CRUs to ensure coherent proposition-level transitions.
Finally, all generated components are assembled through a rule-based procedure to form the final composed reasoning path, as illustrated in \cref{fig:data_pipeline}~(d).
\noindent\textbf{Cognitive Reasoning Patterns.}
We introduce four cognitive reasoning patterns, namely Planning, Reflecting, Verifying, and Backtracking, to provide a human-like problem-solving structure to each path.
Inspired by analyses of cognitive behaviors in text-only CoT reasoning~\cite{gandhi2025stars}, we adapt these patterns to the multimodal reasoning setting by operationalizing them as explicit tool call transformations aligned with the CRU framework. Concretely, we first convert each CRU’s bounding box into a canonical \texttt{crop} call and then instantiate each pattern by editing this initial tool invocation.
\textbf{Planning} provides global context by summarizing the image content and outlining the overall strategy before CRU-level reasoning begins.
\textbf{Reflecting} enables iterative visual focusing by reusing and adjusting earlier \texttt{crop} calls when a new CRU operates on a subregion of a previous view.
\textbf{Verifying} inserts explicit \texttt{display} calls that prompt the model to re-examine relevant visual evidence before advancing to the next proposition.
\textbf{Backtracking} corrects unreliable interleaved visual information by inserting \texttt{scale} calls to rescale the view when necessary.
Together, these patterns structure multimodal reasoning at a higher level than individual steps.
Implementation details are provided in the supplementary material
\cref{supp:modeling_cognitive_reasoning_patterns}.

\subsection{Training strategy}
\label{sec:training_strategy}
As illustrated in \cref{fig:train_stage}, we design a progressive training curriculum that mirrors human learning and gradually unifies structural reasoning (\textbf{Instructional SFT}), perceptual grounding (\textbf{Practice SFT}), and strategic optimization (\textbf{Strategic RL}).
This curriculum is designed to support the agentic reasoning pipeline of \textsc{\textbf{ViRC}} shown in \cref{fig:train_stage}~(b), where the model alternates between CRU-level reasoning and tool invocation until reaching the final answer.



\noindent\textbf{Stage 1: Instructional SFT.}  
This stage aims to internalize the structured reasoning template of Reason Chunking without visual distraction.
As shown in \cref{fig:train_stage}~(a) Stage~1, we treat a 50K subset of the CRUX dataset as a pure text dataset to perform SFT on the \textsc{\textbf{ViRC}} model, since all visual signals $v^{(i)}$ returned by tool execution are masked out.
The model observes the full reasoning path, including the problem statement, the structured format of tool calls, and the internal textual steps of each CRU.


\noindent\textbf{Stage 2: Practice SFT.}  
After acquiring a structural prior over CRUs, we perform SFT using the same 50K subset, but in its full multimodal form, as shown in \cref{fig:train_stage}~(a) Stage~2. When the model generates a tool call $\tau^{(i)}$, we execute it and feed the resulting visual signal $v^{(i)}$ back as contextual input. The model then uses this grounded evidence to complete the current CRU.

\noindent{\textbf{Stage 3: Strategic RL.}}
We introduce a coordinated RL framework that builds on both a carefully curated hard subset derived from the same 50K subset and a tailored reward design, as illustrated in \cref{fig:train_stage}~(a) Stage~3.
The detailed construction rules for this hard subset are provided in the supplementary material 
\cref{supp:hard_subset_curation}.

\begin{table*}[tp]
  \caption{
  \textbf{Comparison with SOTA Models.} Our \textsc{\textbf{ViRC}} is trained on the CRUX dataset and tested on GeoQA~\cite{chen2021geoqa}, MMStar-Math~\cite{chen2024mmstar}, and MathVista-Math~\cite{lu2024mathvista}. Avg. reports the mean of the results on the three benchmarks. \textbf{Bold} and {\ul underlined} results indicate the best and second-best, respectively. Results marked with $\dagger$ are from \cite{chen2025mint}, while all other results are reproduced in this work.
  }
  \label{tab:main_results}
  \centering
  \resizebox{\linewidth}{!}{
    \begin{NiceTabular}{l|w{c}{5em}|w{c}{5em}|w{c}{5em}|w{c}{5em}|*{5}{w{c}{3em}}}
    \toprule
    &  &  &  & MMStar & \multicolumn{5}{c}{MathVista-Math} \\
    \multirow{-2}{*}{Model} & \multirow{-2}{*}{\#Params} & \multirow{-2}{*}{Avg.} & \multirow{-2}{*}{GeoQA} & Math & ALL & GEO & ALG & GPS & TQA  \\
    \midrule
    \multicolumn{10}{c}{\textit{Closed-Source   Model}} \\
    \midrule
    GPT-4o   \cite{openai2024gpt4ocard} & -- & 58.80 & 40.53 & 69.20 & 66.67\rlap{$^{\dagger}$} & 63.68\rlap{$^{\dagger}$} & 67.04\rlap{$^{\dagger}$} & 63.46\rlap{$^{\dagger}$} & {\ul 77.42\rlap{$^{\dagger}$}} \\
    Claude 3.5 Sonnet   \cite{claude35} & -- & 64.74 & 50.80 & {\ul 76.00} & 67.41\rlap{$^{\dagger}$} & 65.09\rlap{$^{\dagger}$} & 67.79\rlap{$^{\dagger}$} & 65.38\rlap{$^{\dagger}$} & 74.19\rlap{$^{\dagger}$} \\
    \midrule
    \multicolumn{10}{c}{\textit{Open-Source   General Model}} \\
    \midrule
    Qwen3-VL-8B-Instruct$_{w/\,tool}$ \cite{qwen3vl} & 8B & 57.67 & {\ul 67.37} & 46.00 & 59.63 & 67.92 & 59.93 & 68.75 & 29.03 \\
    LLaVA-OV-Qwen2-7b-ov   \cite{li2024llavaov} & 7B & 56.27 & 44.16 & 57.60 & 67.04\rlap{$^{\dagger}$} & 69.34\rlap{$^{\dagger}$} & 67.04\rlap{$^{\dagger}$} & 69.71\rlap{$^{\dagger}$} & 58.06\rlap{$^{\dagger}$} \\
    InternVL2.5-8B   \cite{chen2025intervl25} & 8B & 51.73 & 35.01 & 57.60 & 62.59 & 61.32 & 62.55 & 62.02 & 64.52 \\
    InternVL2.5-8B-MPO   \cite{wang202intervl2_5mpo} & 8B & 66.04 & 52.20 & 70.00 & {\ul 75.93} & {\ul 77.83} & {\ul 76.40} & {\ul 78.37} & 67.74 \\
    DeepSeek-VL2   \cite{wu2024deepseekvl2} & 4.5B & 48.56 & 25.33 & 54.80 & 65.56\rlap{$^{\dagger}$} & 63.68\rlap{$^{\dagger}$} & 65.54\rlap{$^{\dagger}$} & 63.94\rlap{$^{\dagger}$} & 70.97\rlap{$^{\dagger}$} \\
    \midrule
    \multicolumn{10}{c}{\textit{Open-Source   Reasoning Model}} \\
    \midrule
    Pixel Reasoner   \cite{su2025pixelreasoner} & 7B & 64.17 & 53.05 & 72.80 & 66.67 & 64.15 & 67.04 & 64.42 & 74.19 \\
    DeepEyes   \cite{zheng2025deepeyes} & 7B & 64.57 & 49.80 & 72.80 & 71.11 & 70.75 & 71.16 & 71.15 & 70.97 \\
    Mini-o3 \cite{lai2025minio3} & 7B & 60.60 & 46.02 & 68.00 & 67.78 & 67.45 & 67.42 & 67.79 & 67.74  \\
    Hint-GRPO-Qwen2.5-VL-3B   \cite{huang2025hintgrpo} & 3B & 52.65 & 44.30 & 54.40 & 59.26 & 57.55 & 59.18 & 58.65 & 61.29 \\
    Hint-GRPO-Qwen2-VL-7B   \cite{huang2025hintgrpo} & 7B & 53.68 & 48.14 & 59.20 & 53.70 & 51.89 & 53.18 & 52.40 & 58.06 \\
    R1-VL-7B \cite{zhang2025r1vl} & 7B & 61.63 & 46.86 & 68.40\rlap{$^{\dagger}$} & 69.63 & 68.87 & 69.66 & 69.71 & 69.35 \\
    MM-Eureka-Qwen-7B   \cite{meng2025mmeureka} & 7B & {\ul 70.35} & 62.86 & 75.60 & 72.59\rlap{$^{\dagger}$} & 71.22\rlap{$^{\dagger}$} & 72.66\rlap{$^{\dagger}$} & 72.60\rlap{$^{\dagger}$} & 72.58\rlap{$^{\dagger}$} \\
    MINT-CoT-7B   \cite{chen2025mint} & 7B & 69.34 & 64.72\rlap{$^{\dagger}$} & 69.60\rlap{$^{\dagger}$} & 73.70\rlap{$^{\dagger}$} & 74.53\rlap{$^{\dagger}$} & 73.78\rlap{$^{\dagger}$} & 75.00\rlap{$^{\dagger}$} & 69.35\rlap{$^{\dagger}$} \\
    \midrule
    Qwen2.5-VL-3B-Instruct   \cite{qwen25VL} & 3B & 51.60 & 39.66 & 59.20 & 55.93 & 53.30 & 56.18 & 52.88 & 66.13 \\
    \rowcolor[HTML]{F2F2F2}
    \textsc{\textbf{ViRC}}\textbf{-3B (Ours)} & 3B & 70.12 & {\ul 67.37} & 70.40 & 72.59 & 75.47 & 72.66 & 75.96 & 61.29 \\
    \midrule
    Qwen2.5-VL-7B-Instruct   \cite{qwen25VL} & 7B & 58.99 & 43.50\rlap{$^{\dagger}$} & 66.80\rlap{$^{\dagger}$} & 66.67\rlap{$^{\dagger}$} & 65.56\rlap{$^{\dagger}$} & 66.29\rlap{$^{\dagger}$} & 65.87\rlap{$^{\dagger}$} & 69.35\rlap{$^{\dagger}$} \\
    \rowcolor[HTML]{F2F2F2}
    \textsc{\textbf{ViRC}}\textbf{-7B (Ours)} & 7B & \textbf{77.79} & \textbf{75.07} & \textbf{77.20} & \textbf{81.11} & \textbf{81.13} & \textbf{81.27} & \textbf{81.73} & \textbf{79.03} \\ \bottomrule
  \end{NiceTabular}
  }
\end{table*}
During training, the policy $\pi_\theta$ generates a group of $G$ rollouts $\{o_1, \dots, o_G\}$ from a shared history state $s_t = (q; \text{CRU}^{(1)}, \dots, \text{CRU}^{(t)})$, where $q$ denotes the multimodal input query.
Each rollout $o_i$ is either a complete CRU or a final answer. 
Visual tokens returned by tool calls are treated as external observations and excluded from gradient updates, ensuring optimization focuses solely on active reasoning behavior. The reward $r(s_t, o_i)$ combines four components:
\textbf{(1) Answer correctness reward ($r_{\text{ans}}$)}: If $o_i$ is an answer, the reward is set to 1 when the answer is correct and 0 otherwise.
\textbf{(2) Multimodal coherence reward ($r_{\text{mm}}$)}: If $o_i$ is a CRU, the reward is computed using Qwen2.5-VL-72B-Instruct~\cite{qwen25VL} as a judge, which scores text reasoning coherence (weight 0.5) and visual relevance to the intermediate proposition (weight 0.4). Both scores are normalized to the range $[0, 1]$.
\textbf{(3) Reasoning pattern alignment reward ($r_{\text{pattern}}$)}: If $o_i$ is a CRU and the invoked tool type matches the ground-truth label in the hard subset, a reward of 0.1 is assigned. This reward encourages the model to select semantically appropriate tools under specific reasoning patterns such as backtracking, verifying, or reflecting, thereby mimicking human expert strategies.
\textbf{(4) Format validity penalty ($r_{\text{format}}$)}: A penalty of -1 is applied to invalid outputs such as malformed JSON, redundant calls, or improperly placed tool invocations.

Specifically, we optimize $\pi_\theta$ using Group Relative Policy Optimization (GRPO)~\cite{shao2024deepseekmath} and apply a token-wise loss mask to ignore loss on observation tokens not generated by the model. 
Because each CRU encapsulates multiple reasoning steps into a single semantically coherent proposition, rewards are assigned at the CRU level rather than at individual steps. 
This design avoids the noise of dense step-level rewards and mitigates the credit assignment problem inherent in sparse answer-only supervision. The Reason Chunking mechanism thus provides a more stable and semantically meaningful signal for reinforcement learning.
\section{Experiment}

\subsection{Experimental Settings}

\noindent{\textbf{Implementation Details.}}
Full implementation details are provided in the supplementary material
\cref{supp:additional_implementation_details}.

\noindent{\textbf{Test Benchmarks.}}
The proposed \textsc{\textbf{ViRC}} is mainly evaluated on mathematical benchmarks.
GeoQA~\cite{chen2021geoqa} is a geometric problem benchmark with annotated solution programs evaluated on the Geo170K test set consistent with R1-V~\cite{chen2025r1v} and Hint-GRPO~\cite{huang2025hintgrpo}.
MathVista-Math and MMStar-Math are extracted from MathVista~\cite{lu2024mathvista} and MMStar~\cite{chen2024mmstar}, respectively, as mathematical capability dimensions, consistent with MINT-CoT~\cite{chen2025mint}.

While initially designed for the mathematical domain, our method exhibits strong generalization capabilities for vision tasks requiring fine-grained perception. Therefore, we assess \textsc{\textbf{ViRC}} on a suite of high-resolution benchmarks including $V^\ast$~\cite{wu2024vstar}, HR-Bench~\cite{wang2025hrbench}, and VisualProbe~\cite{lai2025minio3}. These benchmarks are characterized by image resolutions ranging from 2K to 16K.

\subsection{Quantitative Results}
\label{sec:quantitative_results}

\noindent\textbf{Comparison with Baselines.}
As shown in \cref{tab:main_results}, \textsc{\textbf{ViRC}} significantly improves the performance of baseline models on multimodal mathematical benchmarks.
It achieves an average accuracy gain of 18.52\% over Qwen2.5-VL-3B-Instruct and 18.80\% over Qwen2.5-VL-7B-Instruct.
The improvement is particularly substantial on the GeoQA benchmark,
which highlights \textsc{\textbf{ViRC}}'s strength in geometric problems. We attribute this large gain to the dynamic visual signals introduced by \textit{Reason Chunking}.

\noindent\textbf{Comparison with State-of-the-Art Models.}
We compare our models against state-of-the-art MLLMs, including closed-source, open-source, and specialized open-source reasoning models.
As illustrated in \cref{tab:main_results}, \textsc{\textbf{ViRC}} achieves the highest overall accuracy across all three benchmarks, outperforming all other models.
Specifically, \textsc{\textbf{ViRC}}-7B surpasses the previous leading open-source model, MM-Eureka, by 7.44\% in average accuracy.
Notably, our much smaller \textsc{\textbf{ViRC}}-3B model performs on par with MM-Eureka.
Compared to the best prior result on each benchmark, \textsc{\textbf{ViRC}}-7B posts gains of +7.7\% on GeoQA, +1.2\% on MMStar-Math, and +5.18\% on MathVista-Math.
This demonstrates that our approach, aligned with human cognitive science, holistically enhances multimodal mathematical reasoning.

It is worth noting that our \textsc{\textbf{ViRC}}-7B surpasses MINT-CoT-7B.
In the context of multimodal mathematical reasoning, MINT-CoT~\cite{chen2025mint} has already demonstrated that token-level perception provides more effective grounding than approaches based on bounding boxes.
Building on this observation, we attribute our overall improvement to the combined effect of the \textsc{\textbf{ViRC}} framework.

\begin{table}[tp]
  \caption{
  \textbf{Generalization results.}
  Our \textsc{\textbf{ViRC}} is trained on the CRUX dataset and cross-domain tested on VisualProbe (VP)~\cite{lai2025minio3}, $V^\ast$~\cite{wu2024vstar}, and HR-Bench (HR)~\cite{wang2025hrbench}. Results marked with $\ast$ are from \cite{lai2025minio3}, while all other results are reproduced in this work.
  }
  \label{tab:generalization}
  \centering
  \resizebox{\columnwidth}{!}{
  \begin{NiceTabular}{l|c|ccc}
    \toprule
    Model & \#Params & VP & $V^\ast$ & HR \\
    \midrule
    GPT-4o   \cite{openai2024gpt4ocard} & -- & 24.70 & 65.20\rlap{$^{\ast}$} & 60.15 \\
    Qwen3-VL-8B-Instruct$_{w/\,tool}$ \cite{qwen3vl} & 8B & 23.07 & 41.88 & 44.87 \\
    LLaVA-OV-Qwen2-7b-ov   \cite{li2024llavaov} & 7B & 20.70 & 70.90\rlap{$^{\ast}$} & 57.60 \\
    InternVL2.5-8B   \cite{chen2025intervl25} & 8B & 26.07 & 68.59 & 58.69 \\
    InternVL2.5-8B-MPO   \cite{wang202intervl2_5mpo} & 8B & 21.02 & 72.25 & 59.69 \\
    DeepSeek-VL2   \cite{wu2024deepseekvl2} & 4.5B & 17.12 & 66.49 & 56.37 \\
    Hint-GRPO-Qwen2.5-VL-3B   \cite{huang2025hintgrpo} & 3B & 28.31 & 68.06 & 60.94 \\
    Hint-GRPO-Qwen2-VL-7B   \cite{huang2025hintgrpo} & 7B & 27.85 & {\ul 75.92} & 61.63 \\
    R1-VL-7B \cite{zhang2025r1vl} & 7B & 20.26 & 45.55 & 50.38 \\
    MM-Eureka-Qwen-7B   \cite{meng2025mmeureka} & 7B & 11.69 & 59.16 & 51.50 \\
    MINT-CoT-7B   \cite{chen2025mint} & 7B & 17.84 & 43.46 & 47.63 \\
    \midrule
    Qwen2.5-VL-3B-Instruct   \cite{qwen25VL} & 3B & 27.25 & 68.06 & 65.00 \\
    \rowcolor[HTML]{F2F2F2}
    \textsc{\textbf{ViRC}}\textbf{-3B (Ours)} & 3B & {\ul 33.62} & 72.25 & {\ul 66.75} \\
    \midrule
    Qwen2.5-VL-7B-Instruct   \cite{qwen25VL} & 7B & 29.67 & 75.50\rlap{$^{\ast}$} & 65.50 \\
    \rowcolor[HTML]{F2F2F2}
    \textsc{\textbf{ViRC}}\textbf{-7B (Ours)} & 7B & \textbf{43.57} & \textbf{79.06} & \textbf{69.94} \\ 
    \bottomrule
  \end{NiceTabular}
  }
\end{table}

\noindent\textbf{Generalization Ability.}
To demonstrate the generalization ability of \textsc{\textbf{ViRC}}, we directly evaluate it on high-resolution benchmarks~\cite{lai2025minio3,wu2024vstar,wang2025hrbench}. As shown in \cref{tab:generalization}, our \textsc{\textbf{ViRC}} achieves competitive performance and robust
generalization ability across all three benchmarks. We attribute this superior performance to its capabilities in accurate localization and visual reasoning, which are essential for addressing the challenges posed by extreme resolutions.

\subsection{Ablation Study}
\label{sec:ablation_study}

We evaluate our method on both mathematical and high-resolution benchmarks with Qwen2.5-VL-7B-Instruct~\cite{qwen25VL} as the baseline.
For brevity, we denote the MathVista-Math~\cite{lu2024mathvista}, VisualProbe~\cite{lai2025minio3}, and HR-Bench datasets~\cite{wang2025hrbench} as MV-M, VP, and HR, respectively, in this subsection.

\begin{table}[tp]
  \caption{\textbf{Effectiveness of CRU}. The proposed cognitively-inspired CRU establishes logical hierarchies to guide tool calls, substantially outperforming the step-wise Visual CoT process. For brevity, MV-M stands for MathVista-Math~\cite{lu2024mathvista}.}
  \label{tab:cru_ablation}
  \centering
  \resizebox{\columnwidth}{!}{
  \begin{NiceTabular}{l|cccccc}
    \toprule
    \multirow{2}{*}{Model} & \multirow{2}{*}{GeoQA} & \Block{2-1}{MMStar\\Math} & \multirow{2}{*}{MV-M} & \multirow{2}{*}{VP} & \multirow{2}{*}{$V^\ast$} & \multirow{2}{*}{HR} \\
    &  &  &  &  &  &  \\
    \midrule
    Baseline & 43.50 & 66.80 & 66.67 & 29.67 & 75.50 & 65.50 \\
    Ours $w/o$ CRU & {\ul 74.67} & {\ul 72.80} & {\ul 79.26} & {\ul 37.18} & {\ul 75.92} & {\ul 67.32} \\
    Ours & \textbf{75.07} & \textbf{77.20} & \textbf{81.11} & \textbf{43.57} & \textbf{79.06} & \textbf{69.94} \\
    \bottomrule
  \end{NiceTabular}
  }
\end{table}

\noindent{\textbf{CRU Effectiveness Analysis.}}
The effectiveness of our proposed CRU is demonstrated in \cref{tab:cru_ablation}. In the absence of CRU, the model is burdened with the concurrent tasks of learning when to segment reasoning chunks and how to utilize visual evidence. This lack of an explicit CRU prior leads the model to misinterpret tool calls as standard visual operations. Consequently, it defaults to a step-wise visual CoT process, which often introduces redundant or ambiguous visual signals into the reasoning chain.

\begin{table}[tp]
    \caption{
    \textbf{Analysis of proposed reasoning patterns.}
    $\mathcal{P}$, $\mathcal{B}$, $\mathcal{R}$, and $\mathcal{V}$ stand for four reasoning patterns, namely planning, backtracking, reflecting, and verifying, respectively.
    }
    \label{tab:reasoning_pattern_ablation}
    \centering
    \resizebox{\columnwidth}{!}{
    \begin{NiceTabular}{cccc|cccccc}
        \toprule
        \multicolumn{4}{c}{Reasoning Pattern} & \multirow{2}{*}{GeoQA} & \Block{2-1}{MMStar\\Math} & \multirow{2}{*}{MV-M} & \multirow{2}{*}{VP} & \multirow{2}{*}{$V^\ast$} & \multirow{2}{*}{HR} \\
        $\mathcal{P}$ & $\mathcal{B}$ & $\mathcal{R}$ & $\mathcal{V}$ &  &  &  &  &  &  \\
        \midrule
        \quad & \quad & \quad & \quad & 43.50 & 66.80 & 66.67 & 29.58 & 71.20 & 65.50 \\
        \checkmark & \quad & \quad & \quad & 72.02 & {\ul 74.80} & {\ul 79.26} & 35.61 & 75.39 & 68.56 \\
        \checkmark & \checkmark & \quad & \quad & 72.81 & 68.40 & 78.52 & 38.57 & 78.01 & {\ul 69.25} \\
        \checkmark & \quad & \checkmark & \quad & 66.98 & 74.00 & 68.89 & 39.97 & {\ul 78.53} & 68.75 \\
        \checkmark & \quad & \quad & \checkmark & {\ul 74.93} & 72.40 & 77.41 & {\ul 40.42} & 78.01 & 69.19 \\
        \checkmark & \checkmark & \checkmark & \checkmark & \textbf{75.07} & \textbf{77.20} & \textbf{81.11} & \textbf{43.57} & \textbf{79.06} & \textbf{69.94} \\
        \bottomrule
    \end{NiceTabular}
    }
\end{table}

\noindent{\textbf{Reasoning Pattern Analysis.}}
Our ablation analysis, detailed in \cref{tab:reasoning_pattern_ablation}, systematically dissects the contribution of four distinct reasoning patterns built upon the Qwen2.5-VL-7B-Instruct~\cite{qwen25VL} model. The initial integration of a planning pattern, analogous to that in MINT-CoT~\cite{chen2025mint}, establishes a strong performance benchmark, underscoring its foundational role in orchestrating the reasoning flow. While the independent addition of backtracking, reflecting, or verifying modules each enhances performance by refining fine-grained visual search, this localized optimization can prove suboptimal. We observe instances where single-pattern augmentation underperforms the planning-only configuration, which we posit occurs when a specialized local strategy conflicts with the dynamic requirements of the broader problem-solving process. This finding highlights a crucial insight: peak performance is not achieved through isolated enhancements but through the synergistic integration of all four patterns. Their complementary nature creates a holistic reasoning framework that is both more powerful and robust.

\begin{table}[t]
  \caption{\textbf{Impact of the three training stages.} \textit{I}, \textit{P}, and \textit{S} stand for Instructional SFT, Practice SFT, and Strategic RL, respectively.}
  \label{tab:training_stage_ablation}
  \centering
  \resizebox{\columnwidth}{!}{
  \begin{NiceTabular}{w{c}{1.5em} w{c}{1.5em} w{c}{1.5em}|cccccc}
    \toprule
    \multicolumn{3}{c}{Training Stage} & \multirow{2}{*}{GeoQA} & \Block{2-1}{MMStar\\Math} & \multirow{2}{*}{MV-M} & \multirow{2}{*}{VP} & \multirow{2}{*}{$V^\ast$} & \multirow{2}{*}{HR} \\
    \textit{I} & \textit{P} & \textit{S} &  &  &  &  &  &  \\
    \midrule
     & & & 43.50 & 66.80 & 66.67 & 29.67 & 75.50 & 65.50 \\
    \checkmark & & & 70.27 & 70.80 & 76.30 & {\ul 33.01} & 74.87 & {\ul 68.88} \\
    \checkmark & \checkmark & & {\ul 73.34} & {\ul 72.80} & {\ul 77.78} & 32.82 & {\ul 75.92} & 67.00 \\
    \checkmark & \checkmark & \checkmark & \textbf{75.07} & \textbf{77.20} & \textbf{81.11} & \textbf{43.57} & \textbf{79.06} & \textbf{69.94} \\
    \bottomrule
  \end{NiceTabular}
}
\end{table}

\noindent{\textbf{Training Stage Analysis.}}
We conducted an ablation study on our three-stage training strategy, with results presented in \cref{tab:training_stage_ablation}.
The initial Instructional SFT stage alone yields substantial performance gains, improving GeoQA by 26.77\% and MathVista-Math by 9.63\%.
This underscores the importance of diverse reasoning patterns and the structured reasoning facilitated by Reason Chunking for mathematical problem-solving.
As training progresses through the Practice SFT and Strategic RL stages, the model's performance continues to improve across most benchmarks.
Notably, Strategic RL delivers the most significant gains on high-resolution tasks.
We attribute this to the comprehensive reward signals provided during the RL stage, which in turn enable the model to make more effective tool calls and better exploit fine-grained visual information.
The model achieves its peak performance upon completion of all three stages.
We posit that this three-stage curriculum is highly effective: Instructional SFT establishes the structural prior for Reason Chunking, Practice SFT applies this prior to real-world visual inputs, and Strategic RL refines decision-making on the hard subset.
Consequently, the complete three-stage strategy is essential for achieving efficient and robust chunked multimodal reasoning.



\begin{figure}[t]
  \centering
\includegraphics[width=\linewidth]{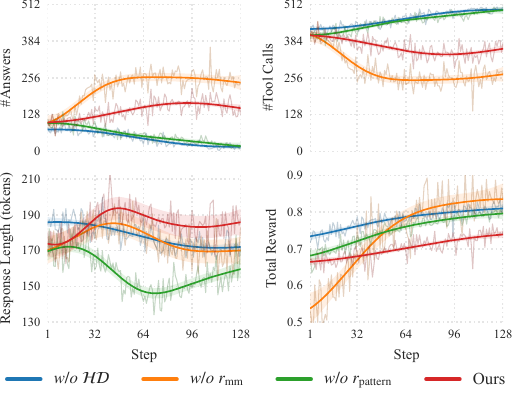}
\vspace{-1em}
  \caption{\textbf{Training dynamics of Strategic RL.}
We visualize how different Strategic RL configurations affect tool call behavior and reasoning progression.
}
  \label{fig:rl_components_ablation}
\end{figure}

\begin{table}[t]
  \caption{\textbf{Ablation of RL components.} The results show that each one of the designs in our proposed Strategic RL plays an important role in multimodal reasoning. For brevity, we denote the curated hard subset derived from CRUX as $\mathcal{HD}$.}
  \label{tab:rl_components_ablation}
  \centering
  \resizebox{\columnwidth}{!}{
  \begin{NiceTabular}{l|cccccc}
    \toprule
    \multirow{2}{*}{Method} & \multirow{2}{*}{GeoQA} & \Block{2-1}{MMStar\\Math} & \multirow{2}{*}{MV-M} & \multirow{2}{*}{VP} & \multirow{2}{*}{$V^\ast$} & \multirow{2}{*}{HR} \\
    &  &  &  &  &  & \\
    \midrule
    Ours & \textbf{75.07} & \textbf{77.20} & \textbf{81.11} & \textbf{43.57} & \textbf{79.06} & \textbf{69.94} \\
    $w/o$ $\mathcal{HD}$ & 66.84 & 70.80 & 73.70 & 35.40 & {\ul 76.44} & 68.75 \\
    $w/o$ $r_{\text{mm}}$ & {\ul 73.60} & 74.00 & 79.63 & 37.92 & \textbf{79.06} & {\ul 69.19} \\
    $w/o$ $r_{\text{pattern}}$ & 73.56 & {\ul 74.80} & {\ul 80.00} & {\ul 38.38} & {\ul 76.44} & {\ul 69.19} \\
    \bottomrule
  \end{NiceTabular}
  }
\end{table}

\noindent{\textbf{RL Component Analysis.}}
In the RL stage, we ablate the key components described in \cref{sec:training_strategy}, including the curated hard subset from CRUX, the multimodal coherence reward $r_{\text{mm}}$, and the reasoning pattern alignment reward $r_{\text{pattern}}$. Note that both the answer correctness reward $r_{\text{ans}}$ and the format validity penalty are always kept.
As shown in \cref{fig:rl_components_ablation}, removing the multimodal reward leads to a classic form of reward hacking, where the model bypasses tool usage and directly guesses answers due to the lack of incentives for maintaining text–image consistency.
On the other hand, removing either the hard subset or $r_{\text{pattern}}$ induces a different failure mode. 
The model overuses tool calls while rarely producing final answers. 
Without the hard subset, RL is driven by randomly sampled and truncated fragments that are too shallow to contain complete solutions, biasing the policy toward repeatedly invoking tools rather than concluding the reasoning. 
Without $r_{\text{pattern}}$, tool usage is no longer grounded in specific reasoning modes, causing unguided exploration and frequent failure to stop and answer. 
Both ablations result in sharply increased tool call frequency and a collapse in answer length, indicating superficial tool manipulation instead of genuine reasoning.
Finally, \cref{tab:rl_components_ablation} shows that the full RL configuration achieves the strongest performance. 
The complete design effectively supports both the exploration and selection of high-quality, image-grounded reasoning paths.

\section{Conclusion}
In this work, we propose \textsc{\textbf{ViRC}}, a novel multimodal reasoning framework specifically designed for visual mathematical problem-solving. 
Drawing inspiration from Miller’s Law, the introduced \textbf{Reason Chunking} mechanism decomposes complex problem-solving into CRUs, enabling dynamic visual verification and coherent intermediate proposition validation. 
To support CRU-aligned reasoning, we construct \textbf{CRUX}, a dataset featuring explicit CRU annotations and diverse reasoning paths. Building on the CRUX dataset, we design a progressive training pipeline comprising Instructional SFT, Practice SFT, and Strategic RL, effectively enhancing the model's ability to alternate and coordinate between visual and textual reasoning. 
Experiments show that \textsc{\textbf{ViRC}}\textbf{-7B} achieves significant performance gains on multiple multimodal mathematical and cross-domain high-resolution image benchmarks. 
The proposed method provides a human-like problem-solving approach for enhanced multimodal mathematical reasoning, with all code publicly released to foster future research.



\section*{Acknowledgement}
\addcontentsline{toc}{section}{Acknowledgement}
This work was supported in part by National Natural Science Foundation of China (No.~62572212), the Ant Group Postdoctoral Programme, the Postdoctoral Science Preferential Funding of Zhejiang Province (No.~ZJ2025214), and the Fundamental Research Funds for the Central Universities.
{
    \small
    \bibliographystyle{ieeenat_fullname}
    \bibliography{main}

@String(CVPR  = {CVPR})

@String(NIPS  = {NeurIPS})

@String(ICLR  = {ICLR})

@String(AAAI  = {AAAI})

@inproceedings{vllm,
  title={Efficient Memory Management for Large Language Model Serving with PagedAttention},
  author={Woosuk Kwon and Zhuohan Li and Siyuan Zhuang and Ying Sheng and Lianmin Zheng and Cody Hao Yu and Joseph E. Gonzalez and Hao Zhang and Ion Stoica},
  booktitle={Proceedings of the ACM SIGOPS 29th Symposium on Operating Systems Principles},
  year={2023}
}

@misc{xiao2024logicvista,
      title={LogicVista: Multimodal LLM Logical Reasoning Benchmark in Visual Contexts}, 
      author={Yijia Xiao and Edward Sun and Tianyu Liu and Wei Wang},
      year={2024},
      eprint={2407.04973},
      archivePrefix={arXiv},
      primaryClass={cs.AI},
      url={https://arxiv.org/abs/2407.04973}, 
}

@article{zhang2024mathverse,
  title={MathVerse: Does Your Multi-modal LLM Truly See the Diagrams in Visual Math Problems?},
  author={Zhang, Renrui and Jiang, Dongzhi and Zhang, Yichi and Lin, Haokun and Guo, Ziyu and Qiu, Pengshuo and Zhou, Aojun and Lu, Pan and Chang, Kai-Wei and Gao, Peng and others},
  journal={arXiv preprint arXiv:2403.14624},
  year={2024}
}

@inproceedings{qiao2025wemath,
  title={We-math: Does your large multimodal model achieve human-like mathematical reasoning?},
  author={Qiao, Runqi and Tan, Qiuna and Dong, Guanting and MinhuiWu, MinhuiWu and Sun, Chong and Song, Xiaoshuai and Wang, Jiapeng and GongQue, Zhuoma and Lei, Shanglin and Zhang, Yifan and others},
  booktitle={Proceedings of the 63rd Annual Meeting of the Association for Computational Linguistics (Volume 1: Long Papers)},
  pages={20023--20070},
  year={2025}
}

@inproceedings{wang2024mathvision,
title={Measuring Multimodal Mathematical Reasoning with MATH-Vision Dataset},
author={Ke Wang and Junting Pan and Weikang Shi and Zimu Lu and Houxing Ren and Aojun Zhou and Mingjie Zhan and Hongsheng Li},
booktitle={The Thirty-eight Conference on Neural Information Processing Systems Datasets and Benchmarks Track},
year={2024},
url={https://openreview.net/forum?id=QWTCcxMpPA}
}

@misc{qwen2vl,
      title={Qwen2-VL: Enhancing Vision-Language Model's Perception of the World at Any Resolution}, 
      author={Qwen Team},
      year={2024},
      eprint={2409.12191},
      archivePrefix={arXiv},
      primaryClass={cs.CV},
      url={https://arxiv.org/abs/2409.12191}, 
}

@article{sheng2024verlhybridflow,
  title   = {HybridFlow: A Flexible and Efficient RLHF Framework},
  author  = {Guangming Sheng and Chi Zhang and Zilingfeng Ye and Xibin Wu and Wang Zhang and Ru Zhang and Yanghua Peng and Haibin Lin and Chuan Wu},
  year    = {2024},
  journal = {arXiv preprint arXiv: 2409.19256}
}

@inproceedings{zheng2024llamafactory,
  title={LlamaFactory: Unified Efficient Fine-Tuning of 100+ Language Models},
  author={Yaowei Zheng and Richong Zhang and Junhao Zhang and Yanhan Ye and Zheyan Luo and Zhangchi Feng and Yongqiang Ma},
  booktitle={Proceedings of the 62nd Annual Meeting of the Association for Computational Linguistics (Volume 3: System Demonstrations)},
  address={Bangkok, Thailand},
  publisher={Association for Computational Linguistics},
  year={2024},
  url={http://arxiv.org/abs/2403.13372}
}

@misc{zhong2025omnir1,
      title={Omni-R1: Reinforcement Learning for Omnimodal Reasoning via Two-System Collaboration}, 
      author={Hao Zhong and Muzhi Zhu and Zongze Du and Zheng Huang and Canyu Zhao and Mingyu Liu and Wen Wang and Hao Chen and Chunhua Shen},
      year={2025},
      eprint={2505.20256},
      archivePrefix={arXiv},
      primaryClass={cs.CV},
      url={https://arxiv.org/abs/2505.20256}, 
}

@misc{liu2025visualagenticreinforcementfinetuning,
      title={Visual Agentic Reinforcement Fine-Tuning}, 
      author={Ziyu Liu and Yuhang Zang and Yushan Zou and Zijian Liang and Xiaoyi Dong and Yuhang Cao and Haodong Duan and Dahua Lin and Jiaqi Wang},
      year={2025},
      eprint={2505.14246},
      archivePrefix={arXiv},
      primaryClass={cs.CV},
      url={https://arxiv.org/abs/2505.14246}, 
}

@misc{hu2024visualsketchpad,
      title={Visual Sketchpad: Sketching as a Visual Chain of Thought for Multimodal Language Models}, 
      author={Yushi Hu and Weijia Shi and Xingyu Fu and Dan Roth and Mari Ostendorf and Luke Zettlemoyer and Noah A Smith and Ranjay Krishna},
      year={2024},
      eprint={2406.09403},
      archivePrefix={arXiv},
      primaryClass={cs.CV},
      url={https://arxiv.org/abs/2406.09403}, 
}

@misc{wu2025reinforcingspatialreasoning,
      title={Reinforcing Spatial Reasoning in Vision-Language Models with Interwoven Thinking and Visual Drawing}, 
      author={Junfei Wu and Jian Guan and Kaituo Feng and Qiang Liu and Shu Wu and Liang Wang and Wei Wu and Tieniu Tan},
      year={2025},
      eprint={2506.09965},
      archivePrefix={arXiv},
      primaryClass={cs.CV},
      url={https://arxiv.org/abs/2506.09965}, 
}

@misc{jiang2025vlmr3,
      title={VLM-R$^3$: Region Recognition, Reasoning, and Refinement for Enhanced Multimodal Chain-of-Thought}, 
      author={Chaoya Jiang and Yongrui Heng and Wei Ye and Han Yang and Haiyang Xu and Ming Yan and Ji Zhang and Fei Huang and Shikun Zhang},
      year={2025},
      eprint={2505.16192},
      archivePrefix={arXiv},
      primaryClass={cs.CV},
      url={https://arxiv.org/abs/2505.16192}, 
}

@misc{su2025openthinkimg,
      title={OpenThinkIMG: Learning to Think with Images via Visual Tool Reinforcement Learning}, 
      author={Zhaochen Su and Linjie Li and Mingyang Song and Yunzhuo Hao and Zhengyuan Yang and Jun Zhang and Guanjie Chen and Jiawei Gu and Juntao Li and Xiaoye Qu and Yu Cheng},
      year={2025},
      eprint={2505.08617},
      archivePrefix={arXiv},
      primaryClass={cs.CV},
      url={https://arxiv.org/abs/2505.08617}, 
}

@misc{kalarani2024unveilinginvisiblecaptioningvideos,
      title={Unveiling the Invisible: Captioning Videos with Metaphors}, 
      author={Abisek Rajakumar Kalarani and Pushpak Bhattacharyya and Sumit Shekhar},
      year={2024},
      eprint={2406.04886},
      archivePrefix={arXiv},
      primaryClass={cs.CV},
      url={https://arxiv.org/abs/2406.04886}, 
}

@misc{wu2024zeroshotlongformvideounderstanding,
      title={Zero-Shot Long-Form Video Understanding through Screenplay}, 
      author={Yongliang Wu and Bozheng Li and Jiawang Cao and Wenbo Zhu and Yi Lu and Weiheng Chi and Chuyun Xie and Haolin Zheng and Ziyue Su and Jay Wu and Xu Yang},
      year={2024},
      eprint={2406.17309},
      archivePrefix={arXiv},
      primaryClass={cs.CV},
      url={https://arxiv.org/abs/2406.17309}, 
}

@misc{min2025morevqa,
      title={MoReVQA: Exploring Modular Reasoning Models for Video Question Answering}, 
      author={Juhong Min and Shyamal Buch and Arsha Nagrani and Minsu Cho and Cordelia Schmid},
      year={2025},
      eprint={2404.06511},
      archivePrefix={arXiv},
      primaryClass={cs.CV},
      url={https://arxiv.org/abs/2404.06511}, 
}

@inproceedings{liang2025improvingmllmocr,
    title = "Improving {MLLM}{'}s Document Image Machine Translation via Synchronously Self-reviewing Its {OCR} Proficiency",
    author = "Liang, Yupu  and
      Zhang, Yaping  and
      Zhang, Zhiyang  and
      Chen, Zhiyuan  and
      Zhao, Yang  and
      Xiang, Lu  and
      Zong, Chengqing  and
      Zhou, Yu",
    editor = "Che, Wanxiang  and
      Nabende, Joyce  and
      Shutova, Ekaterina  and
      Pilehvar, Mohammad Taher",
    booktitle = "ACL",
    month = jul,
    year = "2025",
    address = "Vienna, Austria",
    publisher = "Association for Computational Linguistics",
    url = "https://aclanthology.org/2025.findings-acl.1213/",
    doi = "10.18653/v1/2025.findings-acl.1213",
    pages = "23659--23678",
    ISBN = "979-8-89176-256-5",
}

@INPROCEEDINGS{heng2025rodmllm,
  author={Yin, Heng and Ren, Yuqiang and Yan, Ke and Ding, Shouhong and Hao, Yongtao},
  booktitle={CVPR}, 
  title={ROD-MLLM: Towards More Reliable Object Detection in Multimodal Large Language Models}, 
  year={2025},
  volume={},
  number={},
  pages={14358-14368},
  doi={10.1109/CVPR52734.2025.01339}}

@inproceedings{bao2025mllmi2w,
    title = "{MLLM}-{I}2{W}: Harnessing Multimodal Large Language Model for Zero-Shot Composed Image Retrieval",
    author = "Bao, Tong  and
      Liu, Che  and
      Xu, Derong  and
      Zheng, Zhi  and
      Xu, Tong",
    editor = "Rambow, Owen  and
      Wanner, Leo  and
      Apidianaki, Marianna  and
      Al-Khalifa, Hend  and
      Eugenio, Barbara Di  and
      Schockaert, Steven",
    booktitle = "Proceedings of the 31st International Conference on Computational Linguistics",
    month = jan,
    year = "2025",
    address = "Abu Dhabi, UAE",
    publisher = "Association for Computational Linguistics",
    url = "https://aclanthology.org/2025.coling-main.125/",
    pages = "1839--1849",
}

@misc{deepseekr1,
      title={DeepSeek-R1: Incentivizing Reasoning Capability in LLMs via Reinforcement Learning}, 
      author={DeepSeek-AI},
      year={2025},
      eprint={2501.12948},
      archivePrefix={arXiv},
      primaryClass={cs.CL},
      url={https://arxiv.org/abs/2501.12948}, 
}

@misc{deepseekv3,
      title={DeepSeek-V3 Technical Report}, 
      author={DeepSeek-AI},
      year={2025},
      eprint={2412.19437},
      archivePrefix={arXiv},
      primaryClass={cs.CL},
      url={https://arxiv.org/abs/2412.19437}, 
}

@misc{gpt4,
      title={GPT-4 Technical Report}, 
      author={OpenAI Team},
      year={2024},
      eprint={2303.08774},
      archivePrefix={arXiv},
      primaryClass={cs.CL},
      url={https://arxiv.org/abs/2303.08774}, 
}

@misc{llama3,
      title={The Llama 3 Herd of Models}, 
      author={Llama Team},
      year={2024},
      eprint={2407.21783},
      archivePrefix={arXiv},
      primaryClass={cs.AI},
      url={https://arxiv.org/abs/2407.21783}, 
}

@article{qwen3,
    title={Qwen3 Technical Report}, 
    author={Qwen Team},
    journal = {arXiv preprint arXiv:2505.09388},
    year={2025}
}

@article{qwen25,
    title   = {Qwen2.5 Technical Report}, 
    author  = {Qwen Team},
    journal = {arXiv preprint arXiv:2412.15115},
    year    = {2024}
}

@misc{gandhi2025stars,
      title={Cognitive Behaviors that Enable Self-Improving Reasoners, or, Four Habits of Highly Effective STaRs}, 
      author={Kanishk Gandhi and Ayush Chakravarthy and Anikait Singh and Nathan Lile and Noah D. Goodman},
      year={2025},
      eprint={2503.01307},
      archivePrefix={arXiv},
      primaryClass={cs.CL},
      url={https://arxiv.org/abs/2503.01307}, 
}

@misc{shao2024deepseekmath,
      title={DeepSeekMath: Pushing the Limits of Mathematical Reasoning in Open Language Models}, 
      author={Zhihong Shao and Peiyi Wang and Qihao Zhu and Runxin Xu and Junxiao Song and Xiao Bi and Haowei Zhang and Mingchuan Zhang and Y. K. Li and Y. Wu and Daya Guo},
      year={2024},
      eprint={2402.03300},
      archivePrefix={arXiv},
      primaryClass={cs.CL},
      url={https://arxiv.org/abs/2402.03300}, 
}

@misc{xu2025visualplanning,
      title={Visual Planning: Let's Think Only with Images}, 
      author={Yi Xu and Chengzu Li and Han Zhou and Xingchen Wan and Caiqi Zhang and Anna Korhonen and Ivan Vulić},
      year={2025},
      eprint={2505.11409},
      archivePrefix={arXiv},
      primaryClass={cs.LG},
      url={https://arxiv.org/abs/2505.11409}, 
}

@misc{chern2025thinkingwithgeneratedimages,
      title={Thinking with Generated Images}, 
      author={Ethan Chern and Zhulin Hu and Steffi Chern and Siqi Kou and Jiadi Su and Yan Ma and Zhijie Deng and Pengfei Liu},
      year={2025},
      eprint={2505.22525},
      archivePrefix={arXiv},
      primaryClass={cs.CV},
      url={https://arxiv.org/abs/2505.22525}, 
}

@misc{shen2025zoomeye,
      title={ZoomEye: Enhancing Multimodal LLMs with Human-Like Zooming Capabilities through Tree-Based Image Exploration}, 
      author={Haozhan Shen and Kangjia Zhao and Tiancheng Zhao and Ruochen Xu and Zilun Zhang and Mingwei Zhu and Jianwei Yin},
      year={2025},
      eprint={2411.16044},
      archivePrefix={arXiv},
      primaryClass={cs.CV},
      url={https://arxiv.org/abs/2411.16044}, 
}

@misc{gpt4v,
  title={{GPT-4V(ision)} System Card},
  author={OpenAI},
  year={2023},
url={https://openai.com/research/gpt-4v-system-card}
}

@misc{wang2025simpleo3,
      title={Simple o3: Towards Interleaved Vision-Language Reasoning}, 
      author={Ye Wang and Qianglong Chen and Zejun Li and Siyuan Wang and Shijie Guo and Zhirui Zhang and Zhongyu Wei},
      year={2025},
      eprint={2508.12109},
      archivePrefix={arXiv},
      primaryClass={cs.CV},
      url={https://arxiv.org/abs/2508.12109}, 
}

@misc{zhang2025cof,
      title={Chain-of-Focus: Adaptive Visual Search and Zooming for Multimodal Reasoning via RL}, 
      author={Xintong Zhang and Zhi Gao and Bofei Zhang and Pengxiang Li and Xiaowen Zhang and Yang Liu and Tao Yuan and Yuwei Wu and Yunde Jia and Song-Chun Zhu and Qing Li},
      year={2025},
      eprint={2505.15436},
      archivePrefix={arXiv},
      primaryClass={cs.CV},
      url={https://arxiv.org/abs/2505.15436}, 
}

@misc{yao2024mulberry,
      title={Mulberry: Empowering MLLM with o1-like Reasoning and Reflection via Collective Monte Carlo Tree Search}, 
      author={Huanjin Yao and Jiaxing Huang and Wenhao Wu and Jingyi Zhang and Yibo Wang and Shunyu Liu and Yingjie Wang and Yuxin Song and Haocheng Feng and Li Shen and Dacheng Tao},
      year={2024},
      eprint={2412.18319},
      archivePrefix={arXiv},
      primaryClass={cs.CV},
      url={https://arxiv.org/abs/2412.18319}, 
}

@article{miller1956miller,
author = {Miller, G.},
title = {The magical number seven, plus or minus two: some limits on our capacity for processing information.},
journal = {Psychological Review},
year = {1956},
volume = {63},
issue = {2},
pages = {81-97},
doi = {10.1037/h0043158}
}

@online{qwen3vl,
  title={Qwen3-VL Technical Report}, 
  author={Qwen Team},
  year={2025},
  eprint={2511.21631},
  archivePrefix={arXiv},
  primaryClass={cs.CV},
  url={https://arxiv.org/abs/2511.21631}, 
}

@inproceedings{kojima2022zeroshotreasoners,
author = {Kojima, Takeshi and Gu, Shixiang (Shane) and Reid, Machel and Matsuo, Yutaka and Iwasawa, Yusuke},
booktitle = {NIPS},
editor = {S. Koyejo and S. Mohamed and A. Agarwal and D. Belgrave and K. Cho and A. Oh},
pages = {22199--22213},
publisher = {Curran Associates, Inc.},
title = {Large Language Models are Zero-Shot Reasoners},
volume = {35},
year = {2022}
}

@inproceedings{wei2022cotprompting,
author = {Wei, Jason and Wang, Xuezhi and Schuurmans, Dale and Bosma, Maarten and ichter, brian and Xia, Fei and Chi, Ed and Le, Quoc V and Zhou, Denny},
booktitle = {NIPS},
editor = {S. Koyejo and S. Mohamed and A. Agarwal and D. Belgrave and K. Cho and A. Oh},
pages = {24824--24837},
publisher = {Curran Associates, Inc.},
title = {Chain-of-Thought Prompting Elicits Reasoning in Large Language Models},
volume = {35},
year = {2022}
}

@inproceedings{wang2025hrbench,
  title={Divide, conquer and combine: A training-free framework for high-resolution image perception in multimodal large language models},
  author={Wang, Wenbin and Ding, Liang and Zeng, Minyan and Zhou, Xiabin and Shen, Li and Luo, Yong and Yu, Wei and Tao, Dacheng},
  booktitle={AAAI},
  volume={39},
  number={8},
  pages={7907--7915},
  year={2025}
}

@inproceedings{wu2024vstar,
  title={$V^\ast$: Guided visual search as a core mechanism in multimodal llms},
  author={Wu, Penghao and Xie, Saining},
  booktitle={CVPR},
  pages={13084--13094},
  year={2024}
}

@article{chen2024mmstar,
  title={Are we on the right way for evaluating large vision-language models?},
  author={Chen, Lin and Li, Jinsong and Dong, Xiaoyi and Zhang, Pan and Zang, Yuhang and Chen, Zehui and Duan, Haodong and Wang, Jiaqi and Qiao, Yu and Lin, Dahua and others},
  journal={NIPS},
  volume={37},
  pages={27056--27087},
  year={2024}
}

@inproceedings{lu2024mathvista,
  title={MathVista: Evaluating Mathematical Reasoning of Foundation Models in Visual Contexts},
  author={Lu, Pan and Bansal, Hritik and Xia, Tony and Liu, Jiacheng and Li, Chunyuan and Hajishirzi, Hannaneh and Cheng, Hao and Chang, Kai-Wei and Galley, Michel and Gao, Jianfeng},
  booktitle={ICLR},
  year={2024}
}

@inproceedings{chen2021geoqa,
    title = "{G}eo{QA}: A Geometric Question Answering Benchmark Towards Multimodal Numerical Reasoning",
    author = "Chen, Jiaqi  and
      Tang, Jianheng  and
      Qin, Jinghui  and
      Liang, Xiaodan  and
      Liu, Lingbo  and
      Xing, Eric  and
      Lin, Liang",
    editor = "Zong, Chengqing  and
      Xia, Fei  and
      Li, Wenjie  and
      Navigli, Roberto",
    booktitle = "Findings of the Association for Computational Linguistics: ACL-IJCNLP 2021",
    month = aug,
    year = "2021",
    address = "Online",
    publisher = "Association for Computational Linguistics",
    url = "https://aclanthology.org/2021.findings-acl.46/",
    doi = "10.18653/v1/2021.findings-acl.46",
    pages = "513--523"
}

@misc{chen2025r1v,
  author       = {Chen, Liang and Li, Lei and Zhao, Haozhe and Song, Yifan and Vinci},
  title        = {R1-V: Reinforcing Super Generalization Ability in Vision-Language Models with Less Than \$3},
  howpublished = {\url{https://github.com/Deep-Agent/R1-V}},
  note         = {Accessed: 2025-02-02},
  year         = {2025}
}

@article{chen2025mint,
title={MINT-CoT: Enabling Interleaved Visual Tokens in Mathematical Chain-of-Thought Reasoning},
author={Chen, Xinyan and Zhang, Renrui and Jiang, Dongzhi and Zhou, Aojun and Yan, Shilin and Lin, Weifeng and Li, Hongsheng},
journal=NIPS,
year={2025}
}

@misc{zheng2025deepeyes,
title = {DeepEyes: Incentivizing "Thinking with Images" via Reinforcement Learning},
author = {Zheng, Ziwei and Yang, Michael and Hong, Jack and Zhao, Chenxiao and Xu, Guohai and Yang, Le and Shen, Chao and Yu, Xing},
year = {2025},
eprint = {2505.14362},
archivePrefix = {arXiv},
primaryClass = {cs.CV},
url = {https://arxiv.org/abs/2505.14362}
}

@article{lai2025minio3,
title={Mini-o3: Scaling Up Reasoning Patterns and Interaction Turns for Visual Search},
author={Lai, Xin and Li, Junyi and Li, Wei and Liu, Tao and Li, Tianjian and Zhao, Hengshuang},
journal={arXiv:2509.07969},
year={2025}
}

@misc{qwen25VL,
title = {Qwen2.5-VL},
url = {https://qwenlm.github.io/blog/qwen2.5-vl/},
author = {Qwen Team},
month = {January},
year = {2025}
}

@article{zhang2025r1vl,
title={R1-VL: Learning to Reason with Multimodal Large Language Models via Step-wise Group Relative Policy Optimization},
author={Zhang, Jingyi and Huang, Jiaxing and Yao, Huanjin and Liu, Shunyu and Zhang, Xikun and Lu, Shijian and Tao, Dacheng},
journal={arXiv preprint arXiv:2503.12937},
year={2025}
}

@misc{huang2025hintgrpo,
title={Boosting MLLM Reasoning with Text-Debiased Hint-GRPO},
author={Qihan Huang and Weilong Dai and Jinlong Liu and Wanggui He and Hao Jiang and Mingli Song and Jingyuan Chen and Chang Yao and Jie Song},
year={2025},
eprint={2503.23905},
archivePrefix={arXiv},
primaryClass={cs.CV},
url={https://arxiv.org/abs/2503.23905},
}

@article{su2025pixelreasoner,
title={Pixel Reasoner: Incentivizing Pixel-Space Reasoning with Curiosity-Driven Reinforcement Learning},
author = {Su, Alex and Wang, Haozhe and Ren, Weiming and Lin, Fangzhen and Chen, Wenhu},
journal={arXiv preprint arXiv:2505.15966},
year={2025}
}

@misc{wu2024deepseekvl2,
title={DeepSeek-VL2: Mixture-of-Experts Vision-Language Models for Advanced Multimodal Understanding},
author={Zhiyu Wu and Xiaokang Chen and Zizheng Pan and Xingchao Liu and Wen Liu and Damai Dai and Huazuo Gao and Yiyang Ma and Chengyue Wu and Bingxuan Wang and Zhenda Xie and Yu Wu and Kai Hu and Jiawei Wang and Yaofeng Sun and Yukun Li and Yishi Piao and Kang Guan and Aixin Liu and Xin Xie and Yuxiang You and Kai Dong and Xingkai Yu and Haowei Zhang and Liang Zhao and Yisong Wang and Chong Ruan},
year={2024},
eprint={2412.10302},
archivePrefix={arXiv},
primaryClass={cs.CV},
url={https://arxiv.org/abs/2412.10302},
}

@article{meng2025mmeureka,
title={MM-Eureka: Exploring the Frontiers of Multimodal Reasoning with Rule-based Reinforcement Learning},
author={Fanqing Meng and Lingxiao Du and Zongkai Liu and Zhixiang Zhou and Quanfeng Lu and Daocheng Fu and Tiancheng Han and Botian Shi and Wenhai Wang and Junjun He and Kaipeng Zhang and Ping Luo and Yu Qiao and Qiaosheng Zhang and Wenqi Shao},
year={2025},
journal={arXiv preprint arXiv:2503.07365},
}

@misc{chen2025intervl25,
      title={Expanding Performance Boundaries of Open-Source Multimodal Models with Model, Data, and Test-Time Scaling}, 
      author={Zhe Chen and Weiyun Wang and Yue Cao and Yangzhou Liu and Zhangwei Gao and Erfei Cui and Jinguo Zhu and Shenglong Ye and Hao Tian and Zhaoyang Liu and Lixin Gu and Xuehui Wang and Qingyun Li and Yiming Ren and Zixuan Chen and Jiapeng Luo and Jiahao Wang and Tan Jiang and Bo Wang and Conghui He and Botian Shi and Xingcheng Zhang and Han Lv and Yi Wang and Wenqi Shao and Pei Chu and Zhongying Tu and Tong He and Zhiyong Wu and Huipeng Deng and Jiaye Ge and Kai Chen and Kaipeng Zhang and Limin Wang and Min Dou and Lewei Lu and Xizhou Zhu and Tong Lu and Dahua Lin and Yu Qiao and Jifeng Dai and Wenhai Wang},
      year={2025},
      eprint={2412.05271},
      archivePrefix={arXiv},
      primaryClass={cs.CV},
      url={https://arxiv.org/abs/2412.05271}, 
}

@misc{wang202intervl2_5mpo,
      title={Enhancing the Reasoning Ability of Multimodal Large Language Models via Mixed Preference Optimization}, 
      author={Weiyun Wang and Zhe Chen and Wenhai Wang and Yue Cao and Yangzhou Liu and Zhangwei Gao and Jinguo Zhu and Xizhou Zhu and Lewei Lu and Yu Qiao and Jifeng Dai},
      year={2025},
      eprint={2411.10442},
      archivePrefix={arXiv},
      primaryClass={cs.CL},
      url={https://arxiv.org/abs/2411.10442}, 
}

@misc{li2024llavaov,
      title={LLaVA-OneVision: Easy Visual Task Transfer}, 
      author={Bo Li and Yuanhan Zhang and Dong Guo and Renrui Zhang and Feng Li and Hao Zhang and Kaichen Zhang and Peiyuan Zhang and Yanwei Li and Ziwei Liu and Chunyuan Li},
      year={2024},
      eprint={2408.03326},
      archivePrefix={arXiv},
      primaryClass={cs.CV},
      url={https://arxiv.org/abs/2408.03326}, 
}

@inproceedings{claude35,
  title={Model Card Addendum: Claude 3.5 Haiku and Upgraded Claude 3.5 Sonnet},
  author={Sonnet Anthropic},
  url={https://api.semanticscholar.org/CorpusID:273639283},
year={2024},
}

@misc{openai2024gpt4ocard,
title={GPT-4o System Card},
author={OpenAI Team},
year={2024},
eprint={2410.21276},
archivePrefix={arXiv},
primaryClass={cs.CL},
url={https://arxiv.org/abs/2410.21276},
}
}

\clearpage
\setcounter{section}{0}
\setcounter{page}{1}
\maketitlesupplementary
\appendix





\blfootnote{$^{\dagger}$Work done during Lihong Wang's internship at Ant Digital Technologies, Ant Group.}
\blfootnote{$^{\ast}$Corresponding authors.}


\section*{Overview}
\label{sec:overview}
We organize our supplementary material as follows.
\begin{itemize}
  \item \hyperref[supp:millerslaw]{Background on Miller’s Law}
  \item \hyperref[supp:dataset_details]{Dataset Details}
  \begin{itemize}
    \item \hyperref[supp:dataset_statistic]{Dataset Statistic}
    \item \hyperref[supp:dataset_curation]{Dataset Curation}
    \item \hyperref[supp:modeling_cognitive_reasoning_patterns]{Modeling Cognitive Reasoning Patterns}
  \end{itemize}
  \item \hyperref[supp:additional_implementation_details]{Additional Implementation Details}
  \begin{itemize}
    \item \hyperref[supp:instructional_sft]{Instructional SFT}
    \item \hyperref[supp:practice_sft]{Practice SFT}
    \item \hyperref[supp:strategic_rl]{Strategic RL}
    \item \hyperref[supp:hard_subset_curation]{Hard Subset Curation}
    \item \hyperref[supp:reward_design]{Reward Design}
  \end{itemize}
  \item \hyperref[supp:evaluation_protocol]{Evaluation Protocol}
  \item \hyperref[supp:additional_results]{Additional Results}
  \begin{itemize}
    \item \hyperref[supp:additional_quantitative_results]{Additional Quantitative Results}
    \item \hyperref[supp:detailed_ablation_study_results]{Detailed Ablation Study Results}
    \item \hyperref[supp:additional_qualitative_results]{Additional Qualitative Results}
  \end{itemize}
  \item \hyperref[supp:dataset_and_prompt_details]{Dataset and Prompt Details}
  \begin{itemize}
    \item \hyperref[supp:dataset_example]{Dataset Example}
    \item \hyperref[supp:prompts_for_data_curation]{Prompts for Data Curation}
    \item \hyperref[supp:prompts_for_sft]{Prompts for Supervised Fine-Tuning}
    \item \hyperref[supp:prompts_for_rl]{Prompts for Reinforcement Learning}
  \end{itemize}
\end{itemize}

\section{Background on Miller’s Law}
\label{supp:millerslaw}

For a better understanding of the proposed \textbf{Reason Chunking} mechanism, this section provides a concise explanation of Miller’s Law~\cite{miller1956miller} and its relevance to multimodal mathematical reasoning.
Miller’s Law posits that the capacity of human short-term memory is limited, typically to around seven items.
To overcome this cognitive bottleneck, humans group related low-level information into semantically meaningful higher-level units, or \textit{chunks}.
This principle forms the initial motivation behind structuring complex reasoning into manageable and semantically coherent components.

\noindent\textbf{Miller’s Law and human information organization.}
Miller’s Law suggests that short-term memory can only maintain a small number of items at once.
By applying \textit{information chunking}, multiple low-level elements can be reorganized into higher-level semantic units, substantially reducing cognitive load.
For instance, eight unrelated letters such as “C V P R S U P P” become far easier to recall when reorganized into the two meaningful chunks “CVPR SUPP”.
This illustrates a general cognitive principle: humans naturally process complex inputs by forming compact, interpretable units, which improves reasoning clarity and reduces memory stress.
This principle directly inspires the idea that long reasoning chains should be decomposed into structured intermediate units rather than treated as flat sequences.

\noindent\textbf{Information chunking in mathematical and visual reasoning.}
The principle of information chunking is evident in mathematical problem-solving.
Humans decompose a complex problem into a series of intermediate propositions, perform localized reasoning for each, and then integrate these propositions hierarchically.
This strategy avoids the cognitive overload that would result from maintaining a lengthy, linear reasoning chain in working memory.
The scoring systems of mathematical competitions like the IMO and AMC reinforce this natural structure: \textit{scoring points} are awarded for proving intermediate results, acknowledging that mathematical reasoning is inherently composed of semantically meaningful information chunks.
This cognitive pattern extends to visual reasoning.
Instead of re-examining an entire image at every step, humans selectively attend to relevant visual cues only when needed.
This prevents an excess of visual information from overwhelming short-term memory.
Together, these behaviors align with Miller’s insight that effective reasoning depends on forming and operating on appropriately sized informational chunks.

\noindent\textbf{Operationalizing Miller’s principle in \textsc{\textbf{ViRC}}.}
Guided by this chunk-based cognitive pattern, we design the \textit{Reason Chunking} mechanism to operationalize Miller’s Law within multimodal mathematical reasoning.
The overall reasoning process is structured into a sequence of \textbf{Critical Reasoning Units (CRUs)}, each functioning as a semantically complete chunk that encapsulates a single intermediate proposition.
These CRUs form a hierarchical reasoning path that mirrors human decomposition of complex problems.
Moreover, \textit{selective visual grounding} is applied only between CRUs, simulating the human tendency to consult visual information at critical points rather than continuously.
In this way, \textsc{\textbf{ViRC}} leverages the chunk-based processing principle highlighted by Miller’s Law to produce structured, cognitively aligned multimodal reasoning.

\section{Dataset Details}
\label{supp:dataset_details}

This section provides a comprehensive overview of our constructed \textbf{CRUX} dataset. 
We first present its overall statistics in \cref{supp:dataset_statistic}. 
Next, we elaborate on the automated curation pipeline in \cref{supp:dataset_curation}, including how diverse reasoning paths are sampled and mapped to Critical Reasoning Units (CRUs). 
Finally, we describe how the four cognitive reasoning patterns of Planning, Reflecting, Verifying, and Backtracking are modeled and systematically integrated into the dataset in \cref{supp:modeling_cognitive_reasoning_patterns}.

\subsection{Dataset Statistic}
\label{supp:dataset_statistic}
The key statistics of the CRUX dataset are summarized in \cref{tab:supp_dataset_statistics}. 
This includes the total number of data points, the distribution of Critical Reasoning Units (CRUs) per sample, and the frequency of various tool calls and cognitive reasoning patterns.
\begin{table}[ht]
  \caption{Key statistics of the CRUX dataset.}
  \label{tab:supp_dataset_statistics}
  \centering
  \resizebox{\columnwidth}{!}{
  \begin{NiceTabular}{lc}
    \toprule
    Statistic & Value \\
    \midrule
    Total data points & 100,000 \\
    Average resolution of query images & 431 × 409 px \\
    Minimum CRUs per data point & 1 \\
    Maximum CRUs per data point & 21 \\
    Average CRUs per data point & 4.27 \\
    Total \texttt{crop} tool calls & 281,248 \\
    Total \texttt{scale} tool calls & 11,492 \\
    Total \texttt{display} tool calls & 60,034 \\
    Total “Planning” pattern occurrences & 100,000 \\
    Total “Backtracking” pattern occurrences & 9,682 \\
    Total “Verifying” pattern occurrences & 38,226 \\
    Total “Reflecting” pattern occurrences & 179,008 \\
    \bottomrule
  \end{NiceTabular}
  }
\end{table}

\subsection{Dataset Curation}
\label{supp:dataset_curation}

To support structured multimodal reasoning based on Critical Reasoning Units (CRUs), we construct the CRUX dataset via an automated pipeline involving four specialized LLM/VLM components
(see the main paper Fig.~3).
While CRUX inherits raw problem–image pairs from MINT-CoT~\cite{chen2025mint}, all reasoning paths and CRU annotations are newly generated using carefully designed prompts (see \cref{supp:prompts_for_data_curation}), as detailed below.

\noindent{\textbf{Sampling Diverse Reasoning Paths.}}
For each image, we generate five scaled variants ($f \in \{0.25, 0.5, 1, 2, 4\}$) and constrain image token count to $[4, 16384]$.
Using zero-shot CoT (see \cref{fig:supp_prompt_sampling_sample}), we sample five reasoning paths per scale and verify answer correctness.
We select scales $f_-$ (low accuracy) and $f_+$ (high accuracy) with a $\geq 60\%$ accuracy gap, preferring maximal resolution difference.
From these, we take the longest correct path ($p_0$) and longest incorrect paths ($p_1$ at $f_-$, $p_2$ at $f_+$) as the source paths for subsequent composition.

\noindent{\textbf{Mapping Steps to CRUs.}}
Each path is decomposed into fine-grained steps annotated with a focus object (e.g., ``$\angle ABC$'').
Consecutive steps sharing the same focus object form one CRU, representing a single intermediate proposition.
Using $p_0$ as the semantic anchor, the Mapper aligns each step in $p_1$ and $p_2$ to its most semantically similar step in $p_0$, thereby establishing a correspondence not only between steps but also between their associated intermediate propositions (i.e., CRUs in $p_0$). 
The Mapper then identifies the first erroneous step in each incorrect path ($\text{err}_1$ in $p_1$, $\text{err}_2$ in $p_2$) along with its aligned proposition in $p_0$. 
Only the steps up to and including the first error are retained, and these are grouped according to the CRU in $p_0$ to which they map.
Finally, CRUs from $p_1$, $p_2$, and $p_0$ are concatenated as $p_1 \rightarrow p_2 \rightarrow p_0$ to simulate error-aware reasoning.

\noindent{\textbf{Grounding CRUs.}}
For each CRU, the Linker detects the focus object and associated textual annotations in the original image.
Their bounding boxes are fused into a unified grounding region, prioritizing text when available.
Each CRU is linked to a \texttt{crop} tool call targeting this region.
Additionally, the Linker generates image descriptions, rationales, and guiding questions to ensure coherent transitions between CRUs.

\subsection{Modeling Cognitive Reasoning Patterns}
\label{supp:modeling_cognitive_reasoning_patterns}

Building upon the preliminary dataset described in \cref{supp:dataset_curation}, which contains only canonical \texttt{crop} tool calls derived from CRU bounding boxes, we apply a rule-based post-processing pipeline to instantiate four cognitive reasoning patterns Planning, Reflecting, Verifying, and Backtracking.
These transformations operationalize human-like problem-solving behaviors as explicit edits to the initial tool invocations.

\noindent{\textbf{Planning.}}
At the start of each reasoning path, we generate a global image description and high-level solution rationale from the $f_+$-scaled image to establish strategic context before CRU-level reasoning begins.

\noindent{\textbf{Reflecting.}}
When a CRU's region of interest is fully contained within the output of a preceding \texttt{crop} call from the same source path (i.e., within $p_0$, $p_1$, or $p_2$), we revise its tool invocation to reference that prior output image.
The \texttt{image\_index} is updated accordingly, and the bounding box is adjusted to preserve the physical region, enabling iterative visual focusing.

\noindent{\textbf{Verifying.}}
At the first erroneous step $\text{err}_2$ in path $p_2$, we replace the original \texttt{crop} call with a \texttt{display} invocation that explicitly references the original image, prompting the model to re-examine initial visual evidence and validate the current proposition.
To maintain continuity of this reasoning pattern, all subsequent tool calls in the path operate on the image output produced by this \texttt{display} call, thereby grounding later steps in the verified visual context.

\noindent{\textbf{Backtracking.}}
At the first erroneous step $\text{err}_1$ in path $p_1$, we insert a \texttt{scale} call applied to the original image with factor $f_+ / f_-$ to correct misjudgments caused by reasoning under the low-accuracy scale $f_-$.
To support this mechanism, the visual input for the entire $p_1$ path is rendered at scale $f_-$ during initial sampling, ensuring that upscaling by $f_+ / f_-$ recovers the high-accuracy view used in $p_0$.


All components, including planning statements, CRU sequences, structured tool calls in JSON format, guiding questions, and the final answer, are assembled into a complete cognitively grounded reasoning path following the compositional structure illustrated in
the main paper Fig.~2~(d).

\section{Additional Implementation Details}
\label{supp:additional_implementation_details}

All training experiments are conducted using established open-source frameworks to ensure full reproducibility.
We adopt Qwen2.5-VL-7B~\cite{qwen25VL} and Qwen2.5-VL-3B~\cite{qwen25VL} as the base MLLMs and obtain \textsc{\textbf{ViRC}}-7B and \textsc{\textbf{ViRC}}-3B through the three-stage pipeline of Instructional SFT, Practice SFT, and Strategic RL.
Both model variants use identical training configurations and hyperparameters.
During training, we freeze the vision encoder and the vision-language projector.

\subsection{Instructional SFT}
\label{supp:instructional_sft}

Instructional SFT is conducted on the 50K text-only subset of CRUX, matching the scale used in MINT-CoT~\cite{chen2025mint}.
Tool outputs are removed, and only textual supervision is preserved.
During this stage, we set the system prompt to the template shown in \cref{fig:supp_prompt_system}.
Training is performed using LLaMA Factory~\cite{zheng2024llamafactory} with a learning rate of 5.0e-6, 3 epochs, cosine scheduling, a warmup ratio of 0.03, and a batch size of~64.

\subsection{Practice SFT}
\label{supp:practice_sft}

Practice SFT uses the same 50K samples, but all tool calls are executed offline and the returned images are cached.
These images are inserted into the sequence immediately after each tool call.
We again employ the system prompt in \cref{fig:supp_prompt_system}, but explicitly remove the “\#~Emergency” section and its associated content to ensure that the model fully leverages visual cues during training.
This stage is trained with LLaMA Factory using a learning rate of 1.0e-6, 6 epochs, cosine scheduling, a warmup ratio of 0.1, and a batch size of~64.

\subsection{Strategic RL}
\label{supp:strategic_rl}

For the Strategic RL stage, we use the same system prompt as in Practice SFT and adopt Group Relative Policy Optimization (GRPO)~\cite{shao2024deepseekmath} as implemented in verl~\cite{sheng2024verlhybridflow}.
Unlike standard RL settings where the full question is provided as input, each policy query consists of a truncated reasoning fragment $q$ drawn from a curated hard subset $\mathcal{HD}$.
Each fragment represents an incomplete reasoning prefix from which the model must either reach the final answer or continue tool-guided reasoning.

Given a fragment $q$, the policy produces a group of $G$ candidate continuations $\{o_i\}_{i=1}^{G}$.
Instead of manually defining the reward here, we directly use the reward value $r_i = r(s_t, o_i)$ as specified in
the main paper Sec.~3.4,
where $r(\cdot)$ integrates answer correctness, multimodal coherence, reasoning-pattern alignment, and format validity penalty into a single composite score.
Advantages $A_i$ are computed by normalizing each reward within the group.
In our implementation, we do not include KL or entropy regularization and the policy is updated using the GRPO objective:

\begin{equation}
  \resizebox{0.9\columnwidth}{!}{$
    \begin{split}
    &\mathcal{J}_{GRPO}(\theta) = \mathbb{E}_{[q \sim \mathcal{HD}, \{o_i\}_{i=1}^G \sim \pi_{\theta_{old}}(\cdot|q)]} \frac{1}{G} \sum_{i=1}^{G} \Bigg( \\
    &\min \bigg( \frac{\pi_\theta(o_i|q)}{\pi_{\theta_{old}}(o_i|q)} A_i,\text{clip}\left( \frac{\pi_\theta(o_i|q)}{\pi_{\theta_{old}}(o_i|q)}, 1 - \epsilon, 1 + \epsilon \right) A_i \bigg) \Bigg)
    \end{split}
    \label{eq:supp_GRPO}
  $}
\end{equation}

\begin{equation}
A_i = \frac{r_i - \text{mean}(\{r_1, r_2, ..., r_G\})}{\text{std}(\{r_1, r_2, ..., r_G\})}
\label{eq:supp_advantage}
\end{equation}

We set $\epsilon = 0.2$.
The global batch size is 512 with a mini-batch size of 64.
A constant learning rate of $1\times10^{-6}$ is used.
To maintain computational efficiency, the maximum response length is limited to 512 tokens, and asynchronous rollouts are enabled throughout training.

\subsection{Hard Subset Curation}
\label{supp:hard_subset_curation}

The RL hard subset contains 8K truncated samples selected from CRUX and is designed to balance two abilities: producing final answers when sufficient context is available and continuing tool-guided reasoning when only partial evidence is provided.
The subset is composed of two 4K groups, corresponding to the truncated sample types illustrated in
Fig.3~(a) Stage~3.
The first group, \textit{Long Reasoning Truncation}, contains the 4K samples with the longest reasoning paths, measured by the total number of textual and visual tokens.
For each sample, the final CRU is removed while retaining the ground-truth answer.
These fragments tax the model's long-range information integration capabilities, requiring it to synthesize a final answer from highly complex, accumulated cognitive contexts.

The second group, \textit{Critical Region Truncation}, is constructed from samples not included in the first group.
For each such sample, we identify the tool-returned region that has the smallest area ratio (below 20\%) relative to the original image and belongs to the latest possible CRU.
This CRU and all subsequent steps are truncated while keeping the tool type of that CRU and the final answer as ground truth.
Samples are grouped by tool type, and within each group we retain the top $\frac{4}{3}$K samples with the longest reasoning fragments.
The three tool-specific groups jointly form the 4K Critical Region Truncation set.
These samples impose acute perceptual and strategic challenges, requiring the model to recognize informational deficits from obscure visual cues and autonomously select the correct investigative tool to proceed.

\subsection{Reward Design}
\label{supp:reward_design}
Reward scores for $r_{\text{ans}}$ and $r_{\text{mm}}$ are computed using Qwen2.5-VL-72B-Instruct~\cite{qwen25VL}.
The prompt for $r_{\text{ans}}$ is shown in \cref{fig:supp_prompt_reward_ans}.
The multimodal reward $r_{\text{mm}}$ includes a textual-coherence component $r_{\text{mm}}^{\text{text}}$ (prompt in \cref{fig:supp_prompt_reward_txt}) and a visual-relevance component $r_{\text{mm}}^{\text{vis}}$ (prompt in \cref{fig:supp_prompt_reward_vis}).
All returned scores are strictly required to lie in $[0,1]$.
Otherwise the query is repeated.
We assign weights of 0.4 and 0.5 to $r_{\text{mm}}^{\text{text}}$ and $r_{\text{mm}}^{\text{vis}}$, respectively.
Since $r_{\text{pattern}}$ yields either 0 or 0.1, we constrain the sum of the maxima of all multimodal reward components plus the maximum value of $r_{\text{pattern}}$ to equal 1, matching the maximum value of $r_{\text{ans}}$ and preventing potential reward hacking.

\section{Evaluation Protocol}
\label{supp:evaluation_protocol}

The main paper omits detailed explanations of our evaluation setup due to limited space.
Here, we clarify how we handle models that support tool invocation or integrate visual operations during reasoning.
For fairness, whenever a model provides an official or built-in mechanism for zoom-in, crop, or other visual-grounding tools, we enable these features during evaluation.
Specifically, Qwen3-VL-8B-Instruct~\cite{qwen3vl}, Pixel Reasoner~\cite{su2025pixelreasoner}, DeepEyes~\cite{zheng2025deepeyes}, and Mini-o3~\cite{lai2025minio3} are evaluated with their respective zoom-in or visual tools activated.
MINT-CoT-7B~\cite{chen2025mint} is tested with its interleaved visual tokens mechanism enabled, following its official implementation.
This unified setting ensures that all baselines fully exploit their intended visual reasoning capabilities for a fair comparison.

In addition, we use Qwen2.5-72B-Instruct~\cite{qwen25} to verify the correctness of the answers generated by each model.
The validation prompt is provided in \cref{fig:supp_prompt_sampling_val_ans}.

\section{Additional Results}
\label{supp:additional_results}

This section provides supplementary results to complement and substantiate the findings presented in the main paper. 
We begin by presenting detailed quantitative results and controlled studies in \cref{supp:additional_quantitative_results}.
Subsequently, we provide the full, fine-grained evaluation data from our ablation studies to offer a clearer view of each component's impact in \cref{supp:detailed_ablation_study_results}. 
Finally, we include additional qualitative results that visually demonstrate our model's reasoning capabilities on various problems in \cref{supp:additional_qualitative_results}.

\begin{table*}[t]
  \caption{
  \textbf{Fine-grained generalization results.}
  Our \textsc{\textbf{ViRC}} is trained on the CRUX dataset and cross-domain tested on VisualProbe \cite{lai2025minio3}, $V^\ast$~\cite{wu2024vstar}, and HR-Bench \cite{wang2025hrbench}. Avg. reports the mean of the results on the three benchmarks. Results marked with $\ast$ are from \cite{lai2025minio3}, while all other results are reproduced in this work.
  }
  \label{tab:supp_generalization}
  \centering
  \resizebox{\linewidth}{!}{
    \begin{NiceTabular}{w{l}{16em}|w{c}{4em}|w{c}{4em}|*{3}{w{c}{3.8em}}|w{c}{4em}|*{2}{w{c}{3.8em}}}
    \toprule
    &  &  & \multicolumn{3}{c}{VisualProbe} &  & \multicolumn{2}{c}{HR-Bench} \\
    \multirow{-2}{*}{Model} & \multirow{-2}{*}{\#Params} & \multirow{-2}{*}{Avg.} & easy & medium & hard & \multirow{-2}{*}{$V^\ast$} & 4K & 8K \\
    \midrule
    \multicolumn{9}{c}{\textit{Closed-Source   Model}} \\
    \midrule
    GPT-4o   \cite{openai2024gpt4ocard} & -- & 43.27 & 47.50\rlap{$^{\ast}$} & 15.40\rlap{$^{\ast}$} & 11.20\rlap{$^{\ast}$} & 65.20\rlap{$^{\ast}$} & 62.00\rlap{$^{\ast}$} & 58.30\rlap{$^{\ast}$} \\
    \midrule
    \multicolumn{9}{c}{\textit{Open-Source   General Model}} \\
    \midrule
    Qwen3-VL-8B-Instruct$_{w/\,tool}$ \cite{qwen3vl} & 8B  & 33.47 & 36.68 & 10.82 & 21.70 & 41.88 & 44.12 & 45.62 \\
    LLaVA-OV-Qwen2-7b-ov   \cite{li2024llavaov} & 7B & 41.37 & 36.20\rlap{$^{\ast}$} & 12.50\rlap{$^{\ast}$} & 13.40\rlap{$^{\ast}$} & 70.90\rlap{$^{\ast}$} & 61.20\rlap{$^{\ast}$} & 54.00\rlap{$^{\ast}$} \\
    InternVL2.5-8B   \cite{chen2025intervl25} & 8B & 44.03 & {\ul 55.32} & 11.57 & 11.32 & 68.59 & 62.75 & 54.62 \\
    InternVL2.5-8B-MPO   \cite{wang202intervl2_5mpo} & 8B & 42.45 & 41.13 & 13.43 & 8.49 & 72.25 & 64.12 & 55.25 \\
    DeepSeek-VL2   \cite{wu2024deepseekvl2} & 4.5B & 38.43 & 32.62 & 11.19 & 7.55 & 66.49 & 61.12 & 51.62 \\
    \midrule
    \multicolumn{9}{c}{\textit{Open-Source   Reasoning Model}} \\
    \midrule
    Hint-GRPO-Qwen2.5-VL-3B   \cite{huang2025hintgrpo} & 3B & 45.81 & 46.10 & 20.90 & 17.92 & 68.06 & 62.62 & 59.25 \\
    Hint-GRPO-Qwen2-VL-7B   \cite{huang2025hintgrpo} & 7B & 47.12 & 50.35 & 17.16 & 16.04 & {\ul 75.92} & 65.50 & 57.75 \\
    R1-VL-7B \cite{zhang2025r1vl} & 7B & 34.51 & 39.72 & 5.97 & 15.09 & 45.55 & 54.37 & 46.38 \\
    MM-Eureka-Qwen-7B   \cite{meng2025mmeureka} & 7B & 32.87 & 26.24 & 4.10 & 4.72 & 59.16 & 58.00 & 45.00 \\
    MINT-CoT-7B   \cite{chen2025mint} & 7B & 32.04 & 34.75 & 11.32 & 7.46 & 43.46 & 51.25 & 44.00 \\
    \midrule
    Qwen2.5-VL-3B-Instruct   \cite{qwen25VL} & 3B & 46.64 & 36.17 & 22.01 & 23.58 & 68.06 & 67.25 & 62.75 \\
    \rowcolor[HTML]{F2F2F2}
    \textsc{\textbf{ViRC}}\textbf{-3B (Ours)} & 3B & {\ul 51.10} & 49.65 & 25.75 & {\ul 25.47} & 72.25 & {\ul 70.00} & {\ul 63.50} \\
    \midrule
    Qwen2.5-VL-7B-Instruct   \cite{qwen25VL} & 7B & 49.25 & 39.10\rlap{$^{\ast}$} & {\ul 26.00\rlap{$^{\ast}$}} & 23.90\rlap{$^{\ast}$} & 75.50\rlap{$^{\ast}$} & 68.25\rlap{$^{\ast}$} & 62.75\rlap{$^{\ast}$} \\
    \rowcolor[HTML]{F2F2F2}
    \textsc{\textbf{ViRC}}\textbf{-7B (Ours)} & 7B & \textbf{58.27} & \textbf{56.03} & \textbf{36.94} & \textbf{37.74} & \textbf{79.06} & \textbf{73.12} & \textbf{66.75} \\ \bottomrule
  \end{NiceTabular}
  }
\end{table*}
\begin{table*}[tp]
  \caption{
  \textbf{Robustness to pretrained model.}
  We apply our \textsc{\textbf{ViRC}} training pipeline to two different pretrained MLLMs, Qwen2-VL-2B-Instruct and Qwen2-VL-7B-Instruct, yielding \textsc{\textbf{ViRC}}-Qwen2VL-2B and \textsc{\textbf{ViRC}}-Qwen2VL-7B, respectively.
  }
  \label{tab:supp_robustness}
  \centering
  \resizebox{\linewidth}{!}{
    \begin{NiceTabular}{w{l}{13em}|w{c}{5em}|w{c}{5em}|w{c}{5em}|w{c}{5em}|*{5}{w{c}{3em}}}
    \toprule
    &  &  &  & MMStar & \multicolumn{5}{c}{MathVista-Math} \\
    \multirow{-2}{*}{Model} & \multirow{-2}{*}{\#Params} & \multirow{-2}{*}{Avg.} & \multirow{-2}{*}{GeoQA} & Math & ALL & GEO & ALG & GPS & TQA  \\
    \midrule
    Qwen2-VL-2B-Instruct \cite{qwen2vl} & 2B & 25.23 & 15.52 & 32.40 & 27.78 & 24.06 & 27.72 & 24.52 & 38.71 \\
    \rowcolor[HTML]{F2F2F2}
    \textsc{\textbf{ViRC}}\textbf{-Qwen2VL-2B (Ours)} & 2B & 44.20 & 44.16 & 44.00 & 44.44 & 42.92 & 44.94 & 43.27 & 48.39 \\
    \midrule
    Qwen2-VL-7B-Instruct \cite{qwen2vl} & 7B & 44.49 & 30.37 & 52.00 & 51.11 & 46.70 & 50.94 & 47.60 & 62.90 \\
    \rowcolor[HTML]{F2F2F2}
    \textsc{\textbf{ViRC}}\textbf{-Qwen2VL-7B (Ours)} & 7B & 64.32 & 72.68 & 58.80 & 61.48 & 58.02 & 61.42 & 58.17 & 72.58 \\
    \bottomrule
  \end{NiceTabular}
  }
\end{table*}
\begin{table*}[tp]
  \caption{
  \textbf{Benefits beyond the teacher model.} 
  We compare Qwen2.5-VL-72B-Instruct, the teacher model, with \textsc{\textbf{ViRC}}-7B trained on the CRUX dataset.
  }
  \label{tab:supp_distill_or_cru}
  \centering
  \resizebox{\linewidth}{!}{
    \begin{NiceTabular}{w{l}{13em}|w{c}{5em}|w{c}{5em}|w{c}{5em}|w{c}{5em}|*{5}{w{c}{3em}}}
    \toprule
    &  &  &  & MMStar & \multicolumn{5}{c}{MathVista-Math} \\
    \multirow{-2}{*}{Model} & \multirow{-2}{*}{\#Params} & \multirow{-2}{*}{Avg.} & \multirow{-2}{*}{GeoQA} & Math & ALL & GEO & ALG & GPS & TQA  \\
    \midrule
    Qwen2.5-VL-72B-Instruct \cite{qwen25VL} & 72B & \uline{75.12} & \uline{66.45} & \textbf{80.40} & \uline{78.52} & \uline{77.83} & \uline{78.28} & \uline{77.40} & \textbf{82.26} \\
    \rowcolor[HTML]{F2F2F2}
    \textsc{\textbf{ViRC}}\textbf{-7B (Ours)} & 7B & \textbf{77.79} & \textbf{75.07} & \uline{77.20} & \textbf{81.11} & \textbf{81.13} & \textbf{81.27} & \textbf{81.73} & \uline{79.03} \\
    \bottomrule
  \end{NiceTabular}
  }
\end{table*}

\begin{table*}[tp]
    \caption{
    \textbf{Additional benchmark results.}
    We further compare MINT-CoT-7B, Qwen2.5-VL-7B-Instruct, and \textsc{\textbf{ViRC}}-7B on broader multimodal math benchmarks, including MathVision~\cite{wang2024mathvision}, We-Math~\cite{qiao2025wemath}, MathVerse~\cite{zhang2024mathverse}, and LogicVista~\cite{xiao2024logicvista}.
    }
  \label{tab:supp_more_result_only_math}
  \centering
  \resizebox{\linewidth}{!}{
    \begin{NiceTabular}{w{l}{13em}|*{5}{w{c}{8em}}}
    \toprule
    Model & Avg. & MathVision & We-Math & MathVerse & LogicVista \\
    \midrule
    MINT-CoT-7B \cite{chen2025mint} & 38.54 & 22.52 & \uline{52.00} & \uline{51.56} & 26.12 \\
    Qwen2.5-VL-7B-Instruct \cite{qwen25VL} & \uline{40.07} & \uline{25.13} & 42.67 & 49.19 & \uline{43.30} \\
    \rowcolor[HTML]{F2F2F2}
    \textsc{\textbf{ViRC}}\textbf{-7B (Ours)} & \textbf{48.83} & \textbf{38.64} & \textbf{59.81} & \textbf{52.01} & \textbf{44.87} \\
    \bottomrule
  \end{NiceTabular}
  }
\end{table*}
\begin{table*}[tp]
  \caption{
  \textbf{Inference-time cost.}
  We report token usage and inference latency for different models and variants, including both average statistics and worst-case settings reported for the top 1\% longest token sequences and the slowest 1\% inference cases.
  }
  \label{tab:supp_infer_cost}
  \centering
  \resizebox{\linewidth}{!}{
    \begin{NiceTabular}{w{l}{13em}|w{c}{7em}|*{4}{w{c}{5em}}|*{2}{w{c}{5em}}}
    \toprule
    \multirow{2}{*}{Model} & \multirow{2}{*}{CoT Type} & \multicolumn{4}{c}{Avg. \#Tokens} & \multicolumn{2}{c}{Avg. Latency (s)} \\
     &  & All & Text & Visual & Top 1\% Len. & All & Top 1\% Lat. \\
    \midrule
    MINT-CoT-7B \cite{chen2025mint} & visual & 1540 & 792 & 748 & \textbf{15275} & 9.38 & 19.22 \\
    MM-Eureka-Qwen-7B \cite{meng2025mmeureka} & text-only & {\ul 945} & {\ul 384} & \textbf{561} & 31004 & {\ul 4.09} & \textbf{10.80} \\
    \textsc{\textbf{ViRC}}\textbf{-7B} $w/o$ CRU & visual & 1415 & 522 & 893 & 29473 & 5.26 & 11.98 \\
    \textsc{\textbf{ViRC}}\textbf{-7B (Ours)} & reason chunking & \textbf{879} & \textbf{183} & {\ul 696} & {\ul 24144} & \textbf{2.31} & {\ul 11.00} \\
    \bottomrule
  \end{NiceTabular}
  }
\end{table*}

\subsection{Additional Quantitative Results}
\label{supp:additional_quantitative_results}

\noindent\textbf{Generalization under domain shift.}
The main paper reports averaged scores on generalization benchmarks.
Here we provide more detailed results on VisualProbe~\cite{lai2025minio3}, $V^\ast$~\cite{wu2024vstar}, and HR-Bench~\cite{wang2025hrbench} in \cref{tab:supp_generalization}.
Although \textsc{\textbf{ViRC}} is trained only on CRUX, which consists of low-resolution schematic mathematical diagrams with CRU-align reasoning and tool calls, it generalizes well to high-resolution natural images (2K--16K) and perception-centric questions without any domain-specific fine-tuning.
These results suggest that the learned reason chunking and selective tool usage transfer beyond the training distribution.

\noindent\textbf{Robustness to pretrained models.}
\cref{tab:supp_robustness} shows that \textsc{\textbf{ViRC}} yields consistent improvements when instantiated with different pretrained model families.
In particular, the gains hold when switching from Qwen2-VL, which uses a fixed-resolution strategy, to Qwen2.5-VL, which adopts a dynamic-resolution strategy, indicating that our method is robust to the pretrained model choice and its image-resolution policy.

\noindent\textbf{Benefits beyond distillation from the teacher model.}
Results in \cref{tab:supp_distill_or_cru} show that \textsc{\textbf{ViRC}}-7B can even surpass the teacher model on multiple benchmarks.
This suggests that the reason chunking mechanism provides additional benefits over the base model by introducing chunk-level organization and denser training signals.

\noindent\textbf{Broader multimodal math benchmarks.}
\cref{tab:supp_more_result_only_math} reports additional results on broader multimodal math benchmarks, including MathVision~\cite{wang2024mathvision}, We-Math~\cite{qiao2025wemath}, MathVerse~\cite{zhang2024mathverse}, and LogicVista~\cite{xiao2024logicvista}.
For a fair comparison against a method that also uses multimodal CoT, we include MINT-CoT~\cite{chen2025mint} as a strong baseline.
Notably, MINT-CoT is within 1\% of the best model compared in the main paper, MM-Eureka-Qwen-7B~\cite{meng2025mmeureka}, while \textsc{\textbf{ViRC}}-7B consistently outperforms MINT-CoT across these benchmarks.

\noindent\textbf{Inference-time cost.}
\cref{tab:supp_infer_cost} reports token usage and inference latency measured with vLLM~\cite{vllm} v0.14.1 on a single A100 80G GPU (batch size 1, FP16) across all benchmarks reported in \cref{tab:supp_more_result_only_math,tab:supp_distill_or_cru}.
We report both average statistics and worst-case settings based on the top 1\% longest token sequences and the slowest 1\% inference cases.
\textsc{\textbf{ViRC}}-7B achieves the lowest average token usage and inference latency while maintaining strong performance.
Removing CRU increases both token consumption and latency, highlighting the efficiency benefit of reason chunking.
Under worst-case settings, \textsc{\textbf{ViRC}}-7B achieves the second-best results, remaining competitive in inference-time cost.

\begin{table*}[t]
  \caption{
  \textbf{Fine-grained results for CRU effectiveness analysis.} Performance is broken down by subset and category.
  }
  \vspace{-1em}
  \label{tab:supp_cru_ablation}
  \centering
  \resizebox{\linewidth}{!}{
  \begin{NiceTabular}{w{l}{8em}|w{c}{3em}|w{c}{3em}|*{5}{w{c}{2em}}|*{3}{w{c}{2em}}|w{c}{3em}|*{5}{w{c}{2em}}}
    \toprule
    &  & MMStar & \multicolumn{5}{c}{MathVista-Math} & \multicolumn{3}{c}{VisualProbe} &  & \multicolumn{2}{c}{HR-Bench} \\
    \multirow{-2}{*}{Model} & \multirow{-2}{*}{GeoQA} & Math & ALL & GEO & ALG & GPS & TQA & easy & medium & hard & \multirow{-2}{*}{$V^\ast$} & 4K & 8K \\
    \midrule
    Baseline & 43.50 & 66.80 & 66.67 & 65.56 & 66.29 & 65.87 & 69.35 & 39.10 & 26.00 & 23.90 & 75.50 & 68.25 & 62.75 \\
    Ours $w/o$ CRU & {\ul 74.67} & {\ul 72.80} & {\ul 79.26} & {\ul 78.77} & {\ul 79.40} & {\ul 79.33} & {\ul 79.03} & {\ul 47.52} & {\ul 29.10} & {\ul 34.91} & {\ul 75.92} & {\ul 71.00} & {\ul 63.63} \\
    Ours & \textbf{75.07} & \textbf{77.20} & \textbf{81.11} & \textbf{81.13} & \textbf{81.27} & \textbf{81.73} & \textbf{79.03} & \textbf{56.03} & \textbf{36.94} & \textbf{37.74} & \textbf{79.06} & \textbf{73.12} & \textbf{66.75} \\
    \bottomrule
  \end{NiceTabular}
  }
\end{table*}
\begin{table*}[t]
    \caption{
    \textbf{Fine-grained results for reasoning pattern ablation.} Each row shows performance when a specific reasoning pattern is removed from training and evaluation. $\mathcal{P}$, $\mathcal{B}$, $\mathcal{R}$, and $\mathcal{V}$ stand for 4 reasoning patterns--planning, backtracking, reflecting, and verifying, respectively.
    }
    \vspace{-1em}
    \label{tab:supp_reasoning_pattern_ablation}
    \centering
    \resizebox{\textwidth}{!}{
    \begin{NiceTabular}{*{4}{w{c}{1.1em}}|w{c}{3em}|w{c}{3em}|*{5}{w{c}{2em}}|*{3}{w{c}{2em}}|w{c}{3em}|*{2}{w{c}{2em}}}
    \toprule
    \multicolumn{4}{c|}{Reasoning Pattern} &  & MMStar & \multicolumn{5}{c|}{MathVista-Math} & \multicolumn{3}{c|}{VisualProbe} &  & \multicolumn{2}{c}{HR-Bench} \\
     $\mathcal{P}$ & $\mathcal{B}$ & $\mathcal{R}$ & $\mathcal{V}$ & \multirow{-2}{*}{GeoQA} & Math & ALL & GEO & ALG & GPS & TQA & easy & medium & hard & \multirow{-2}{*}{$V^\ast$} & 4K & 8K \\
        \midrule
        \quad & \quad & \quad & \quad & 43.50 & 66.80 & 66.66 & 65.56 & 66.29 & 65.87 & 69.35 & 39.10 & 25.75 & 23.90 & 71.20 & 68.25 & 62.75 \\
        \checkmark & \quad & \quad & \quad & 72.02 & {\ul 74.80} & {\ul 79.26} & {\ul 79.25} & {\ul 79.03} & 79.33 & \textbf{79.03} & 47.52 & 27.24 & 32.08 & 75.39 & 71.62 & 65.50 \\
        \checkmark & \checkmark & \quad & \quad & 72.81 & 68.40 & 78.52 & {\ul 79.25} & 78.65 & {\ul 79.81} & 74.19 & 50.35 & 27.61 & \textbf{37.74} & 78.01 & 71.75 & \textbf{66.75} \\
        \checkmark & \quad & \checkmark & \quad & 66.98 & 74.00 & 68.89 & 67.92 & 69.29 & 68.75 & 69.35 & 51.77 & 31.34 & {\ul 36.79} & {\ul 78.53} & 70.75 & \textbf{66.75} \\
        \checkmark & \quad & \quad & \checkmark & {\ul 74.93} & 72.40 & 77.41 & 77.83 & 77.53 & 77.88 & {\ul 75.81} & {\ul 53.90} & {\ul 32.46} & 34.91 & 78.01 & {\ul 72.50} & {\ul 65.88} \\
        \checkmark & \checkmark & \checkmark & \checkmark & \textbf{75.07} & \textbf{77.20} & \textbf{81.11} & \textbf{81.13} & \textbf{81.27} & \textbf{81.73} & \textbf{79.03} & \textbf{56.03} & \textbf{36.94} & \textbf{37.74} & \textbf{79.06} & \textbf{73.12} & \textbf{66.75} \\
        \bottomrule
    \end{NiceTabular}
    }
\end{table*}
\begin{table*}[th]
  \caption{
  \textbf{Fine-grained results for the impact of the three training stages.}  \textit{I}, \textit{P}, and \textit{S} stand for Instructional SFT, Practice SFT, and Strategic RL, respectively.
  }
  \vspace{-1em}
  \label{tab:supp_training_stage_ablation}
  \centering
  \resizebox{\linewidth}{!}{
    \begin{NiceTabular}{*{3}{w{c}{1.87em}}|w{c}{3em}|w{c}{3em}|*{5}{w{c}{2em}}|*{3}{w{c}{2em}}|w{c}{3em}|*{2}{w{c}{2em}}}
    \toprule
    \multicolumn{3}{c|}{Training Stage} &  & MMStar & \multicolumn{5}{c|}{MathVista-Math} & \multicolumn{3}{c|}{VisualProbe} &  & \multicolumn{2}{c}{HR-Bench} \\
    I & P & S & \multirow{-2}{*}{GeoQA} & Math & ALL & GEO & ALG & GPS & TQA & easy & medium & hard & \multirow{-2}{*}{$V^\ast$} & 4K & 8K \\
    \midrule
    & & & 43.50 & 66.80 & 66.67 & 65.56 & 66.29 & 65.87 & 69.35 & 39.10 & 26.00 & 23.90 & 75.50 & 68.25 & 62.75 \\
    \checkmark & & & 70.27 & 70.80 & 76.30 & 77.36 & 76.40 & 77.88 & {\ul 70.97} & {\ul 42.55} & 27.24 & {\ul 29.25} & 74.87 & {\ul 71.75} & {\ul 66.00} \\
    \checkmark & \checkmark & & {\ul 73.34} & {\ul 72.80} & {\ul 77.78} & {\ul 79.72} & {\ul 77.90} & {\ul 80.29} & 69.35 & {\ul 42.55} & {\ul 29.48} & 26.42 & {\ul 75.92} & 70.25 & 63.75 \\
    \checkmark & \checkmark & \checkmark & \textbf{75.07} & \textbf{77.20} & \textbf{81.11} & \textbf{81.13} & \textbf{81.27} & \textbf{81.73} & \textbf{79.03} & \textbf{56.03} & \textbf{36.94} & \textbf{37.74} & \textbf{79.06} & \textbf{73.12} & \textbf{66.75} \\
    \bottomrule
  \end{NiceTabular}
  }
\end{table*}
\begin{table*}[t]
  \caption{
  \textbf{Fine-grained results for RL component ablation.} The results show that each one of the designs in our proposed Strategic RL plays an important role in multimodal reasoning. For brevity, we denote the curated hard subset derived from CRUX as $\mathcal{HD}$.
  }
  \vspace{-1em}
  \label{tab:supp_rl_components_ablation}
  \centering
  \resizebox{\textwidth}{!}{
  \begin{NiceTabular}{w{l}{8em}|w{c}{3em}|w{c}{3em}|*{5}{w{c}{2em}}|*{3}{w{c}{2em}}|w{c}{3em}|*{2}{w{c}{2em}}}
    \toprule
    &  & MMStar & \multicolumn{5}{c}{MathVista-Math} & \multicolumn{3}{c}{VisualProbe} &  & \multicolumn{2}{c}{HR-Bench} \\
    \multirow{-2}{*}{Model} & \multirow{-2}{*}{GeoQA} & Math & ALL & GEO & ALG & GPS & TQA & easy & medium & hard & \multirow{-2}{*}{$V^\ast$} & 4K & 8K \\
    \midrule
    Ours & \textbf{75.07} & \textbf{77.20} & \textbf{81.11} & \textbf{81.13} & \textbf{81.27} & \textbf{81.73} & {\ul 79.03} & \textbf{56.03} & \textbf{36.94} & \textbf{37.74} & \textbf{79.06} & \textbf{73.12} & \textbf{66.75} \\
    $w/o$ $\mathcal{HD}$ & 66.84 & 70.80 & 73.70 & 75.47 & 73.78 & 76.44 & 64.52 & 45.39 & 28.73 & 32.08 & {\ul 76.44} & 71.75 & 65.75 \\
    $w/o$ $r_{\text{mm}}$ & {\ul 73.60} & 74.00 & 79.63 & {\ul 80.19} & 79.40 & {\ul 80.77} & 75.81 & {\ul 50.35} & 31.34 & 32.08 & \textbf{79.06} & {\ul 72.50} & 65.88 \\
    $w/o$ $r_{\text{pattern}}$ & 73.56 & {\ul 74.80} & {\ul 80.00} & 78.77 & {\ul 80.25} & 79.33 & \textbf{82.26} & 46.10 & {\ul 35.07} & {\ul 33.96} & {\ul 76.44} & 71.88 & {\ul 66.50} \\
    \bottomrule
  \end{NiceTabular}
  }
\end{table*}

\subsection{Detailed Ablation Study Results}
\label{supp:detailed_ablation_study_results}

This section provides additional details of the ablation configurations in
the main paper Sec.~4.3.
and report the full ablation study results.
Specifically, the main paper reports only averaged benchmark scores.
Here we additionally provide subset- and category-level metrics for all ablated variants.
Detailed results are shown in 
\cref{tab:supp_cru_ablation,tab:supp_reasoning_pattern_ablation,tab:supp_training_stage_ablation,tab:supp_rl_components_ablation}.

\noindent{\textbf{Ablation Setup.}}
To ensure fair and controlled comparisons, all ablation studies were conducted on the same base model, Qwen2.5-VL-7B-Instruct~\cite{qwen25VL}.
The training protocol for each ablation run was kept identical to the one used to train our final \textsc{ViRC}-7B model, whose results are reported in the state-of-the-art comparison
(Sec. 4.2).
Unless a component was intentionally modified as part of the study (e.g., skipping a training stage), all other conditions were held constant.
These conditions include all training hyperparameters, such as learning rate, batch size, and the number of epochs per stage.
This methodology isolates the impact of each ablated component, ensuring that performance differences can be attributed directly to its removal.

\noindent{\textbf{CRU Effectiveness Analysis.}}  
To evaluate the model’s performance without explicit CRU guidance, we ablated the Instructional SFT stage.
The model was trained directly with Practice SFT, followed by Strategic RL.
As described in
Sec.~3.4
of the main paper, the Instructional SFT stage is responsible for teaching the model the internal textual structure of each CRU.
Skipping this stage therefore simulates an environment where this structural prior is absent.

\noindent{\textbf{Reasoning Pattern Analysis.}}  
For each reasoning pattern under ablation, we follow the procedure in \cref{supp:modeling_cognitive_reasoning_patterns} to selectively exclude its modeling during dataset construction.
We then trained a separate model on each dataset variant using the full three-stage training strategy.
Comparing the performance of these models allowed us to isolate the impact of each pattern.

\noindent{\textbf{RL Component Analysis.}}  
We ablate two components in the Strategic RL stage: the curated hard subset and the reward design.  
For hard subset ablation, we replace the curated 8K examples with a randomly sampled 8K subset from CRUX.
These samples were created by taking full reasoning paths and truncating them at an intermediate CRU.
The ground truth for each resulting fragment consisted of the tool call at the truncation point and the final answer from the original path.
For reward ablation, we consider two settings.
In $w/o\ r_{\text{mm}}$, only $r_{\text{pattern}}$ was used for tool invocation.
In $w/o\ r_{\text{pattern}}$, only $r_{\text{mm}}$ was used.

\onecolumn
\subsection{\texorpdfstring{\raggedright Additional Qualitative Results}{Additional Qualitative Results}}
\label{supp:additional_qualitative_results}
\centering
\toolsection
{
Qualitative Comparison on the GeoQA Benchmark
}
{
\toolitem{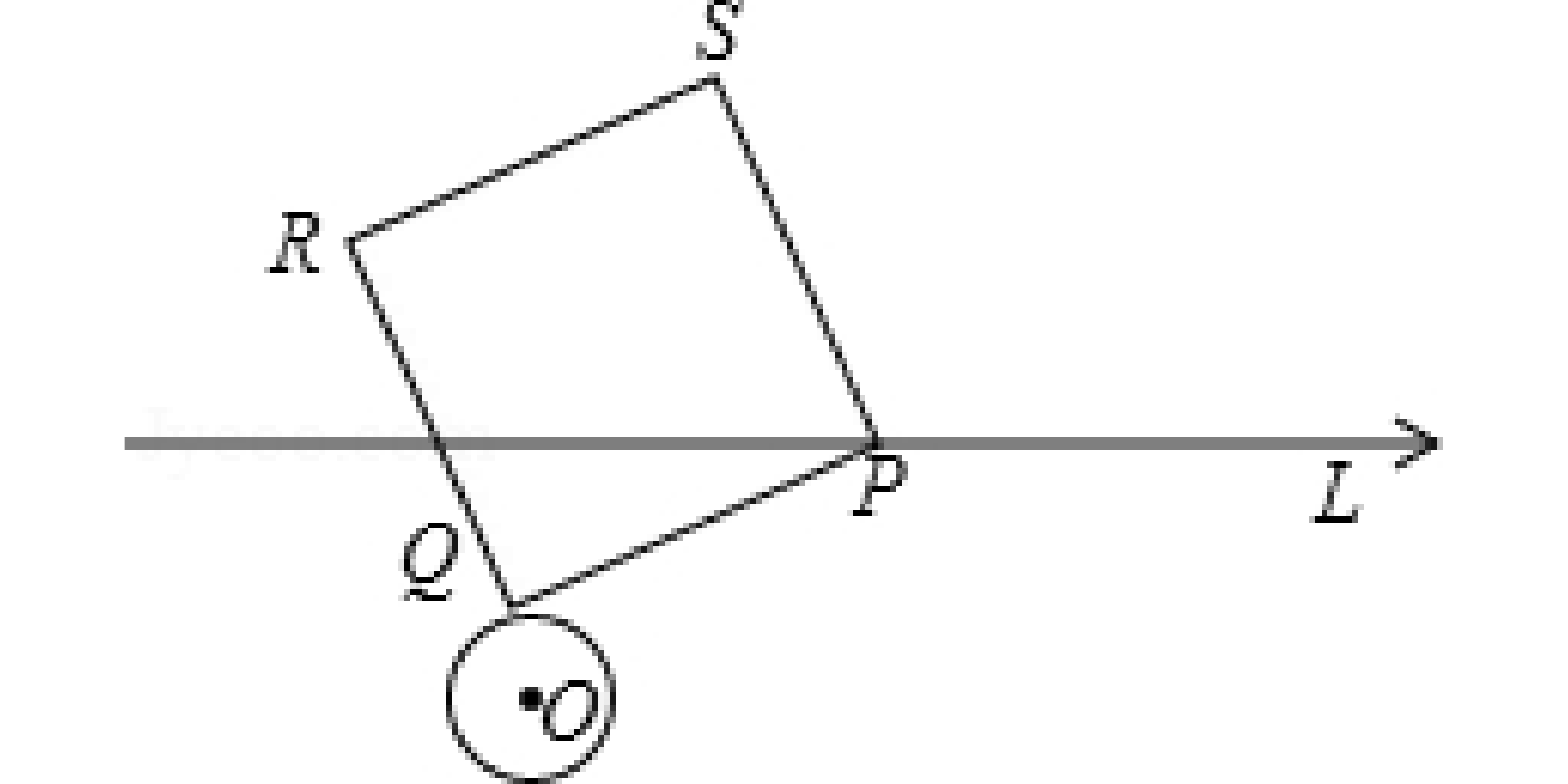}{
circle O is a circle with a radius of 1.0, the distance from point O to line L is 3.0, draw a tangent of circle O through any point P on the straight line L , and the tangent point is Q; if PQ is taken as the edge to make the square PQRS, then the minimum area of the square PQRS is ()
}{Initial Query}
\toolitem{assets/figure/example_math_1/0.pdf}{
\textless think\textgreater\textbackslash nLet's think step by step. A circle with center O and radius 1 is positioned such that its distance to a line L is 3; from any point P on L, a tangent is drawn to the circle touching at Q, and a square PQRS is constructed on the tangent segment PQ. Apply the Pythagorean theorem to the right triangle OPQ, where O is the circle's center, P is on line L, and Q is the point of tangency, to express the length of PQ in terms of the fixed distance OP and the known radius OQ. How can I choose a coordinate system that simplifies the given geometric setup?\textbackslash n\textless \textbackslash think\textgreater\textbackslash n\textbackslash n\textless tool\_call\textgreater\textbackslash n\{"name": "crop\_image", "arguments": \{"bbox\_2d": [0, 0, 224, 139], "image\_index": 0\}\}\textbackslash n\textless/tool\_call\textgreater
}{\textsc{\textbf{ViRC}}-7B Turn 1: Planning}
\toolitem{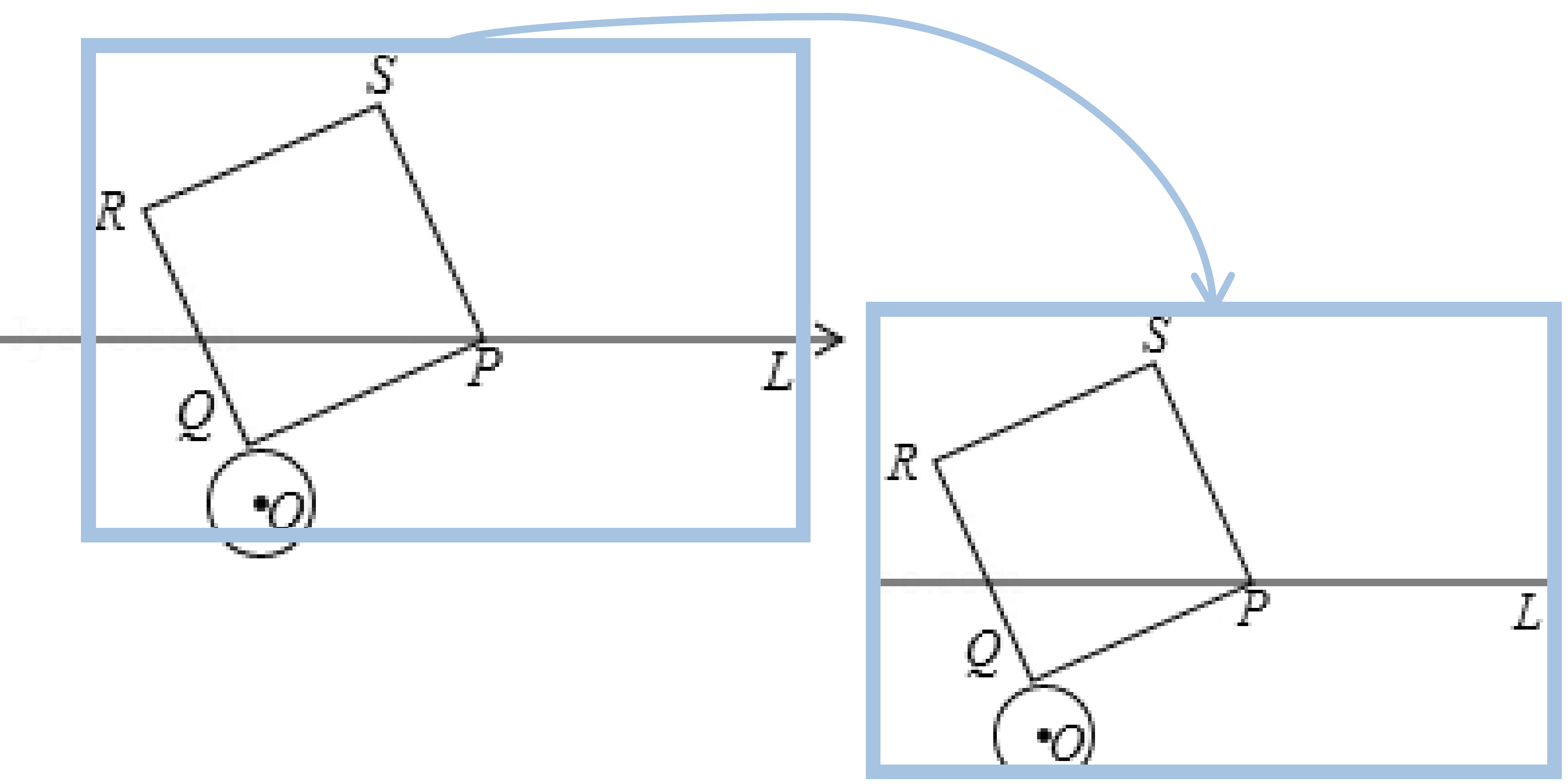}{
\textless think\textgreater\textbackslash nI set up a coordinate system with line L as the x-axis (y=0) and the circle's center O at (0, 3). This aligns with the problem's description where the distance from O to L is 3 units, simplifying calculations involving distances and coordinates. How can I use coordinates to represent a general point P on line L and relate it to the circle's center and radius?\textbackslash n\textless \textbackslash think\textgreater\textbackslash n\textbackslash n\textless tool\_call\textgreater\textbackslash n\{"name": "crop\_image", "arguments": \{"bbox\_2d": [35, 64, 139, 139], "image\_index": 1\}\}\textbackslash n\textless/tool\_call\textgreater
}{\textsc{\textbf{ViRC}}-7B Turn 2: Reflecting}
\toolitem{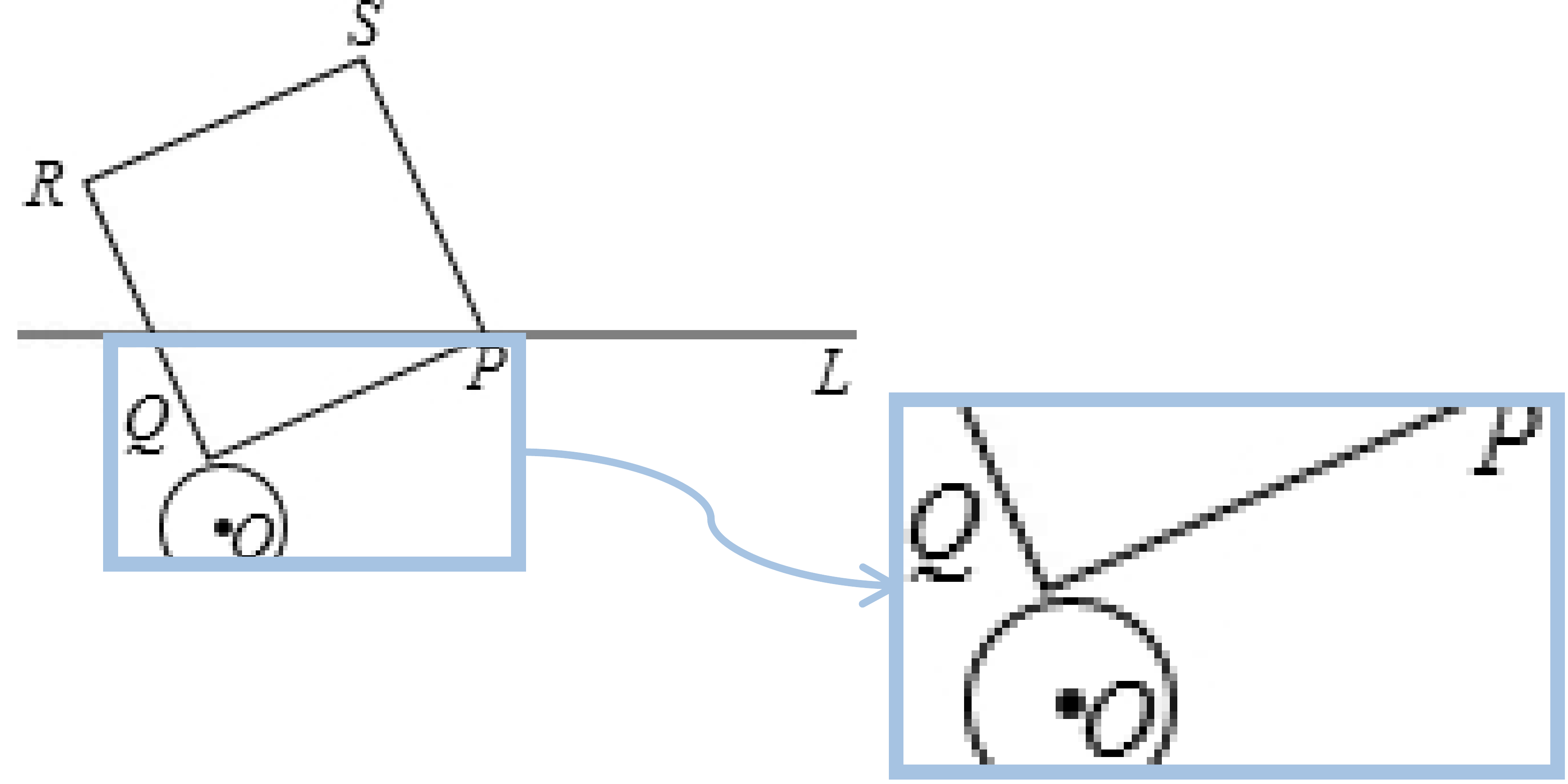}{
\textless think\textgreater\textbackslash nI let point P be any point on line L with coordinates (x, 0). Drawing a tangent from P to the circle, the point of contact is Q. Since the radius OQ is perpendicular to the tangent PQ, triangle OPQ is a right-angled triangle at Q. What relationship can I use in right triangle OPQ to connect the sides OP, OQ, and PQ?\textbackslash n\textless \textbackslash think\textgreater\textbackslash n\textbackslash n\textless tool\_call\textgreater\textbackslash n\{"name": "crop\_image", "arguments": \{"bbox\_2d": [35, 0, 139, 139], "image\_index": 1\}\}\textbackslash n\textless/tool\_call\textgreater
}{\textsc{\textbf{ViRC}}-7B Turn 3}
\toolitem{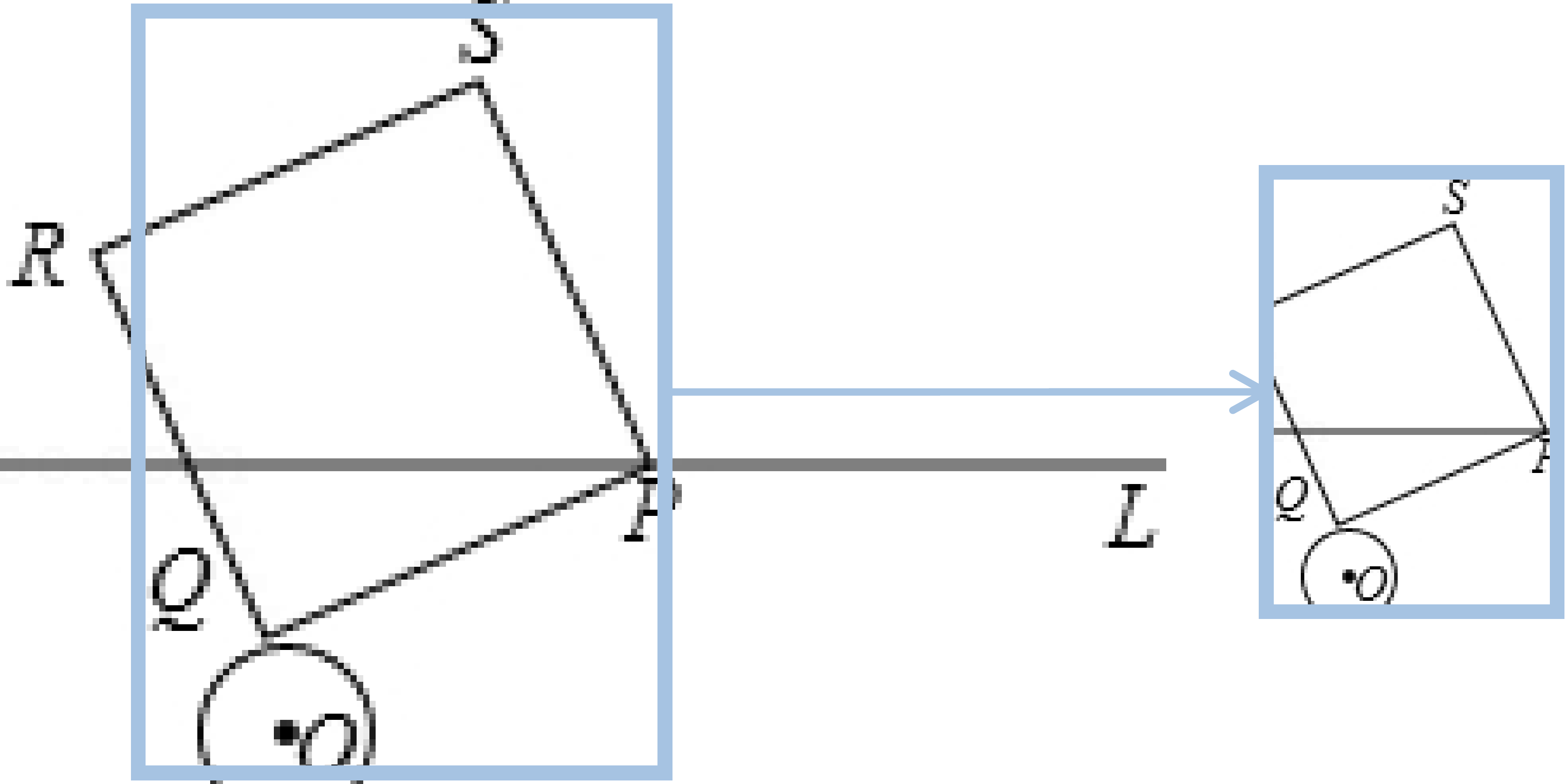}{
\textless think\textgreater\textbackslash nUsing the Pythagorean theorem in triangle OPQ: \textbackslash ( PQ\^{}2 = OP\^{}2 - OQ\^{}2 \textbackslash ). Here, \textbackslash ( OQ = 1 \textbackslash ) (radius), so substituting gives \textbackslash ( PQ\^{}2 = x\^{}2 + 3\^{}2 - 1\^{}2 = x\^{}2 + 9 - 1 = x\^{}2 + 8 \textbackslash ). Since PQ is a side of the square, the area of the square PQRS is \textbackslash ( (PQ)\^{}2 = x\^{}2 + 8 \textbackslash ). To minimize the area, I need to minimize \textbackslash ( x\^{}2 \textbackslash ), which occurs when \textbackslash ( x = 0 \textbackslash ). When x=0, how does the expression for the square’s area simplify, and what does that tell me about the minimum area?\textbackslash n\textless \textbackslash think\textgreater\textbackslash n\textbackslash n\textless tool\_call\textgreater\textbackslash n\{"name": "crop\_image", "arguments": \{"bbox\_2d": [0, 0, 104, 106], "image\_index": 1\}\}\textbackslash n\textless/tool\_call\textgreater
}{\textsc{\textbf{ViRC}}-7B Turn 4}

\begin{tcolorbox}[
width=\linewidth,
colback=box_section_back,
arc=0mm,
boxrule=0pt,
left=4pt,
right=4pt,
breakable,
halign=center,
valign=center,
height=6em,
]
(continued on next page)
\end{tcolorbox}

\toolitem{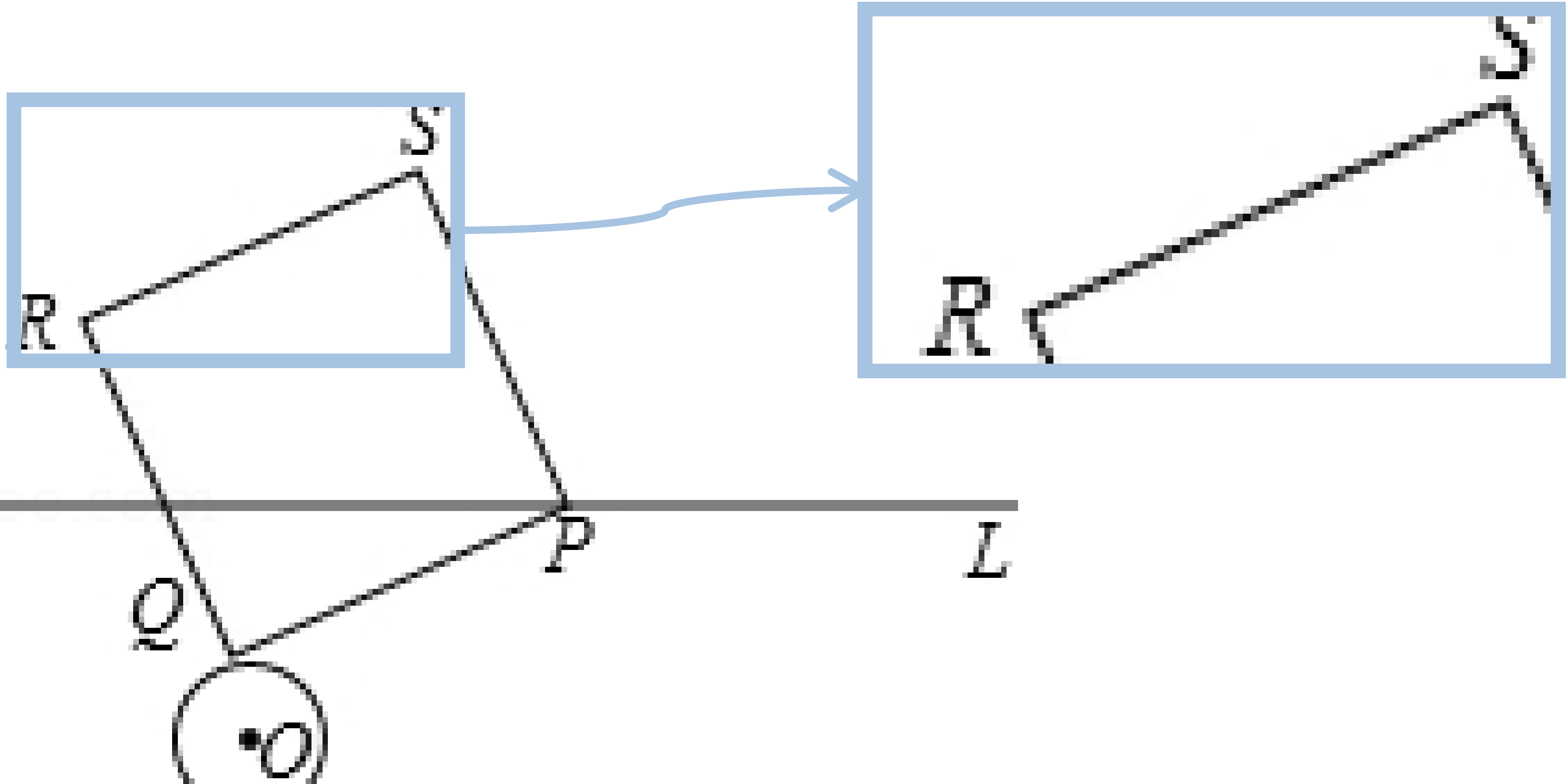}{
\textless think\textgreater\textbackslash nSubstituting \textbackslash ( x = 0 \textbackslash ) into the equation, the minimal area becomes \textbackslash ( 0\^{}2 + 8 = 8 \textbackslash ). Thus, the smallest possible area of square PQRS is achieved when point P lies directly below or above the center O on line L. Why does the case where \$x = -\textbackslash sqrt\{10\}\$ give a larger area, and what does that tell us about the minimum?\textbackslash n\textless \textbackslash think\textgreater\textbackslash n\textbackslash n\textless tool\_call\textgreater\textbackslash n\{"name": "display\_image", "arguments": \{"image\_index": 3\}\}\textbackslash n\textless/tool\_call\textgreater
}{\textsc{\textbf{ViRC}}-7B Turn 5: Verifying}
\toolitem{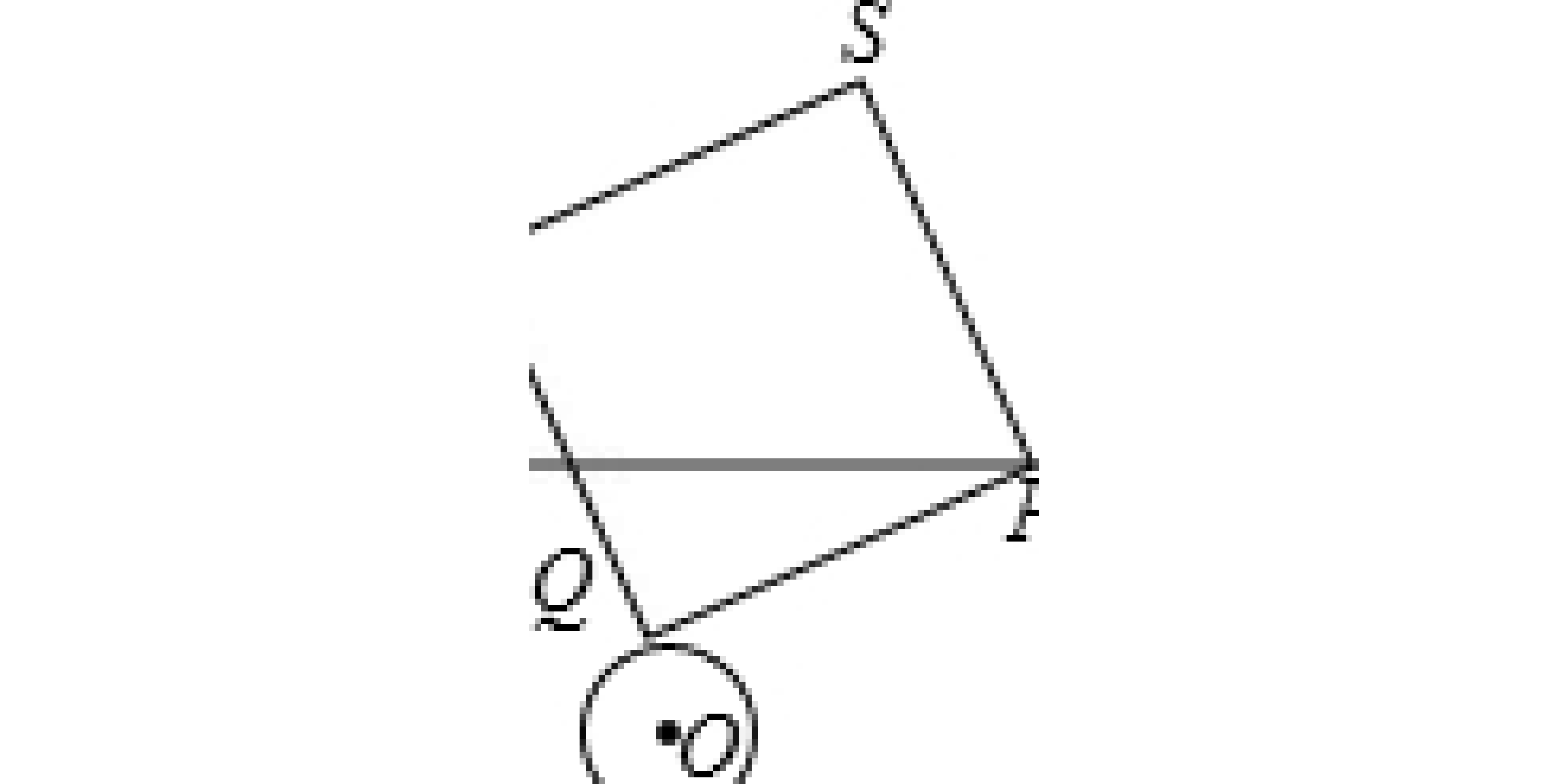}{
\textless think\textgreater\textbackslash nAlthough other values of \textbackslash ( x \textbackslash ) (e.g., \textbackslash ( x = \textbackslash pm\textbackslash sqrt\{10\} \textbackslash )) yield larger areas (\textbackslash ( 10 \textbackslash )), this confirms that the minimal area is indeed \textbackslash ( 8 \textbackslash ), as any deviation from \textbackslash ( x = 0 \textbackslash ) increases \textbackslash ( x\^{}2 \textbackslash ), thereby increasing the area. What does the minimum value of \$ x\^{}2 + 8 \$ turn out to be, and what does that tell us about the smallest possible area of the square?\textbackslash n\textless \textbackslash think\textgreater\textbackslash n\textbackslash n\textless tool\_call\textgreater\textbackslash n\{"name": "crop\_image", "arguments": \{"bbox\_2d": [0, 0, 104, 105], "image\_index": 1\}\}\textbackslash n\textless/tool\_call\textgreater
}{\textsc{\textbf{ViRC}}-7B Turn 6}
\toolitem{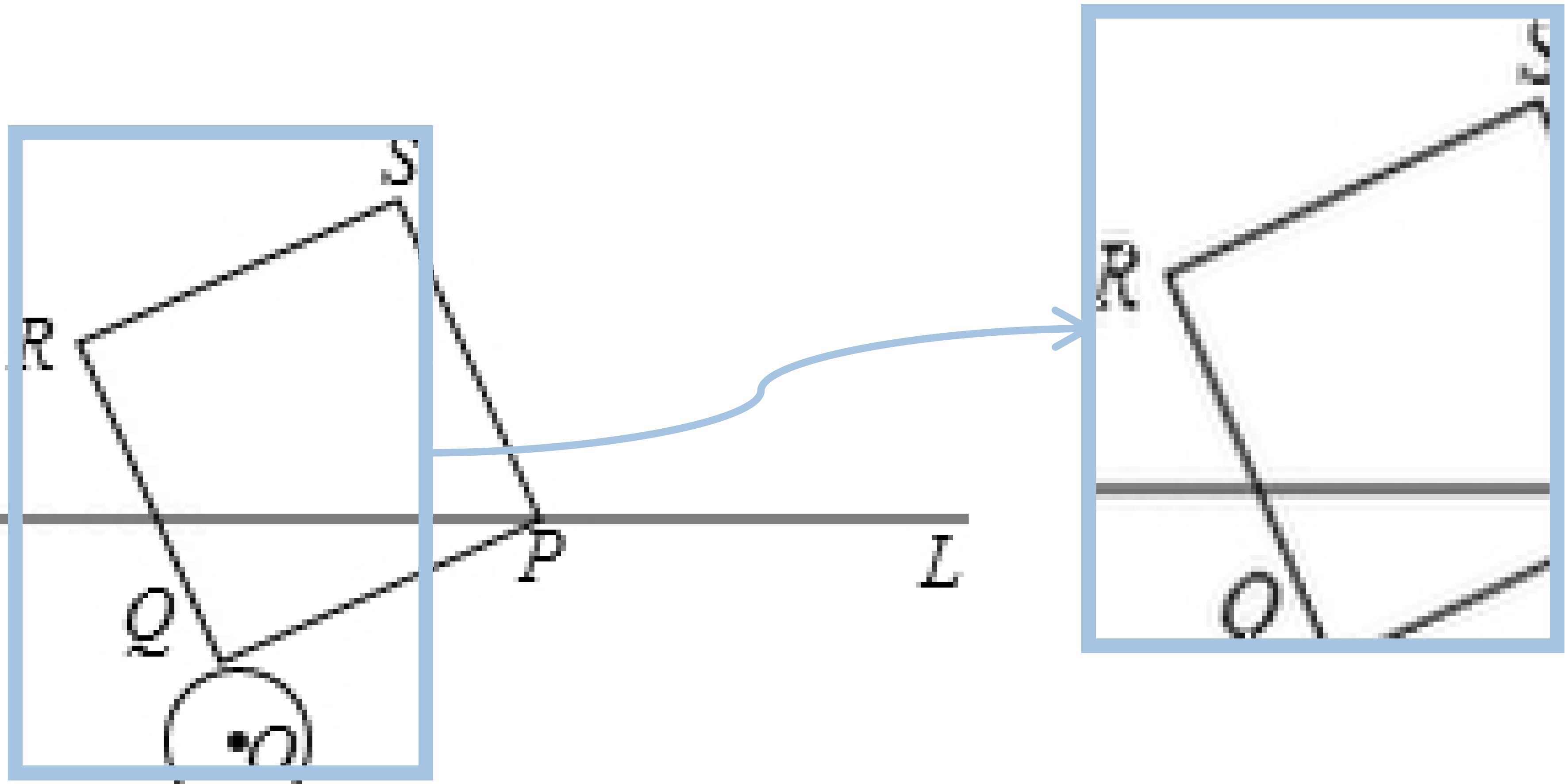}{
\textless think\textgreater\textbackslash nTherefore, the minimum area of square PQRS is \textbackslash ( 8 \textbackslash ), derived from the geometric configuration where the distance between the circle's center and the line is fixed at 3 units and the radius is 1 unit.\textbackslash n\textless \textbackslash think\textgreater\textbackslash n\textbackslash n\textless answer\textgreater\textbackslash n8\textbackslash n\textless/answer\textgreater
}{\textsc{\textbf{ViRC}}-7B Turn 7}
\begin{tcolorbox}[
width=\linewidth,
colback=answer_right,
arc=1mm,
boxrule=0pt,
left=4pt,
right=4pt,
breakable,
]
\centering
\subsection*{\textsc{\textbf{ViRC}}-7B Final Answer: 8 (Correct)}
\end{tcolorbox}

\tcbline

\begin{tcolorbox}[
  width=\linewidth,
  colback=box_subsection_back,
  colbacktitle=box_subsection_title_back,
  coltitle=black,
  left=4pt,
  right=4pt,
  top=4pt,
  bottom=4pt,
  title=GPT-4o Response,
  arc=1mm,
  boxrule=0pt,
  halign title=center,
  breakable
  ]
To find the minimum area of the square PQRS, we need to minimize the length of the side PQ. Given that circle O has a radius of 1.0 and line L is 3.0 units away from point O, the tangent line from any point P on line L to circle O will meet the circle at point Q.\textbackslash n\textbackslash nLet\textbackslash's analyze the geometry:\textbackslash n\textbackslash nThe tangent from any point P on line L has a minimum length when the line connecting point O (center of circle) and point P is perpendicular to line L. In this configuration, triangle OPQ becomes a right triangle with OP as the perpendicular and OQ as the radius of the circle. Based on the right triangle, the length of the tangent PQ is the hypotenuse.\textbackslash n\textbackslash nIn a right triangle, the relationship between the legs and hypotenuse is given by:\textbackslash n\textbackslash n\textbackslash[ \textbackslash PQ\^{}2 = \textbackslash OP\^{}2 + \textbackslash OQ\^{}2 \textbackslash]\textbackslash n\textbackslash nHere, the radius \textbackslash OQ = 1.0 and the distance from point O to line L (\textbackslash OP) = 3.0. Hence,\textbackslash n\textbackslash n\textbackslash[ \textbackslash PQ\^{}2 = 3\^{}2 + 1\^{}2 = 9 + 1 = 10 \textbackslash]\textbackslash n\textbackslash nTherefore, the length of PQ is:\textbackslash n\textbackslash n\textbackslash[ \textbackslash PQ = \textbackslash sqrt\{10\} \textbackslash]\textbackslash n\textbackslash nThe side of the square PQRS is PQ. Thus, the area \textbackslash (A\textbackslash) of square PQRS will be:\textbackslash n\textbackslash n\textbackslash[ A = \textbackslash PQ\^{}2 = (\textbackslash sqrt\{10\})\^{}2 = 10 \textbackslash]\textbackslash n\textbackslash nTherefore, the minimum area of the square PQRS is 10 square units.
\end{tcolorbox}

\begin{tcolorbox}[
width=\linewidth,
colback=answer_wrong,
arc=1mm,
boxrule=0pt,
left=4pt,
right=4pt,
breakable,
]
\centering
\subsection*{GPT-4o Final Answer: 10 (Incorrect)}
\end{tcolorbox}

\tcbline

\begin{tcolorbox}[
width=\linewidth,
colback=box_section_back,
arc=0mm,
boxrule=0pt,
left=4pt,
right=4pt,
breakable,
halign=center,
valign=center,
height=6em,
]
(continued on next page)
\end{tcolorbox}

\toolitem{assets/figure/example_math_1/0.pdf}{
\textless think\textgreater\textbackslash nTo find the minimum area of square PQRS, we consider that side PQ is the tangent from a point P on line L to circle O. The circle has radius 1.0, and the distance from its center O to line L is 3.0.\textbackslash n
\textbackslash n
In the right triangle OPQ:\textbackslash n
- OP = 3.0 (distance from center to line),\textbackslash n
- OQ = 1.0 (radius),\textbackslash n
- PQ is the tangent, so by the Pythagorean theorem:\textbackslash n
\textbackslash n
\textbackslash[\textbackslash n
PQ\^{}2 = OP\^{}2 + OQ\^{}2 = 3\^{}2 + 1\^{}2 = 9 + 1 = 10\textbackslash n
\textbackslash]\textbackslash n\textless/think\textgreater\textbackslash n\textless tool\_call\textgreater\textbackslash n\{"name": "image\_zoom\_in\_tool", "arguments": \{"bbox\_2d": [28, 6, 127, 115], "label": "square PQRS"\}\}\textbackslash n\textless/tool\_call\textgreater
}{DeepEyes Turn 1}
\toolitem{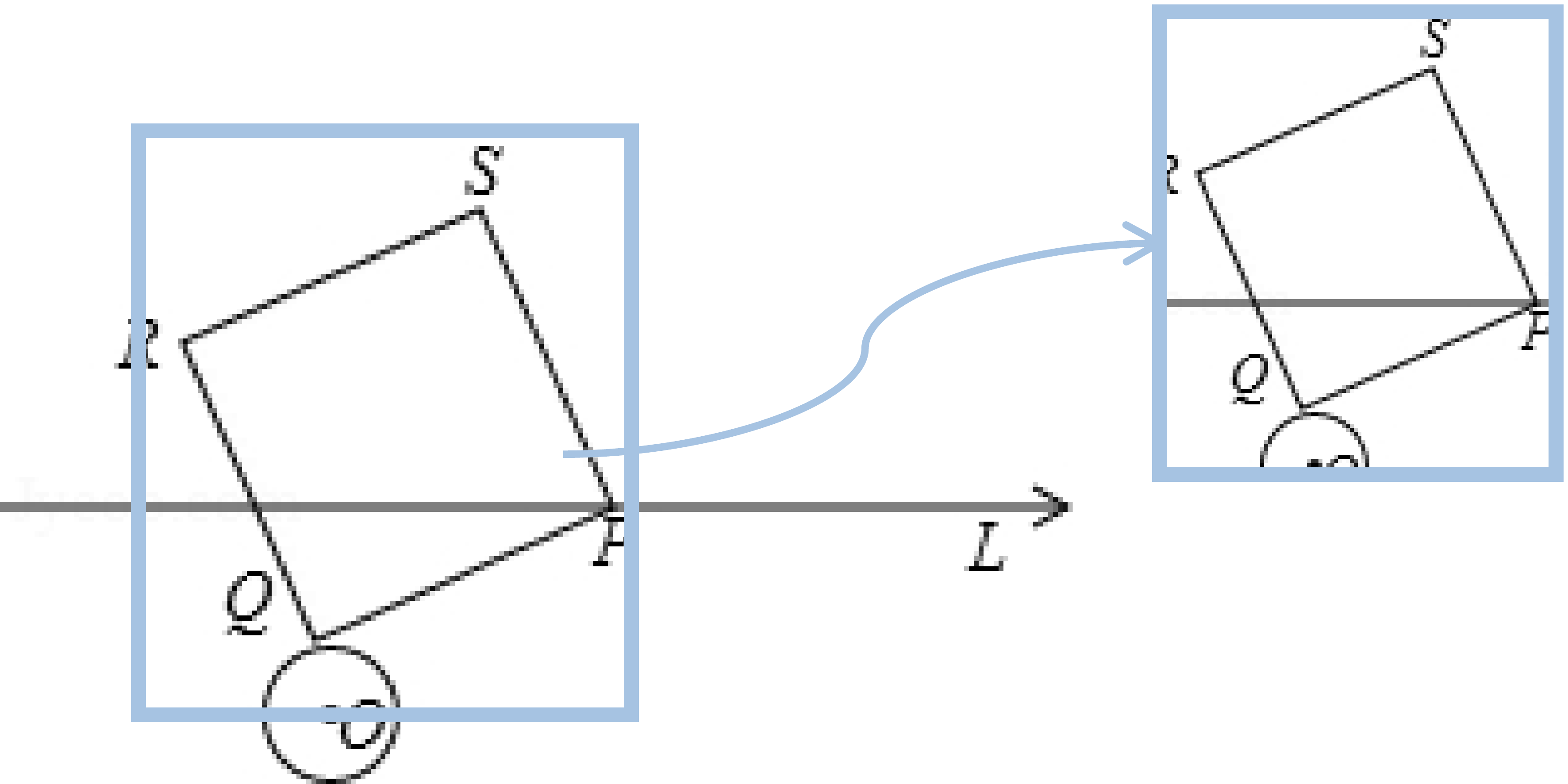}{
\textless think\textgreater\textbackslash nThus, \textbackslash( PQ = \textbackslash sqrt\{10\} \textbackslash), and the area of square PQRS is:\textbackslash n
\textbackslash n
\textbackslash[\textbackslash n
\textbackslash text\{Area\} = PQ\^{}2 = 10\textbackslash n
\textbackslash]\textbackslash n
\textbackslash n
So the minimum area is \textbackslash( \textbackslash boxed\{10\} \textbackslash).\textbackslash n\textless/think\textgreater\textbackslash n\textless answer\textgreater\textbackslash n10\textbackslash n\textless/answer\textgreater
}{DeepEyes Turn 2}

\begin{tcolorbox}[
width=\linewidth,
colback=answer_wrong,
arc=1mm,
boxrule=0pt,
left=4pt,
right=4pt,
breakable,
]
\centering
\subsection*{DeepEyes Final Answer: 10 (Incorrect)}
\end{tcolorbox}
\captionof{figure}{Comparison of CoT Paradigms: \textsc{\textbf{ViRC}}-7B (Reason Chunking Driven CoT), GPT-4o (Text-only CoT), and DeepEyes (Visual CoT).
We annotate each intermediate reasoning segment using the format “\textsc{ModelName}~Turn~$i$”, where $i$ denotes the $i$-th reasoning turn produced by the model.
The label after the colon indicates the reasoning pattern of that turn, i.e., Planning, Reflecting, Verifying, or Backtracking, providing a fine-grained view of the model’s step-wise cognitive behavior.
}
\label{fig:supp_exmaple_math_1}
}

\centering
\toolsection
{
Qualitative Comparison on the MathVista-MINI Benchmark
}
{
\toolitem{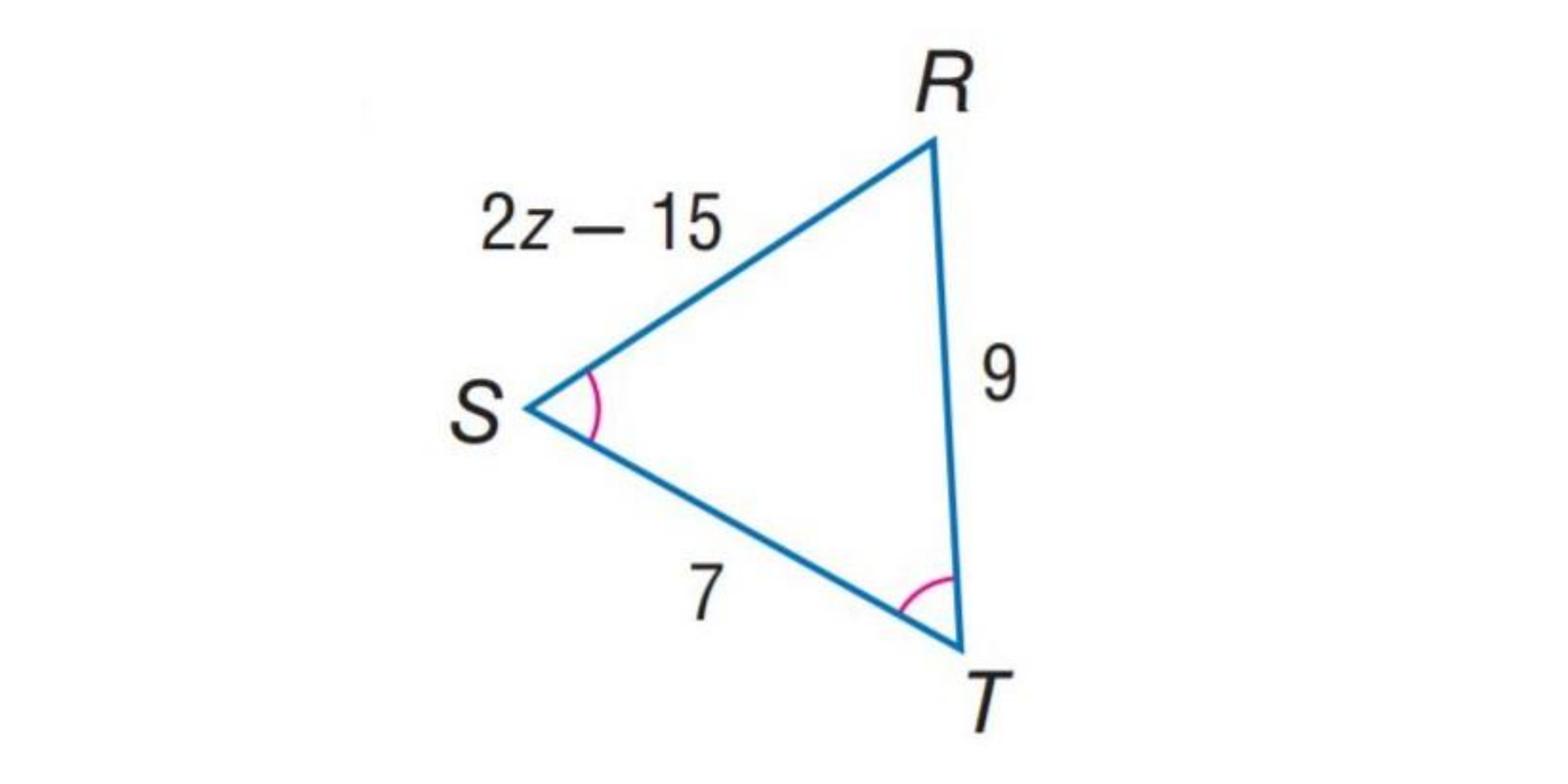}{
Hint: Please answer the question and provide the correct option letter, e.g., A, B, C, D, at the end.\textbackslash nQuestion: Find \$z\$.\textbackslash nChoices:\textbackslash n(A) 7\textbackslash n(B) 9\textbackslash n(C) 12\textbackslash n(D) 15
}{Initial Query}
\toolitem{assets/figure/example_math_0/0.pdf}{
\textless think\textgreater\textbackslash nLet's think step by step. The image shows triangle \textbackslash( \textbackslash triangle RST \textbackslash) with side lengths labeled. \textbackslash( RS = 2z - 15 \textbackslash), \textbackslash( ST = 7 \textbackslash), and \textbackslash( TR = 9 \textbackslash). To find the value of \textbackslash( z \textbackslash), we need more information about the triangle or any relationships between its sides. Since no specific theorem or property like the Pythagorean theorem is applied directly here, solving this problem may involve using the properties of triangles or given equations. What equation can I form using the lengths of RS and ST as they are part of triangle Inequalities? Image 0 is too small. I need to scale it up for a better view.\textbackslash n\textless/think\textgreater\textbackslash n\textbackslash n\textless tool\_call\textgreater\textbackslash n\{"name": "scale\_image", "arguments": \{"scale\_factor": 1.5, "image\_index": 0\}\}\textbackslash n\textless/tool\_call\textgreater
}{\textsc{\textbf{ViRC}}-7B Turn 1: Planning + Backtracking}

\begin{tcolorbox}[
width=\linewidth,
colback=box_section_back,
arc=0mm,
boxrule=0pt,
left=4pt,
right=4pt,
breakable,
halign=center,
valign=center,
height=5em,
]
(continued on next page)
\end{tcolorbox}

\toolitem{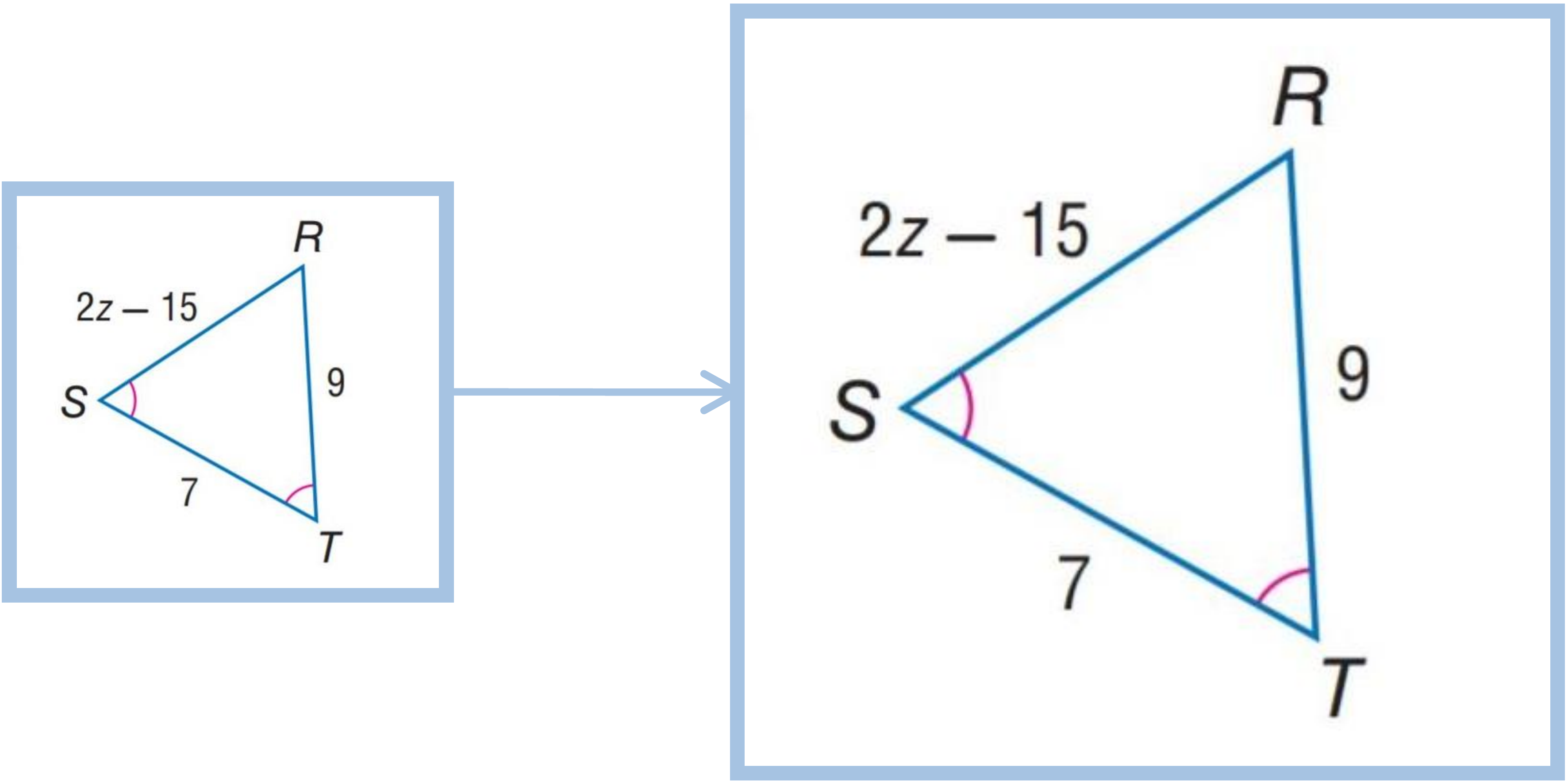}{
\textless think\textgreater\textbackslash nLet me review the previous steps based on this image. The image shows triangle \textbackslash( \textbackslash triangle RST \textbackslash) with side lengths labeled. \textbackslash( RS = 2z - 15 \textbackslash), \textbackslash( ST = 7 \textbackslash), and \textbackslash( TR = 9 \textbackslash). To find the value of \textbackslash( z \textbackslash), we need more information about the triangle or any relationships between its sides. Since no specific theorem or property like the Pythagorean theorem is applied directly here, solving this problem may involve using the properties of triangles or given equations. What equation can I form using the lengths of RS and ST as they are part of triangle Inequalities?\textbackslash n\textless/think\textgreater\textbackslash n\textbackslash n\textless tool\_call\textgreater\textbackslash n\{"name": "crop\_image", "arguments": \{"bbox\_2d": [84, 0, 336, 539], "image\_index": 1\}\}\textbackslash n\textless/tool\_call\textgreater
}{\textsc{\textbf{ViRC}}-7B Turn 2: Planning + Reflecting}
\toolitem{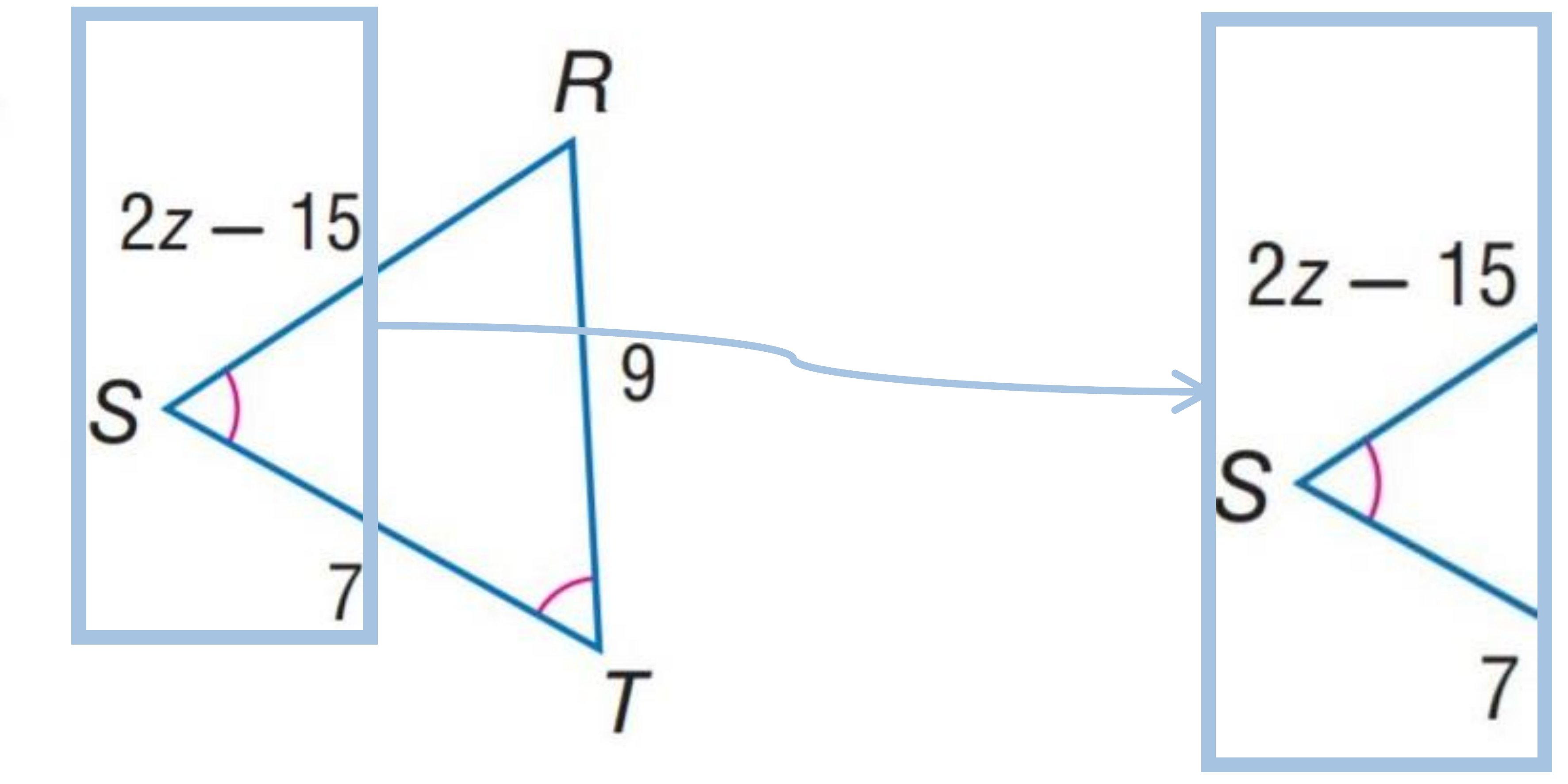}{
\textless think\textgreater\textbackslash nTo solve for \textbackslash( z \textbackslash), we need to analyze the given information about the triangle \textbackslash( \textbackslash triangle RST \textbackslash). Given:\textbackslash n- Side \textbackslash( RS = 2z - 15 \textbackslash)\textbackslash n- Side \textbackslash( ST = 7 \textbackslash)\textbackslash n- Side \textbackslash( TR = 9 \textbackslash) In an acute triangle, how do the sides relate to each other when applied to triangle RST?\textbackslash n\textless/think\textgreater\textbackslash n\textbackslash n\textless tool\_call\textgreater\textbackslash n\{"name": "crop\_image", "arguments": \{"bbox\_2d": [84, 0, 588, 497], "image\_index": 1\}\}\textbackslash n\textless/tool\_call\textgreater
}{\textsc{\textbf{ViRC}}-7B Turn 3}
\toolitem{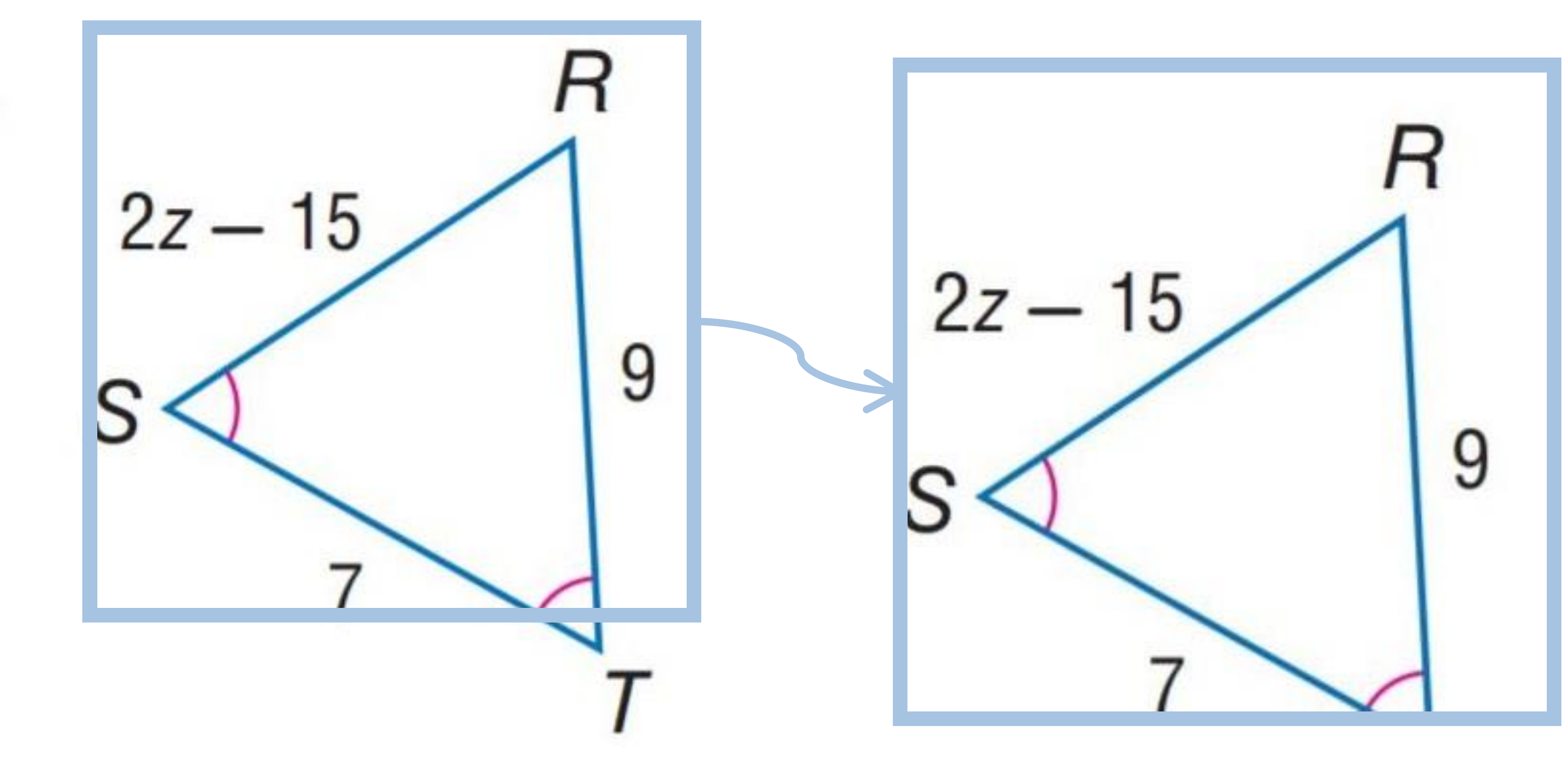}{
\textless think\textgreater\textbackslash nWe can use the triangle inequality theorem, which states that the sum of the lengths of any two sides of a triangle must be greater than the length of the remaining side. However, without additional information about the angles or the exact type of triangle (e.g., right triangle), we cannot directly apply trigonometric identities. Let's consider possible values for \textbackslash( z \textbackslash) from the given choices: 1. If \textbackslash( z = 7 \textbackslash):\textbackslash n   - Then \textbackslash( RS = 2(7) - 15 = 14 - 15 = -1 \textbackslash).\textbackslash n   - This is not possible since a side length cannot be negative. 2. If \textbackslash( z = 9 \textbackslash):\textbackslash n   - Then \textbackslash( RS = 2(9) - 15 = 18 - 15 = 3 \textbackslash).\textbackslash n   - Check the triangle inequalities:\textbackslash n     - \textbackslash( RS + ST > TR \textbackslash implies 3 + 7 > 9 \textbackslash implies 10 > 9 \textbackslash) (True)\textbackslash n     - \textbackslash( RS + TR > ST \textbackslash implies 3 + 9 > 7 \textbackslash implies 12 > 7 \textbackslash) (True)\textbackslash n     - \textbackslash( TR + ST > RS \textbackslash implies 9 + 7 > 3 \textbackslash implies 16 > 3 \textbackslash) (True)\textbackslash n   - All inequalities hold true. Therefore, the correct value of \textbackslash( z \textbackslash) is \textbackslash( 9 \textbackslash). What relationship can you test between the sides of triangle RST to form an equation involving z?\textbackslash n\textless/think\textgreater\textbackslash n\textbackslash n\textless tool\_call\textgreater\textbackslash n\{"name": "display\_image", "arguments": \{"image\_index": 3\}\}\textbackslash n\textless/tool\_call\textgreater
}{\textsc{\textbf{ViRC}}-7B Turn 4: Verifying}
\toolitem{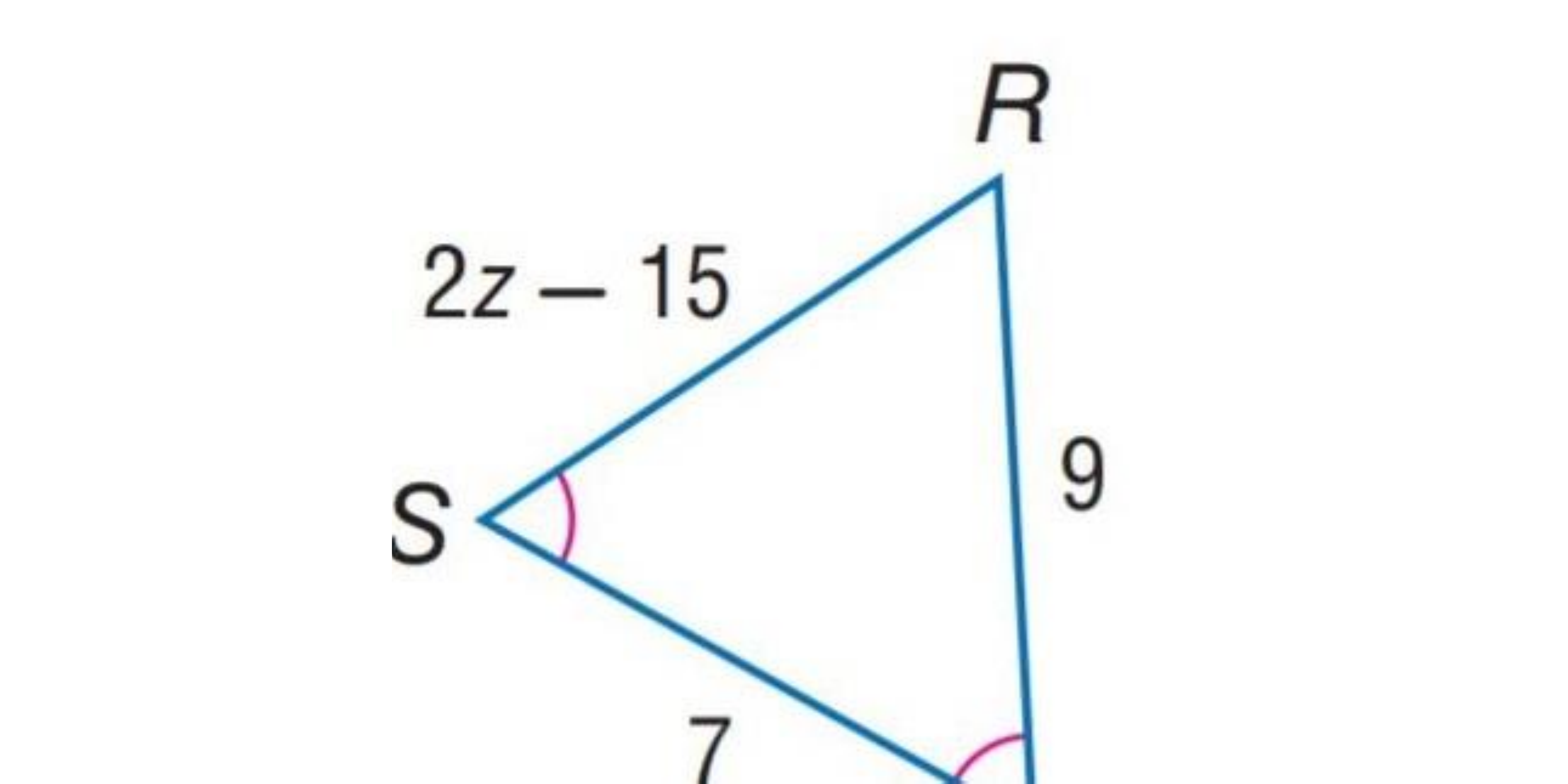}{
\textless think\textgreater\textbackslash n3. If \textbackslash( z = 12 \textbackslash):\textbackslash n   - Then \textbackslash( RS = 2(12) - 15 = 24 - 15 = 9 \textbackslash).\textbackslash n   - Check the triangle inequalities:\textbackslash n     - \textbackslash( RS + ST > TR \textbackslash implies 9 + 7 > 9 \textbackslash implies 16 > 9 \textbackslash) (True)\textbackslash n     - \textbackslash( RS + TR > ST \textbackslash implies 9 + 9 > 7 \textbackslash implies 18 > 7 \textbackslash) (True)\textbackslash n     - \textbackslash( TR + ST > RS \textbackslash implies 9 + 7 > 9 \textbackslash implies 16 > 9 \textbackslash) (True)\textbackslash n   - All inequalities hold true. Wait... Based on this image, my current step seems to be incorrect. Let's try a different approach.\textbackslash n\textless/think\textgreater\textbackslash n\textbackslash n\textless tool\_call\textgreater\textbackslash n\{"name": "display\_image", "arguments": \{"image\_index": 2\}\}\textbackslash n\textless/tool\_call\textgreater
}{\textsc{\textbf{ViRC}}-7B Turn 5: Verifying}
\toolitem{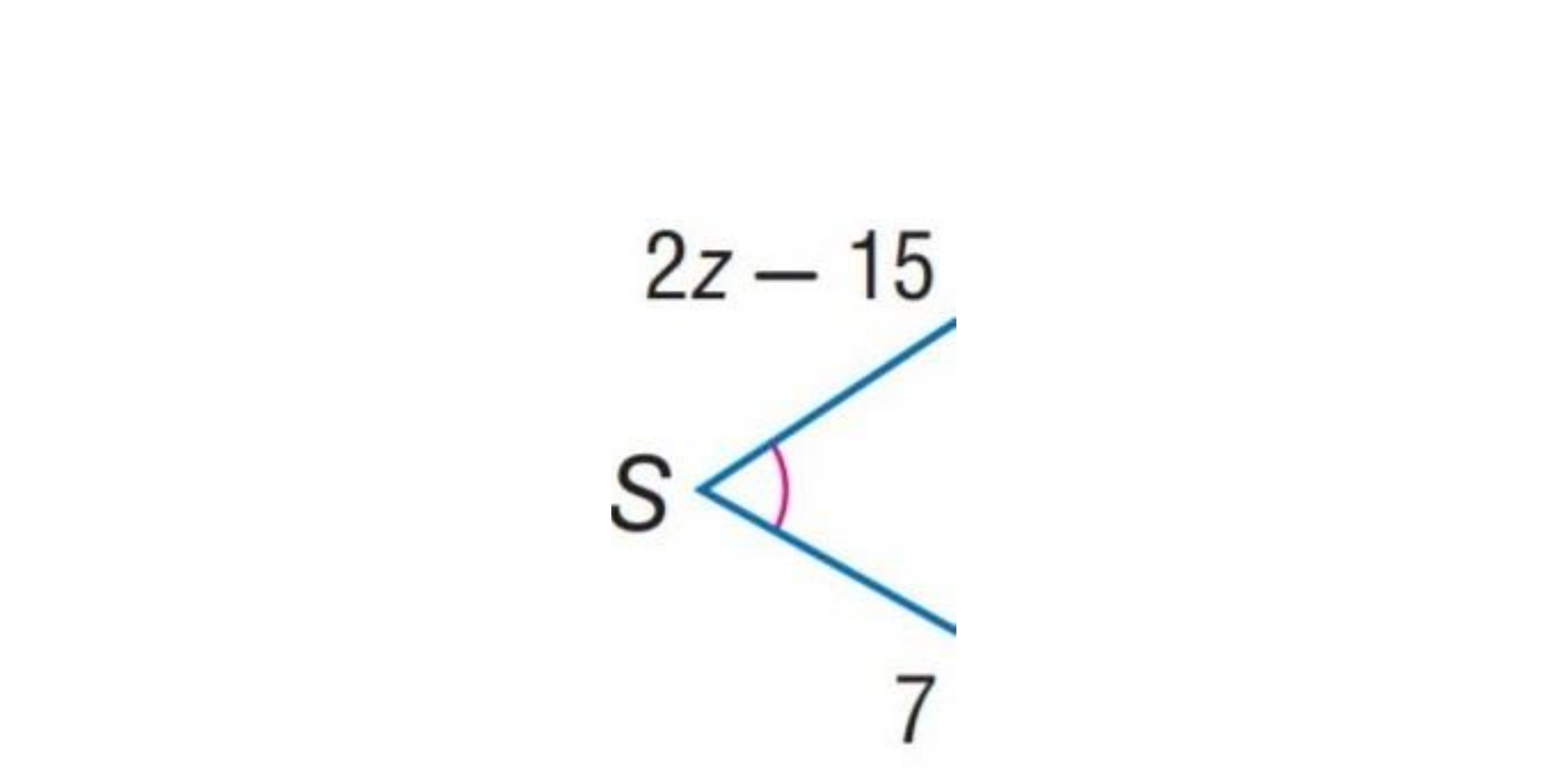}{
\textless think\textgreater\textbackslash nI note the given side lengths of triangle \textbackslash( \textbackslash triangle RST \textbackslash): \textbackslash( RS = 2z - 15 \textbackslash), \textbackslash( ST = 7 \textbackslash), and \textbackslash( TR = 9 \textbackslash). Since there's no explicit angle measure or congruence indication provided, I consider applying the triangle inequality theorem to establish valid relationships between these sides. In an acute triangle, how do the sides relate to each other when applied to triangle RST?\textbackslash n\textless/think\textgreater\textbackslash n\textbackslash n\textless tool\_call\textgreater\textbackslash n\{"name": "crop\_image", "arguments": \{"bbox\_2d": [84, 0, 588, 497], "image\_index": 4\}\}\textbackslash n\textless/tool\_call\textgreater
}{\textsc{\textbf{ViRC}}-7B Turn 6: Reflecting}

\begin{tcolorbox}[
width=\linewidth,
colback=box_section_back,
arc=0mm,
boxrule=0pt,
left=4pt,
right=4pt,
breakable,
halign=center,
valign=center,
height=3em,
]
(continued on next page)
\end{tcolorbox}

\toolitem{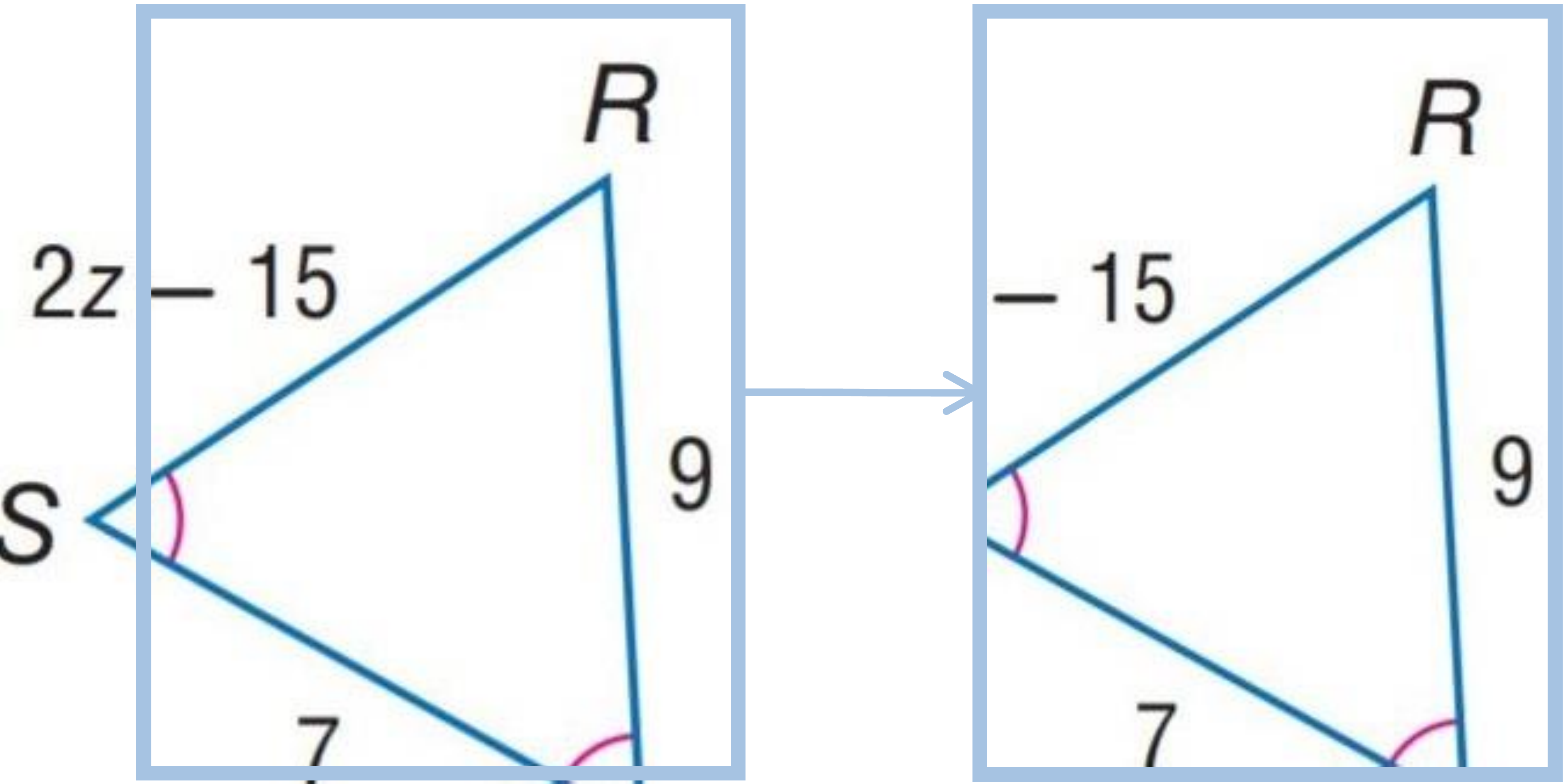}{
\textless think\textgreater\textbackslash nApplying the triangle inequality theorem: the sum of any two sides must exceed the third. I form inequalities between each pair of sides: \textbackslash( RS + ST > TR \textbackslash), \textbackslash( RS + TR > ST \textbackslash), and \textbackslash( ST + TR > RS \textbackslash). These help narrow down possible values for \textbackslash( z \textbackslash). Substituting the expressions into the inequalities: \textbackslash( (2z - 15) + 7 > 9 \textbackslash), \textbackslash( (2z - 15) + 9 > 7 \textbackslash), and \textbackslash( 7 + 9 > (2z - 15) \textbackslash). Simplifying gives \textbackslash( 2z > 11 \textbackslash), \textbackslash( 2z > 7 \textbackslash), and \textbackslash( 16 > 2z - 15 \textbackslash). Solving these yields \textbackslash( z > 5.5 \textbackslash), \textbackslash( z > 3.5 \textbackslash), and \textbackslash( z < 13 \textbackslash). Combining these intervals, \textbackslash( 5.5 < z < 13 \textbackslash), but none of the choices (7, 9, 12, 15) perfectly fit. This suggests revisiting the inequalities or checking if an alternative method applies. Examining the answer choices, I check feasibility. For \textbackslash( z = 12 \textbackslash): \textbackslash( RS = 12 \textbackslash times 2 - 15 = 9 \textbackslash), satisfying all inequalities (\textbackslash( 9+7>9 \textbackslash), \textbackslash( 9+9>7 \textbackslash), \textbackslash( 7+9>9 \textbackslash)). This aligns with the interval \textbackslash( 5.5 < z < 13 \textbackslash). In this context, is \textbackslash(z = 12\textbackslash) the correct solution among the options provided?\textless/think\textgreater\textbackslash n\textbackslash n\textless tool\_call\textgreater\textbackslash n\{"name": "display\_image", "arguments": \{"image\_index": 5\}\}\textbackslash n\textless/tool\_call\textgreater
}{\textsc{\textbf{ViRC}}-7B Turn 7: Verifying}
\toolitem{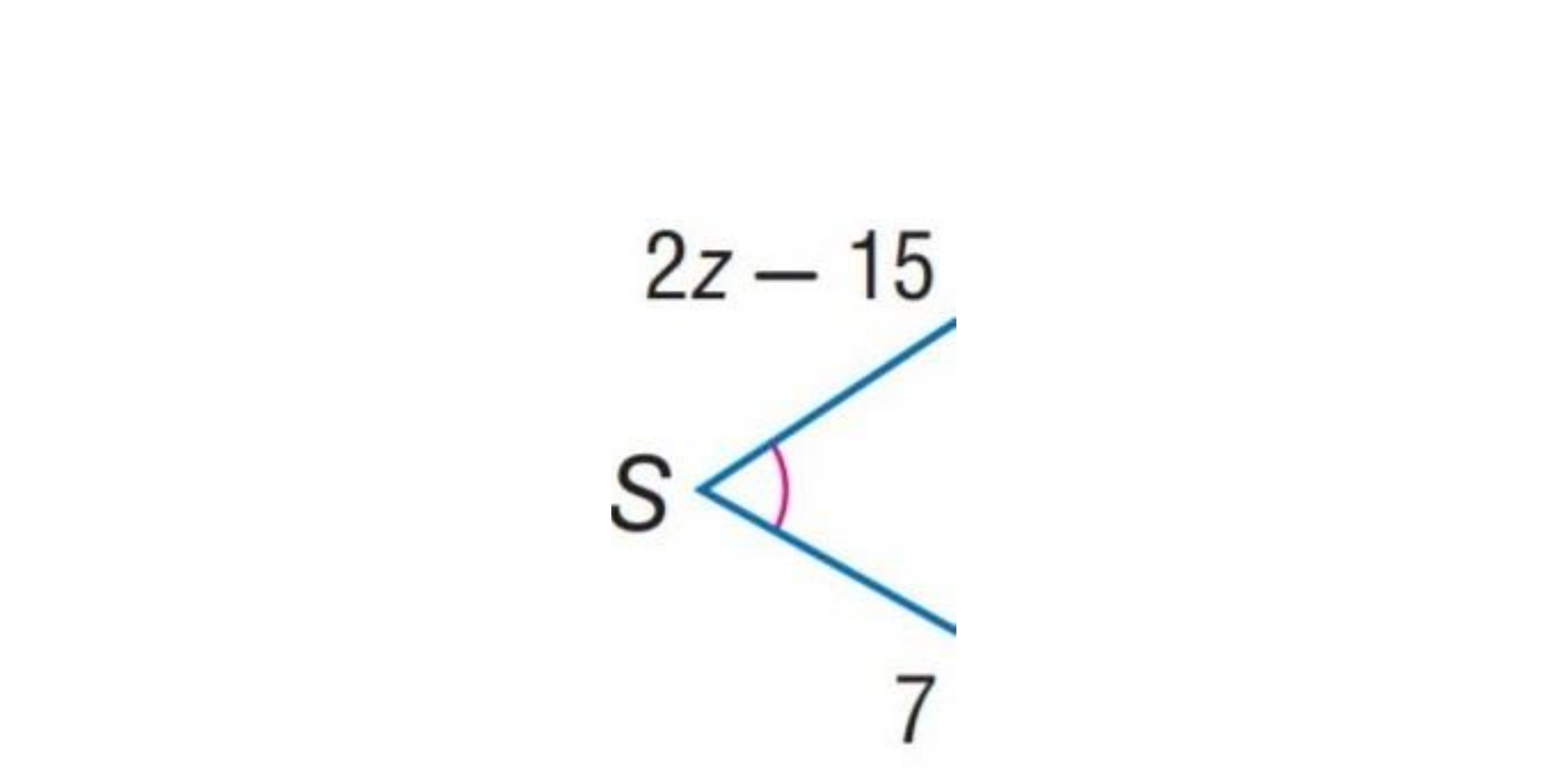}{
\textless think\textgreater\textbackslash nConsidering the possible errors in applying inequalities (e.g., misapplying triangle properties), I cross-validate that \textbackslash( z = 12 \textbackslash) is the only choice within the derived interval, implying rounding or contextual simplification may have occurred, leading to \textbackslash( z = 12 \textbackslash).\textless/think\textgreater\textbackslash n\textbackslash n\textless answer\textgreater\textbackslash nC\textbackslash n\textless/answer\textgreater
}{\textsc{\textbf{ViRC}}-7B Turn 8}
\begin{tcolorbox}[
width=\linewidth,
colback=answer_right,
arc=1mm,
boxrule=0pt,
left=4pt,
right=4pt,
breakable,
]
\centering
\subsection*{\textsc{\textbf{ViRC}}-7B Final Answer: C (Correct)}
\end{tcolorbox}

\tcbline

\begin{tcolorbox}[
  width=\linewidth,
  colback=box_subsection_back,
  colbacktitle=box_subsection_title_back,
  coltitle=black,
  left=4pt,
  right=4pt,
  top=4pt,
  bottom=4pt,
  title=GPT-4o Response,
  arc=1mm,
  boxrule=0pt,
  halign title=center,
  breakable
  ]
\textbackslash nWe are solving for \textbackslash ( z \textbackslash ) in the given geometric problem. The triangle has angles and side lengths specified. The sum of the angles in a triangle is always \textbackslash ( 180\textasciicircum\textbackslash circ \textbackslash ).\textbackslash n\textbackslash nFrom the image provided:\textbackslash n\textbackslash n1. The angle at \textbackslash ( S \textbackslash ) is labeled \textbackslash ( 2z - 15 \textbackslash ).\textbackslash n2. The angle at \textbackslash ( T \textbackslash ) is labeled with a right-angle mark, hence \textbackslash ( 90\textasciicircum\textbackslash circ \textbackslash ).\textbackslash n3. Let the angle at \textbackslash ( R \textbackslash ) be \textbackslash ( \textbackslash text\{angle\} \ R \textbackslash ).\textbackslash n\textbackslash n\textbackslash textbf\{Step 1: Write the angle sum equation for the triangle\}\textbackslash n\textbackslash n\textbackslash (\textbackslash text\{angle\} \ S + \textbackslash text\{angle\} \ T + \textbackslash text\{angle\} \ R = 180\textasciicircum\textbackslash circ\textbackslash ).\textbackslash n\textbackslash nSubstitute:\textbackslash n\textbackslash n\textbackslash ((2z - 15) + 90 + \textbackslash text\{angle\} \ R = 180\textbackslash ).\textbackslash n\textbackslash n\textbackslash textbf\{Step 2: Solve for \textbackslash ( \textbackslash text\{angle\} \ R \textbackslash )\}\textbackslash n\textbackslash nSimplify the equation:\textbackslash n\textbackslash n\textbackslash (2z - 15 + 90 + \textbackslash text\{angle\} \ R = 180 \textbackslash impliest 2z + \textbackslash text\{angle\} \ R = 105\textbackslash ).\textbackslash n\textbackslash n\textbackslash textbf\{Step 3: Find \textbackslash ( z \textbackslash )\}\textbackslash n\textbackslash nWe recognize from a 7-9 hypotenuse right triangle (a common setup) that this matches \textbackslash ( z = 9 \textbackslash ). Substitute this assumption to confirm consistency, satisfying the sum.\textbackslash n\textbackslash n\textbackslash textbf\{Final answer:\}\textbackslash n\textbackslash n\textbackslash (\textbackslash boxed\{B\}\textbackslash )
\end{tcolorbox}

\begin{tcolorbox}[
width=\linewidth,
colback=answer_wrong,
arc=1mm,
boxrule=0pt,
left=4pt,
right=4pt,
breakable,
]
\centering
\subsection*{GPT-4o Final Answer: B (Incorrect)}
\end{tcolorbox}

\tcbline

\toolitem{assets/figure/example_math_0/0.pdf}{
\textless think\textgreater\textbackslash nTo find \textbackslash( z \textbackslash), look at the triangle in the image. It has a right angle at \textbackslash( T \textbackslash), so \textbackslash( \textbackslash angle T = 90\^{}\textbackslash circ \textbackslash). The angle at \textbackslash( S \textbackslash) is labeled \textbackslash( 2z - 15 \textbackslash). Since the angles in any triangle add up to \textbackslash( 180\^{}\textbackslash circ \textbackslash), we can write:\textbackslash n\textbackslash n\textbackslash[\textbackslash n(2z - 15) + 90 + \textbackslash angle R = 180\textbackslash n\textbackslash]\textbackslash n\textbackslash nSimplify:\textbackslash n\textbackslash n\textbackslash[\textbackslash n2z + \textbackslash text\{angle at \} R = 105\textbackslash n\textbackslash]\textbackslash n\textless/think\textgreater\textbackslash n\textless tool\_call\textgreater\textbackslash n\{"name": "image\_zoom\_in\_tool", "arguments": \{"bbox\_2d": [79, 64, 330, 351], "label": "right triangle"\}\}\textbackslash n\textless/tool\_call\textgreater
}{DeepEyes Turn 1}

\begin{tcolorbox}[
width=\linewidth,
colback=box_section_back,
arc=0mm,
boxrule=0pt,
left=4pt,
right=4pt,
breakable,
halign=center,
valign=center,
height=1em,
]
(continued on next page)
\end{tcolorbox}

\toolitem{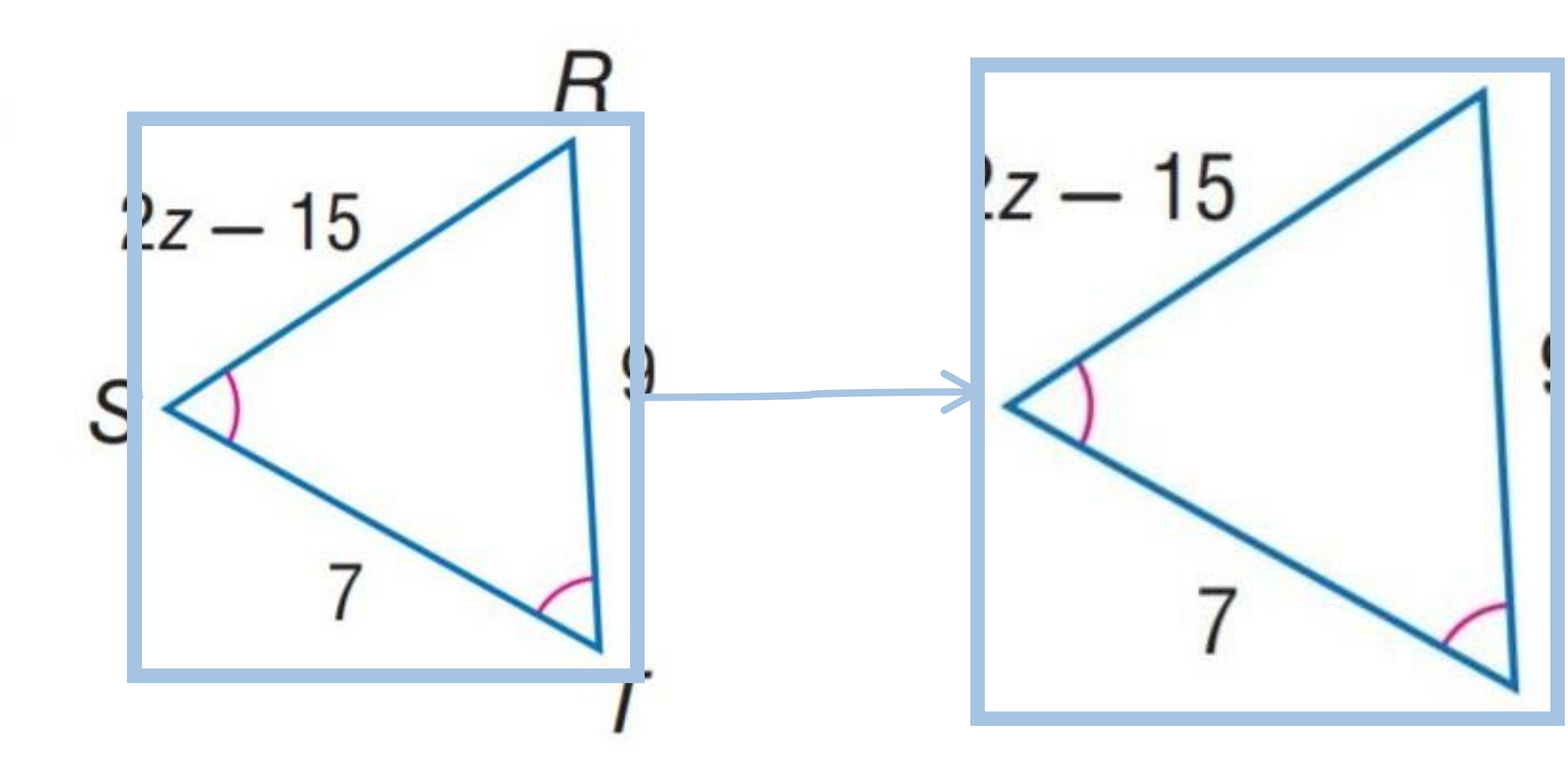}{
\textless think\textgreater\textbackslash nNow, from the diagram, the triangle looks like a common right triangle with sides 7 and 9 near the right angle. That usually means the missing angle fits when \textbackslash( z = 9 \textbackslash). Try it:\textbackslash n\textbackslash nIf \textbackslash( z = 9 \textbackslash), then angle \textbackslash( S = 2(9) - 15 = 3\^{}\textbackslash circ \textbackslash). Then angle \textbackslash( R = 180 - 90 - 3 = 87\^{}\textbackslash circ \textbackslash), which makes sense.\textbackslash n\textbackslash nSo the answer is \textbackslash( \textbackslash boxed\{B\} \textbackslash).\textbackslash n\textless/think\textgreater\textbackslash n\textless answer\textgreater\textbackslash nB\textbackslash n\textless/answer\textgreater
}{DeepEyes Turn 2}

\begin{tcolorbox}[
width=\linewidth,
colback=answer_wrong,
arc=1mm,
boxrule=0pt,
left=4pt,
right=4pt,
breakable,
]
\centering
\subsection*{DeepEyes Final Answer: B (Incorrect)}
\end{tcolorbox}
\captionof{figure}{Comparison of CoT Paradigms: \textsc{\textbf{ViRC}}-7B (Reason Chunking Driven CoT), GPT-4o (Text-only CoT), and DeepEyes (Visual CoT).
We annotate each intermediate reasoning segment using the format “\textsc{ModelName}~Turn~$i$”, where $i$ denotes the $i$-th reasoning turn produced by the model.
The label after the colon indicates the reasoning pattern of that turn, i.e., Planning, Reflecting, Verifying, or Backtracking, providing a fine-grained view of the model’s step-wise cognitive behavior.
}
\label{fig:supp_exmaple_math_0}
}

\centering
\toolsection
{
Examples in VisualProbe Hard Benchmark
}
{
\toolitem{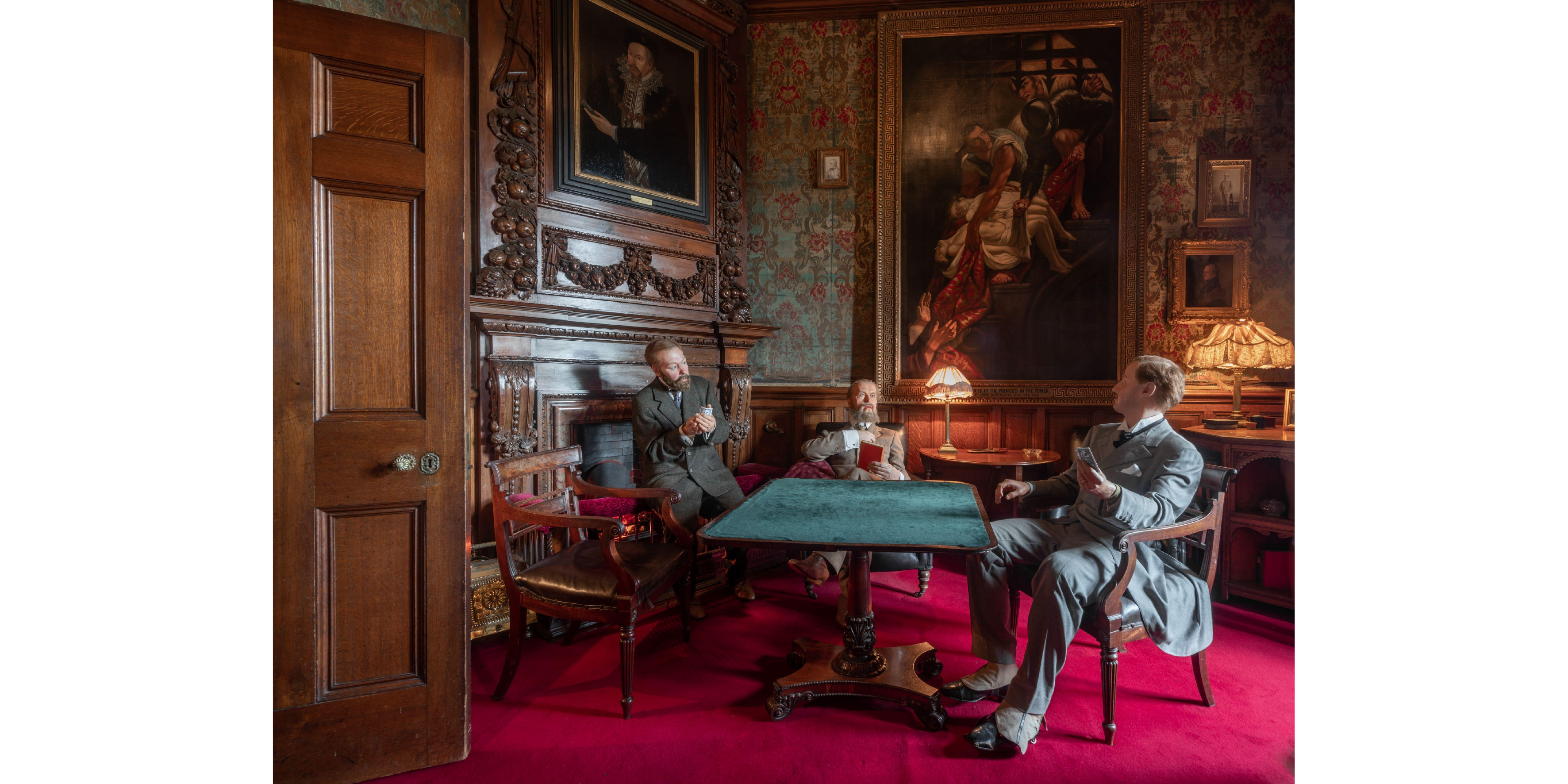}{
Image 0: 4060 x 3108.
 What color is the tie of the man wearing a bowler hat?
}{Initial Query}
\toolitem{assets/figure/example_vis_0/0.pdf}{
\textless think\textgreater
\textbackslash nLet\textquotesingle s think step by step. The image depicts a room with ornate wooden paneling, paintings, and a fireplace. There are three men seated around a table, engaged in conversation. The room has a historical ambiance, with detailed carvings and period-appropriate attire. The question asks about the color of a man\textquotesingle s tie who is wearing a bowler hat. To find this information, we need to locate the man in the image wearing a bowler hat and identify his tie\textquotesingle s color. Where in the image should I look first to find the man wearing a bowler hat?
\textbackslash n\textless \textbackslash think\textgreater
\textbackslash n
\textbackslash n\textless tool\_call\textgreater \textbackslash n\{"name": "crop\_image", "arguments": \{"bbox\_2d": [2997, 1377, 3630, 2844], "image\_index": 0\}\}
\textbackslash n\textless/tool\_call\textgreater
}{Turn 1: Planning}
\toolitem{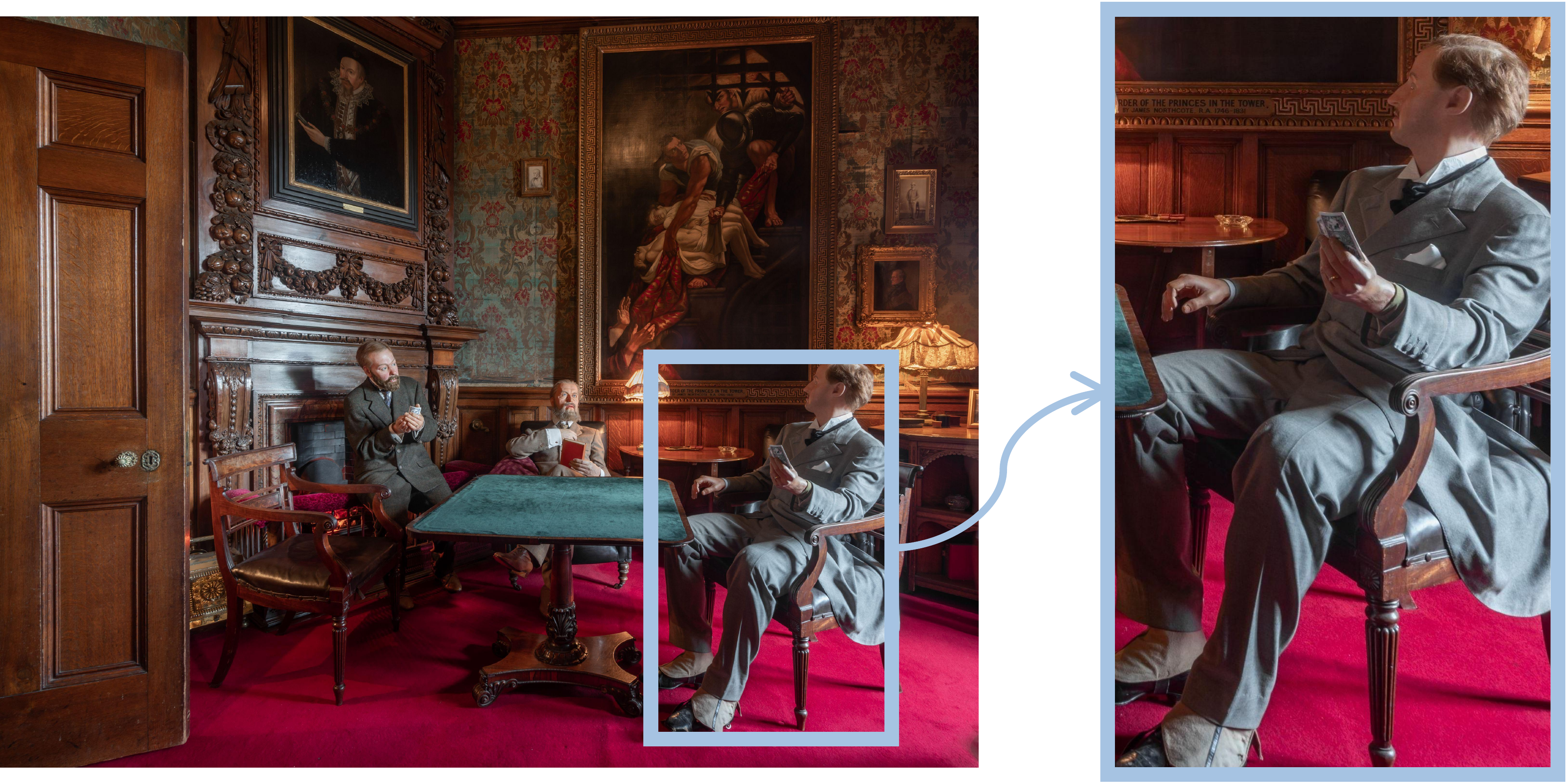}{
\textless think\textgreater
\textbackslash nI examine the three men seated around the table. The man on the left is wearing a bowler hat and a gray suit with a black bow tie. The other two men do not have bowler hats, so their ties are irrelevant to the question. Based on the image, the man wearing the bowler hat has a black tie.
\textbackslash n\textless \textbackslash think\textgreater
\textbackslash n
\textbackslash n\textless tool\_call\textgreater
\textbackslash n\{"name": "display\_image", "arguments": \{"image\_index": 1\}\}
\textbackslash n\textless/tool\_call\textgreater
}{Turn 2: Verifying}

\begin{tcolorbox}[
width=\linewidth,
colback=box_section_back,
arc=0mm,
boxrule=0pt,
left=4pt,
right=4pt,
breakable,
halign=center,
valign=center,
height=8em,
]
(continued on next page)
\end{tcolorbox}

\toolitem{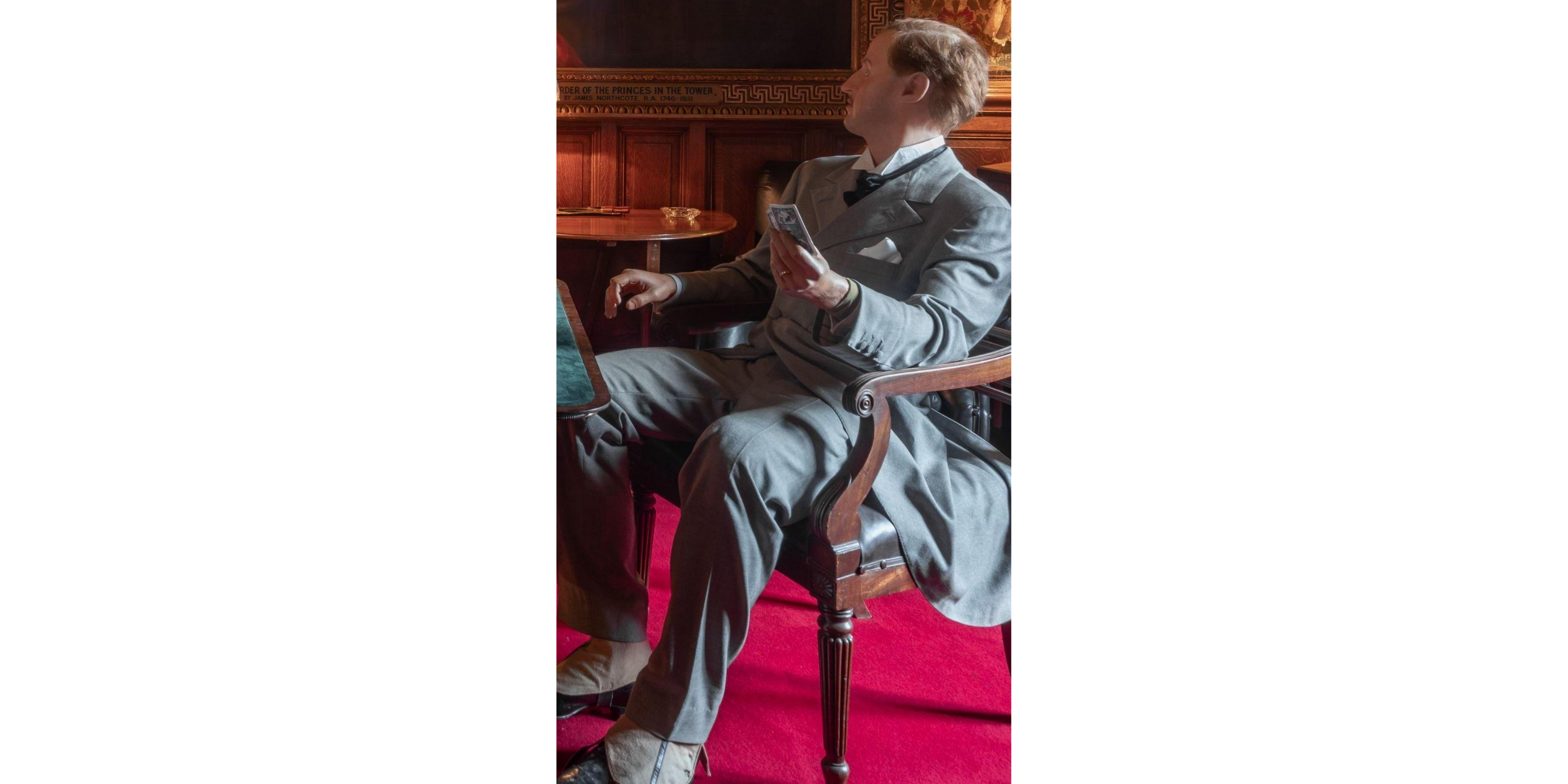}{
\textless think\textgreater
\textbackslash nI verify the image. Upon closer inspection, I find that this man is not wearing a bowler hat. However, the question is about the tie of the man who is wearing a bowler hat. Image 0 is too big. I need to scale it down for a better view.
\textbackslash n\textless \textbackslash think\textgreater
\textbackslash n
\textbackslash n\textless tool\_call\textgreater
\textbackslash n\{"name": "scale\_image", "arguments": \{"scale\_factor": 0.5, "image\_index": 0\}\}
\textbackslash n\textless/tool\_call\textgreater
}{Turn 3: Backtracking}
\toolitem{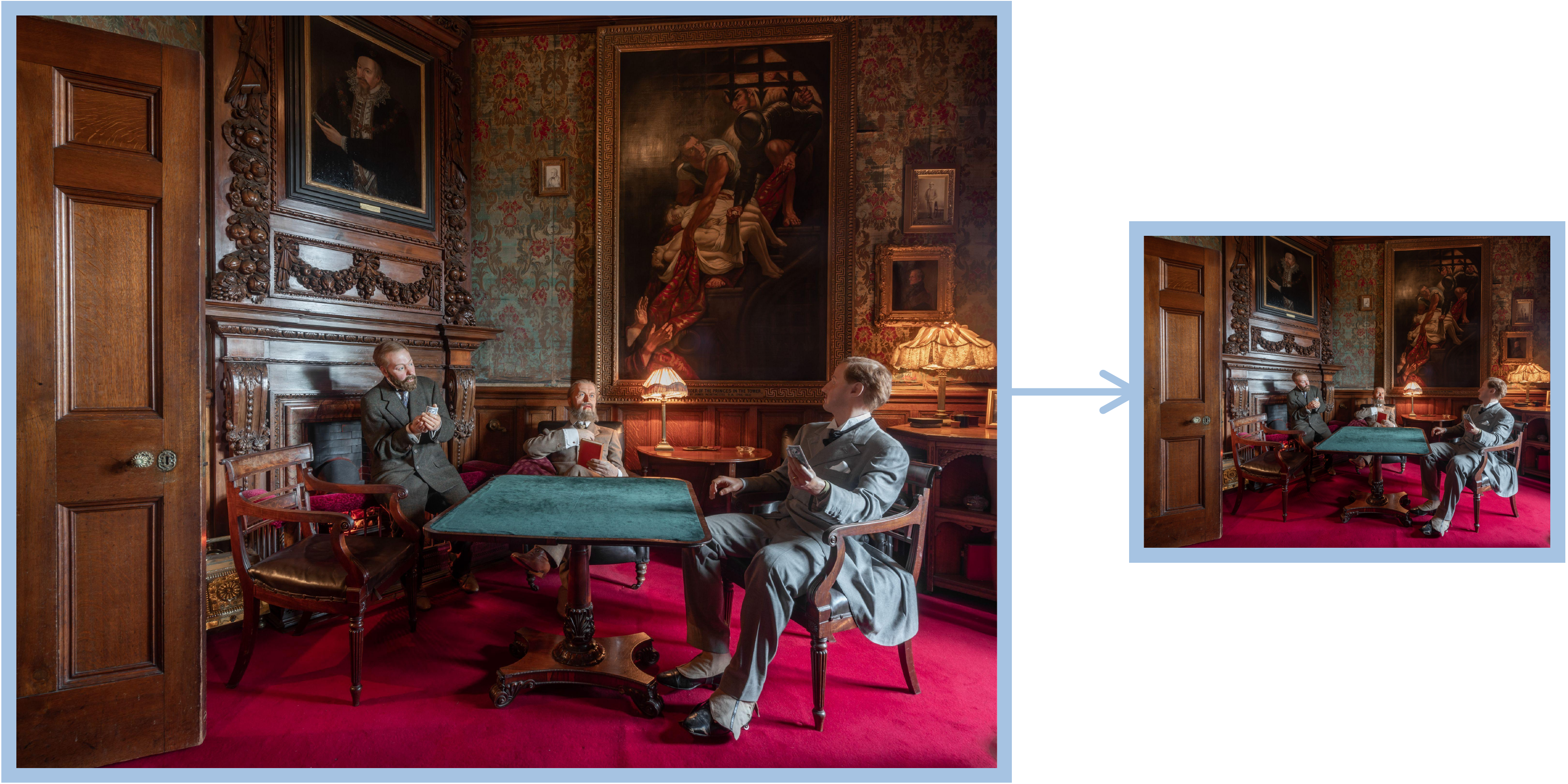}{
\textless think\textgreater
\textbackslash nThe image shows a luxurious interior, possibly from the late 19th or early 20th century, with dark wood, carved fireplaces, red carpet, and warm lighting. Large paintings, including religious and historical scenes, are on the walls. Three men in period suits sit around a green table. I observe a portrait painting with a man wearing a bowler hat. What is the color of his tie?
\textbackslash n\textless \textbackslash think\textgreater
\textbackslash n
\textbackslash n\textless tool\_call\textgreater
\textbackslash n\{"name": "crop\_image", "arguments": \{"bbox\_2d": [1474, 193, 1537, 285], "image\_index": 2\}\}
\textbackslash n\textless/tool\_call\textgreater
}{Turn 4}
\toolitem{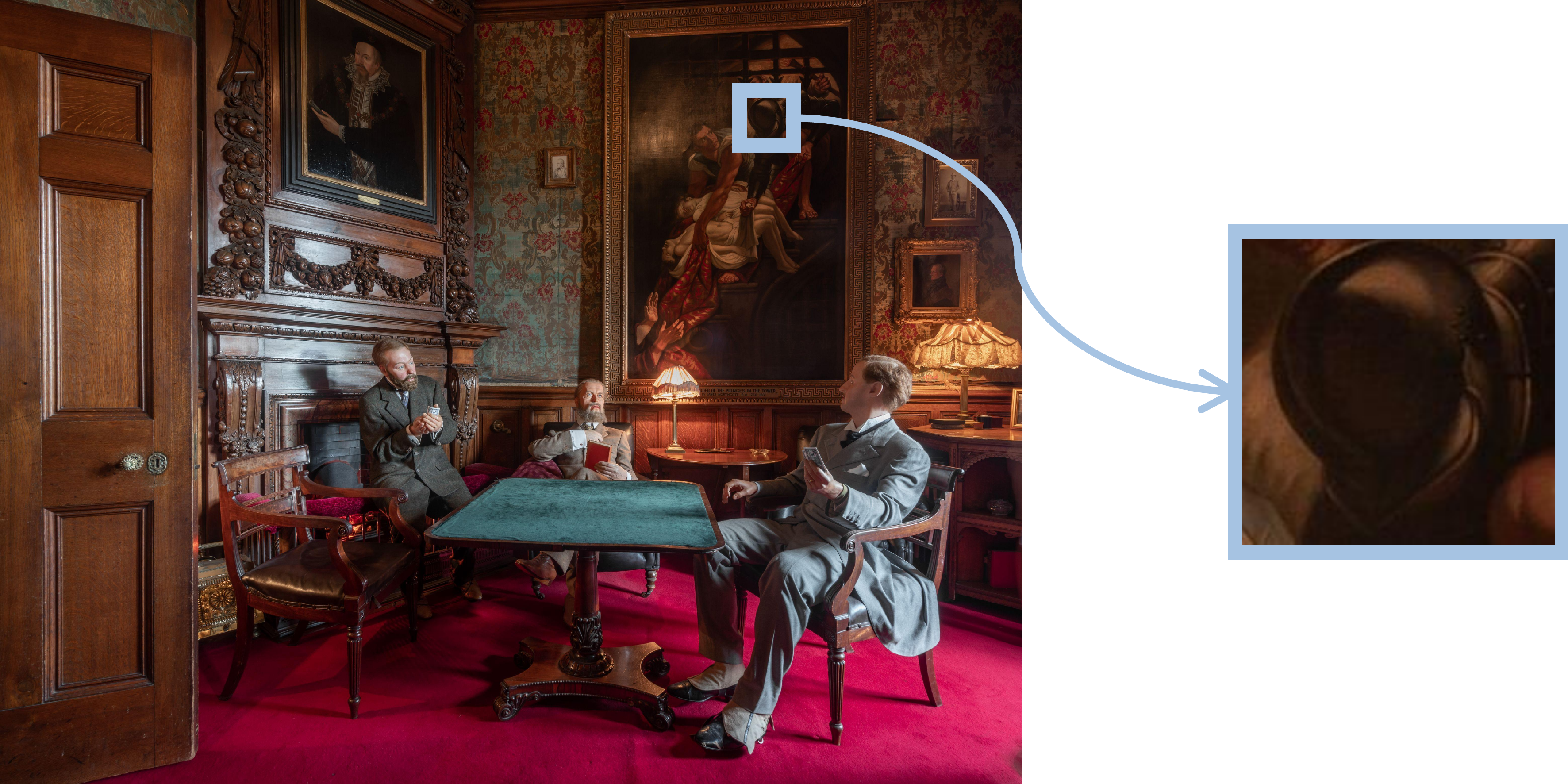}{
\textless think\textgreater
\textbackslash nThis matches the information about the man with the bowler hat, but I confirm that this man does not have a tie. How to find the man wearing a bowler hat?
\textbackslash n\textless \textbackslash think\textgreater
\textbackslash n
\textbackslash n\textless tool\_call\textgreater
\textbackslash n\{"name": "crop\_image", "arguments": \{"bbox\_2d": [1821, 306, 1940, 456], "image\_index": 2\}\}
\textbackslash n\textless/tool\_call\textgreater
}{Turn 5}
\toolitem{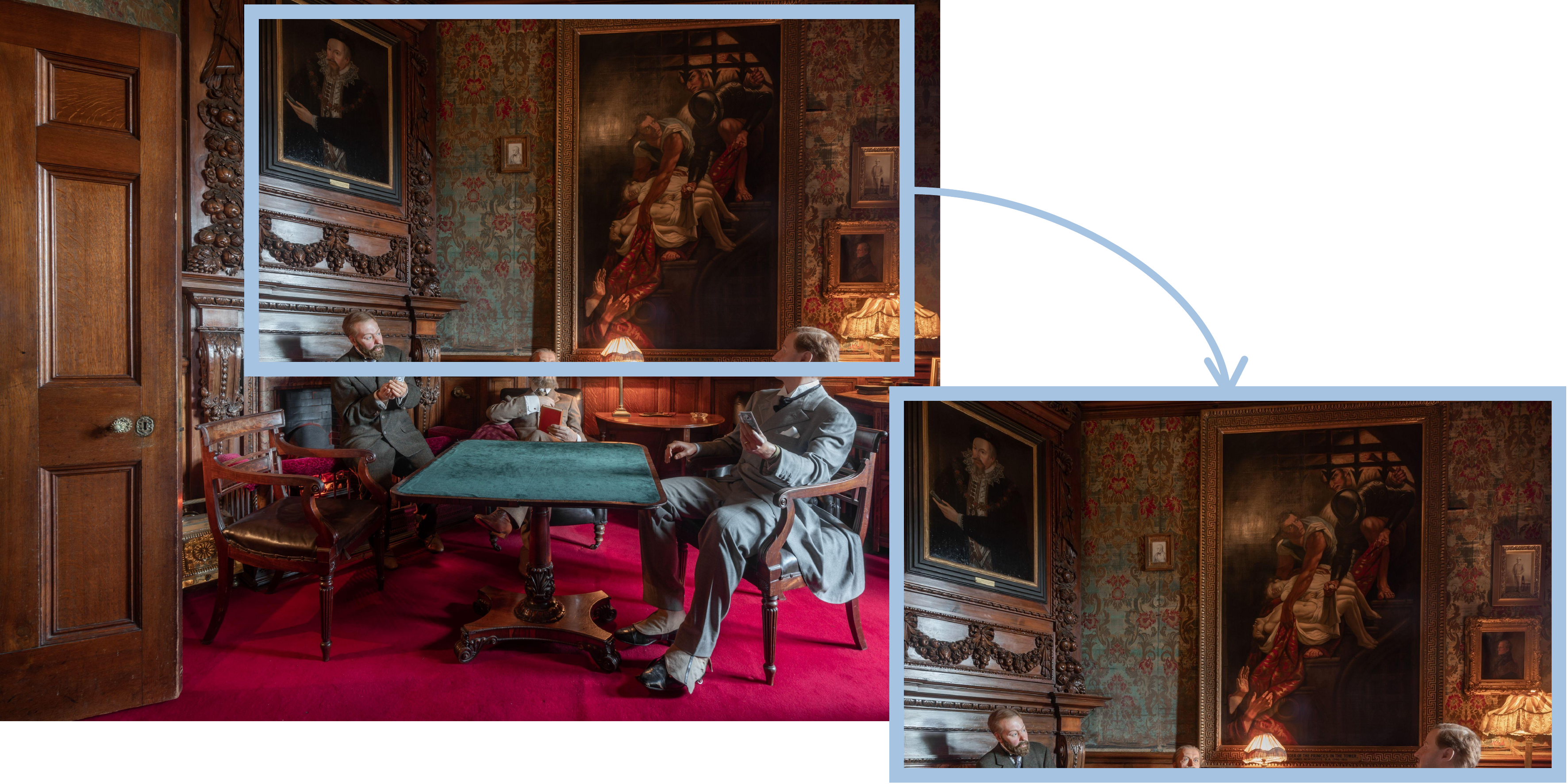}{
\textless think\textgreater
\textbackslash nNext, I check other portrait paintings. I find a black-and-white portrait of a man wearing a bowler hat and a tie. I note that the tie appears to be black. How to confirm the color of the tie in a black-and-white portrait?
\textbackslash n\textless \textbackslash think\textgreater
\textbackslash n
\textbackslash n\textless tool\_call\textgreater
\textbackslash n\{"name": "crop\_image", "arguments": \{"bbox\_2d": [552, 0, 1949, 643], "image\_index": 2\}\}
\textbackslash n\textless/tool\_call\textgreater
}{Turn 6: Reflecting}
\toolitem{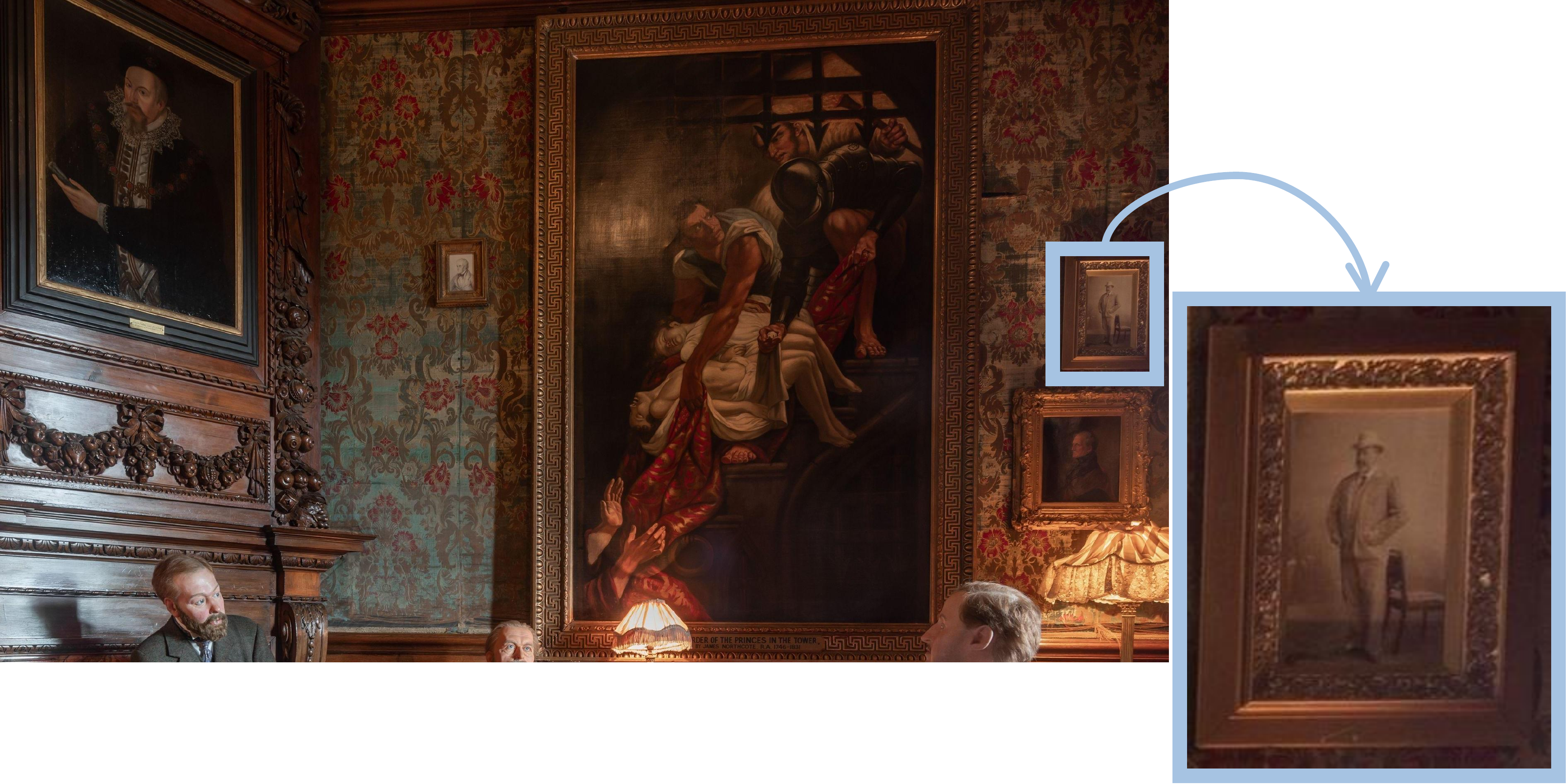}{
\textless think\textgreater
\textbackslash nI check all paintings. If the information does not specify the tie's color, substitute black as the assumed color. I write that the tie is black.
\textbackslash n\textless \textbackslash think\textgreater
\textbackslash n
\textbackslash n\textless answer\textgreater
\textbackslash nblack
\textbackslash n\textless/answer\textgreater
}{Turn 7}
\begin{tcolorbox}[
width=\linewidth,
colback=answer_right,
arc=1mm,
boxrule=0pt,
left=4pt,
right=4pt,
breakable,
]
\centering
\subsection*{\textsc{\textbf{ViRC}}-7B Final Answer: black (Correct)}
\end{tcolorbox}
\captionof{figure}{Inference Results on the VisualProbe Hard Benchmark.}
\label{fig:supp_exmaple_high_res}
}

\section{\texorpdfstring{\raggedright Dataset and Prompt Details}{Dataset and Prompt Details}}
\label{supp:dataset_and_prompt_details}

\subsection{\texorpdfstring{\raggedright Dataset Example}{Dataset Example}}
\label{supp:dataset_example}
\centering
\toolsection
{
A CRU-aligned reasoning path in the \textbf{CRUX} dataset
}
{
\toolitem{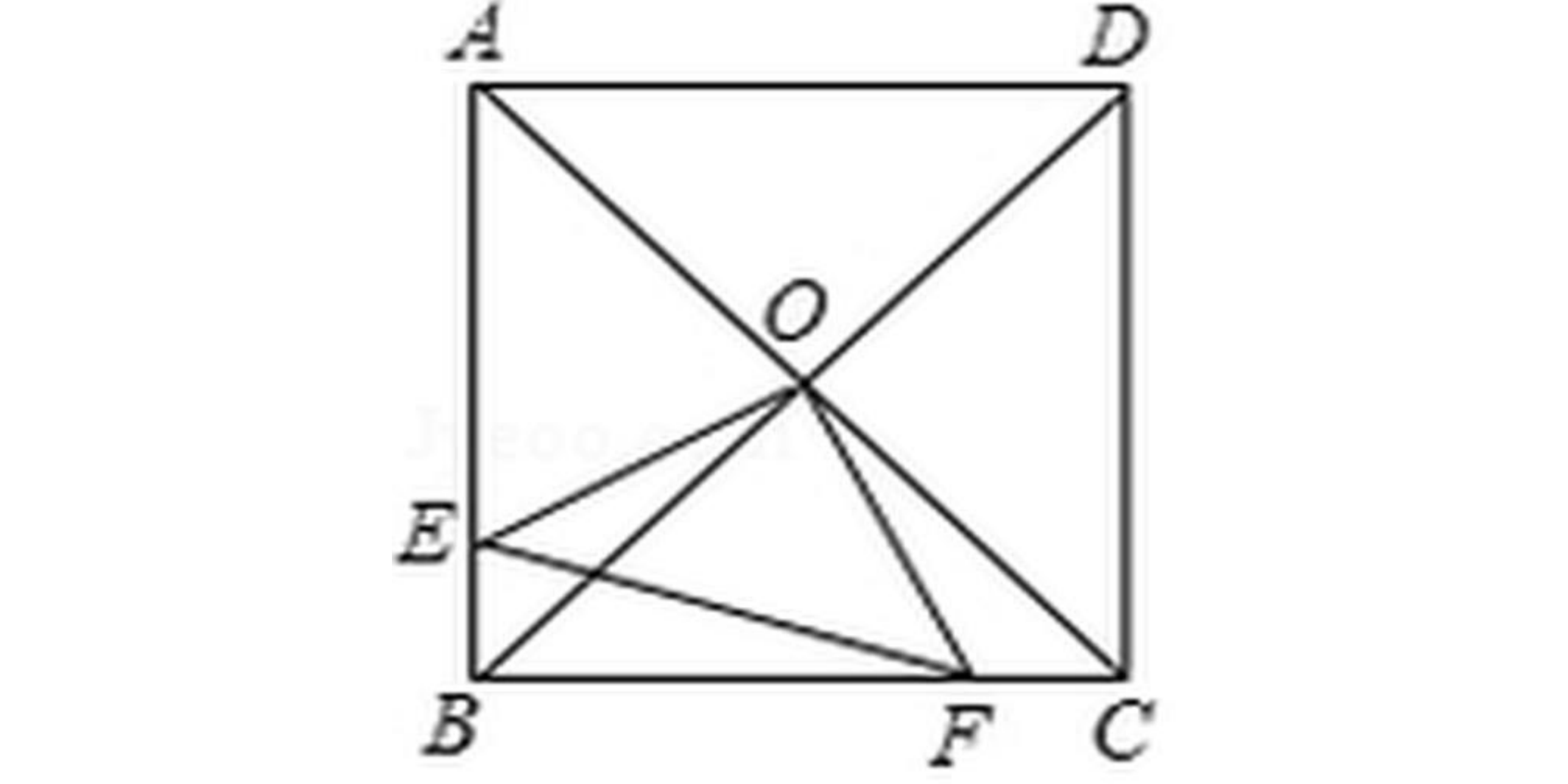}{
\textless image\textgreater Image 0: 504 x 504.\textbackslash nQuestion: Consider the square ABCD shown in the figure. The diagonals AC and BD intersect at point O. From point O, two perpendicular lines OE and OF are drawn to the sides AB and BC respectively. If the length of segment AE is 3.0 units and the length of segment CF is 1.0 unit, what is the value of EF?\textbackslash nChoices:\textbackslash nA: 2\textbackslash nB: \textbackslash sqrt\{10\}\textbackslash nC: 4\textbackslash nD: 2\textbackslash sqrt\{2\}\textbackslash n\textbackslash nLet’s think step by step. Call \text{**tool**} if needed, then answer. Format strictly as: \textless think\textgreater...\textless/think\textgreater \textless tool\_call\textgreater...\textless/tool\_call\textgreater{} (if tools needed) \textless answer\textgreater...\textless/answer\textgreater (if available).
}{Initial Query}
\toolitem{assets/figure/example_dataset/0.pdf}{
\textless think\textgreater\textbackslash nLet's think step by step. The image depicts a square ABCD. Diagonals AC and BD intersect at point O. Two perpendicular lines, OE and OF, are drawn from O to the sides AB and BC, respectively. AE measures 3.0 units and CF measures 1.0 unit. The problem involves a geometric shape (a square) and requires calculating the distance between two points (E and F) using given segment lengths related to the sides of the square. The key insight is recognizing and applying properties of perpendicular and intersecting lines within the square. - \text{**Step 1:**} Understand that diagonals bisect each other in a square, implying that O is equidistant from all four vertices of the square. - \text{**Step 2:**} Identify the right triangles involved: \textbackslash Delta AEO and \textbackslash Delta COF. Both are right triangles with known lengths of AE and CF, respectively. - \text{**Step 3:**} Recognize that OE and OF are perpendicular, forming the right triangle EOF. - \text{**Step 4:**} Apply the Pythagorean Theorem in the right triangle EOF to find EF. Use OE and OF as the perpendicular sides. - \text{**Step 5:**} Calculate OE and OF based on their perpendicular distances from the square's sides. How can the coordinates of points E and F be determined using the given lengths AE and CF within the square?\textbackslash n\textless/think\textgreater\textbackslash n\textbackslash n\textless tool\_call\textgreater{}\textbackslash n\{"name": "crop\_image", "arguments": \{"bbox\_2d": [0, 252, 336, 420], "image\_index": 0\}\}\textbackslash n\textless/tool\_call\textgreater
}{Turn 1}
\toolitem{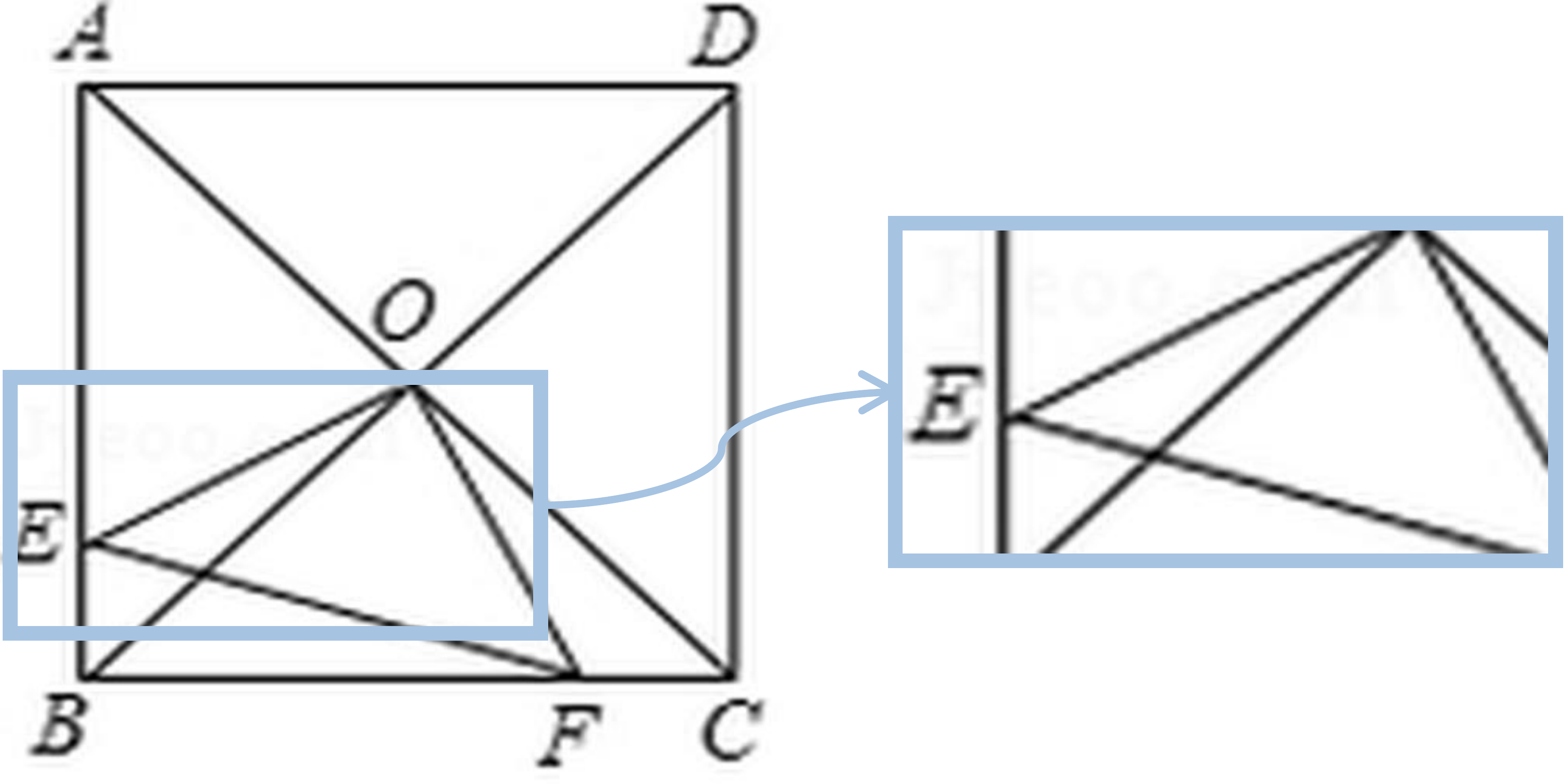}{
\textless think\textgreater\textbackslash nUnderstanding the Geometry: Given a square \$ABCD\$ with diagonals \$AC\$ and \$BD\$ intersecting at point \$O\$. Since \$O\$ is the intersection of the diagonals of a square, it is the midpoint of both diagonals, meaning \$O\$ divides each diagonal into two equal parts. Determine Coordinates: Place the square in a coordinate system with \$A = (0, s)\$, \$B = (s, s)\$, \$C = (s, 0)\$, and \$D = (0, 0)\$, where \$s\$ is the side length of the square. Then \$O\$ is at \$\textbackslash left(\textbackslash frac\{s\}\{2\}, \textbackslash frac\{s\}\{2\}\textbackslash right)\$. How can the coordinates of points E and F be determined using the given lengths AE and CF within the square? Perpendicular Lines OE and OF: From point \$O\$, two perpendicular lines \$OE \textbackslash perp AB\$ and \$OF \textbackslash perp BC\$. Lengths Given: \$AE = 3\$ and \$CF = 1\$. How can the positions of points E and F be determined using the given lengths AE and CF in terms of the square's side length?\textbackslash n\textless/think\textgreater\textbackslash n\textbackslash n\textless tool\_call\textgreater{}\textbackslash n\{"name": "crop\_image", "arguments": \{"bbox\_2d": [0, 0, 336, 336], "image\_index": 0\}\}\textbackslash n\textless/tool\_call\textgreater
}{Turn 2}

\begin{tcolorbox}[
width=\linewidth,
colback=box_section_back,
arc=0mm,
boxrule=0pt,
left=4pt,
right=4pt,
breakable,
halign=center,
valign=center,
height=7em,
]
(continued on next page)
\end{tcolorbox}

\toolitem{assets/figure/example\_dataset/2.pdf}{
\textless think\textgreater\textbackslash nLengths of OE and OF: Since \$OE \textbackslash perp AB\$, \$E\$ lies on \$AB\$ and thus has coordinates \$(x, s)\$. Given \$AE = 3\$, \$E\$ is at \$(3, s)\$. Therefore, \$OE = \textbackslash frac\{s\}\{2\} - 3\$. Similarly, since \$OF \textbackslash perp BC\$, \$F\$ lies on \$BC\$ and thus has coordinates \$(s, y)\$. Given \$CF = 1\$, \$F\$ is at \$(s, 1)\$. Therefore, \$OF = \textbackslash frac\{s\}\{2\} - 1\$. Wait... Based on this image, my current step seems to be incorrect. Let's try a different approach. Image 0 is too big. I need to scale it down for a better view.\textbackslash n\textless/think\textgreater\textbackslash n\textbackslash n\textless tool\_call\textgreater{}\textbackslash n\{"name": "scale\_image", "arguments": \{"scale\_factor": 0.25, "image\_index": 0\}\}\}\textbackslash n\textless/tool\_call\textgreater
}{Turn 3}
\toolitem{assets/figure/example\_dataset/3.pdf}{
\textless think\textgreater\textbackslash nLet me review the previous steps based on this image. The image depicts a square ABCD. Diagonals AC and BD intersect at point O. Two perpendicular lines, OE and OF, are drawn from O to the sides AB and BC, respectively. AE measures 3.0 units and CF measures 1.0 unit. The problem involves a geometric shape (a square) and requires calculating the distance between two points (E and F) using given segment lengths related to the sides of the square. The key insight is recognizing and applying properties of perpendicular and intersecting lines within the square. - \text{**Step 1:**} Understand that diagonals bisect each other in a square, implying that O is equidistant from all four vertices of the square. - \text{**Step 2:**} Identify the right triangles involved: \textbackslash Delta AEO and \textbackslash Delta COF. Both are right triangles with known lengths of AE and CF, respectively. - \text{**Step 3:**} Recognize that OE and OF are perpendicular, forming the right triangle EOF. - \text{**Step 4:**} Apply the Pythagorean Theorem in the right triangle EOF to find EF. Use OE and OF as the perpendicular sides. - \text{**Step 5:**} Calculate OE and OF based on their perpendicular distances from the square's sides. How can the coordinates of points E and F be determined using the given lengths AE and CF within the square?\textbackslash n\textless/think\textgreater\textbackslash n\textbackslash n\textless tool\_call\textgreater{}\textbackslash n\{"name": "crop\_image", "arguments": \{"bbox\_2d": [0, 63, 84, 105], "image\_index": 3\}\}\textbackslash n\textless/tool\_call\textgreater
}{Turn 4}
\toolitem{assets/figure/example\_dataset/4.pdf}{
\textless think\textgreater\textbackslash nIdentify Key Properties: \$ABCD\$ is a square. Diagonals \$AC\$ and \$BD\$ intersect at point \$O\$. \$O\$ is the midpoint of both diagonals, so \$AO = OC\$ and \$BO = OD\$. \$OE \textbackslash perp AB\$ and \$OF \textbackslash perp BC\$. Calculate Side Length of the Square: Let the side length of the square be \$s\$. Since \$O\$ is the midpoint of the diagonals, \$AO = OC = BO = OD = \textbackslash frac\{s\textbackslash sqrt\{2\}\}\{2\}\$. How can the coordinates of points E and F be determined using the given lengths AE and CF within the square? Given Values: \$AE = 3\$ units. \$CF = 1\$ unit. Determine \$BE\$ and \$BF\$: Since \$AE = 3\$ and \$AB = s\$, then \$BE = s - 3\$. Since \$CF = 1\$ and \$BC = s\$, then \$BF = s - 1\$. How can the positions of points E and F be determined using the given lengths AE and CF in terms of the square's side length?\textbackslash n\textless/think\textgreater\textbackslash n\textbackslash n\textless tool\_call\textgreater{}\textbackslash n\{"name": "crop\_image", "arguments": \{"bbox\_2d": [0, 0, 84, 84], "image\_index": 3\}\}\textbackslash n\textless/tool\_call\textgreater
}{Turn 5}
\toolitem{assets/figure/example\_dataset/5.pdf}{
\textless think\textgreater\textbackslash nUse Right Triangles \$AEO\$ and \$COF\$: In \$\textbackslash triangle AEO\$, \$AE = 3\$ and \$OE\$ is perpendicular to \$AB\$. So, \$OE = \textbackslash frac\{s\}\{2\}\$. In \$\textbackslash triangle COF\$, \$CF = 1\$ and \$OF\$ is perpendicular to \$BC\$. So, \$OF = \textbackslash frac\{s\}\{2\}\$. Wait... Based on this image, my current step seems to be incorrect. Let's try a different approach.\textbackslash n\textless/think\textgreater\textbackslash n\textbackslash n\textless tool\_call\textgreater{}\textbackslash n\{"name": "display\_image", "arguments": \{"image\_index": 4\}\}\textbackslash n\textless/tool\_call\textgreater
}{Turn 6}

\begin{tcolorbox}[
width=\linewidth,
colback=box_section_back,
arc=0mm,
boxrule=0pt,
left=4pt,
right=4pt,
breakable,
halign=center,
valign=center,
height=5em,
]
(continued on next page)
\end{tcolorbox}

\toolitem{assets/figure/example\_dataset/6.pdf}{
\textless think\textgreater\textbackslash nI recognize that point O is the center of square ABCD, so its coordinates are halfway between the vertices. I assign a coordinate system where vertex A is at \$(0,0)\$, allowing me to express coordinates for O, E, and F based on the square's side length. How can the coordinates of points E and F be determined using the given lengths AE and CF within the square? Since OE and OF are perpendicular lines from O to sides AB and BC respectively, I note their directions must satisfy the perpendicularity condition. I denote the square's side length as \$s\$, so coordinates become: O is at \$(\textbackslash frac\{s\}\{2\}, \textbackslash frac\{s\}\{2\})\$, E is on AB with AE = 3, placing it at \$(3, 0)\$, and F is on BC with CF = 1, placing it at \$(s, s-1)\$. How can the positions of points E and F be determined using the given lengths AE and CF in terms of the square's side length?\textbackslash n\textless/think\textgreater\textbackslash n\textbackslash n\textless tool\_call\textgreater{}\textbackslash n\{"name": "display\_image", "arguments": \{"image\_index": 5\}\}\textbackslash n\textless/tool\_call\textgreater
}{Turn 7}
\toolitem{assets/figure/example\_dataset/7.pdf}{
\textless think\textgreater\textbackslash nI represent vectors OE and OF using coordinates. Vector OE is from O to E: \$(3 - \textbackslash frac\{s\}\{2\}, -\textbackslash frac\{s\}\{2\})\$, and vector OF is from O to F: \$(\textbackslash frac\{s\}\{2\}, \textbackslash frac\{s\}\{2\} - 1)\$. Since OE and OF are perpendicular, their dot product equals zero. How can I represent the coordinates of points E and F based on the given side lengths AE and CF, and use the perpendicularity of OE and OF to find the side length of the square?\textbackslash n\textless/think\textgreater\textbackslash n\textbackslash n\textless tool\_call\textgreater{}\textbackslash n\{"name": "crop\_image", "arguments": \{"bbox\_2d": [63, 42, 126, 126], "image\_index": 3\}\}\textbackslash n\textless/tool\_call\textgreater
}{Turn 8}
\toolitem{assets/figure/example\_dataset/8.pdf}{
\textless think\textgreater\textbackslash nSetting up the equation for the dot product: \$(3 - \textbackslash frac\{s\}\{2\})(\textbackslash frac\{s\}\{2\}) + (-\textbackslash frac\{s\}\{2\})(\textbackslash frac\{s\}\{2\} - 1) = 0\$. Solving this yields \$s = 4\$, determining the square's side length. How do the given lengths AE and CF relate to the side length of the square?\textbackslash n\textless/think\textgreater\textbackslash n\textbackslash n\textless tool\_call\textgreater{}\textbackslash n\{"name": "crop\_image", "arguments": \{"bbox\_2d": [21, 84, 105, 105], "image\_index": 3\}\}\textbackslash n\textless/tool\_call\textgreater
}{Turn 9}
\toolitem{assets/figure/example\_dataset/9.pdf}{
\textless think\textgreater
\textbackslash nWith \$s = 4\$, coordinates of E are \$(3,0)\$ and F are \$(4,3)\$. Applying the distance formula, \$EF = \textbackslash sqrt{(4 - 3)\^{}2 + (3 - 0)\^{}2} = \textbackslash sqrt{1 + 9} = \textbackslash sqrt{10}\$. Thus, EF is \$\textbackslash sqrt{10}\$.
\textbackslash n\textless/think\textgreater
\textbackslash n\textbackslash n\textless answer\textgreater
\textbackslash nB
\textbackslash n\textless/answer\textgreater
}{Turn 10}
\captionof{figure}{A CRU-aligned reasoning path in the \textbf{CRUX} dataset.}
\label{fig:supp_exmaple_dataset}
}

\subsection{\texorpdfstring{\raggedright Prompts for Data Curation}{Prompts for Data Curation}}
\label{supp:prompts_for_data_curation}

\centering
\begin{tcolorbox}[
  width=\linewidth,
  colback=box_section_back,
  colbacktitle=box_section_title_back,
  coltitle=black,
  arc=1mm,
  boxrule=0pt,
  title={\parbox{\linewidth}{\subsection*{
Zero-shot CoT Prompt Template (Sampling Stage)
  }}
  },
  left=4pt,
  right=4pt,
  breakable,
  ]
\verb|{image}|\\
\verb|{question}| Let’s think step by step.
\captionof{figure}{Zero-shot CoT Prompt Template. Given this prompt, a reasoning path is sampled from the model in a zero-shot CoT manner based on the input image–text pair.}
\label{fig:supp_prompt_sampling_sample}
\end{tcolorbox}

\centering
\begin{tcolorbox}[
  width=\linewidth,
  colback=box_section_back,
  colbacktitle=box_section_title_back,
  coltitle=black,
  arc=1mm,
  boxrule=0pt,
  title={\parbox{\linewidth}{\subsection*{
Answer Validation Prompt Template (Sampling Stage)
  }}
  },
  left=4pt,
  right=4pt,
  breakable,
  ]
  Below are two answers to the same question: [Question]. [Standard Answer] is the correct answer, and [Model Answer] is from a model's output. Compare them.\\
  If [Model Answer] has the same meaning as [Standard Answer], even if expressed differently, they are consistent.\\
  The model's output will contain the answer, regardless of its certainty. Just focus on the consistency of the answer, not the solution process. If they are consistent, Judement is 1; if they are different, Judement is 0. Just output Judgement as \textbackslash boxed\{0\} or \textbackslash boxed\{1\}.

  \vspace{1em}

  \text{[Question]}: Who is wearing pants?\\
  \text{[Standard Answer]}: A. The boy is wearing pants.\\
  \text{[Model Answer]}: C. The girl in the picture is wearing pants.\\
  \text{Judgment}: \textbackslash boxed\{0\}

  \vspace{1em}

  \text{[Question]}: Is the man phone both blue and closed?\\
  \text{[Standard Answer]}: A. Yes, the man phone is both blue and closed.\\
  \text{[Model Answer]}: No.\\
  \text{Judgment}: \textbackslash boxed\{0\}

  \vspace{1em}

  \text{[Question]}: What color is the towel in the center of the picture?\\
  \text{[Standard Answer]}: A. The towel in the center of the picture is blue.\\
  \text{[Model Answer]}: The towel in the center of the picture is pink.\\
  \text{Judgment}: \textbackslash boxed\{0\}

  \vspace{1em}

  \text{[Question]}: Is the countertop tan or blue?\\
  \text{[Standard Answer]}: A. The countertop is tan.\\
  \text{[Model Answer]}: tan\\
  \text{Judgment}: \textbackslash boxed\{1\}

  \vspace{1em}

  \text{[Question]}: On which side of the picture is the barrier?\\
  \text{[Standard Answer]}: A. The barrier is on the left side of the picture.\\
  \text{[Model Answer]}: A\\
  \text{Judgment}: \textbackslash boxed\{1\}

  \vspace{1em}

  \text{[Question]}: Is the kite brown and large?\\
  \text{[Standard Answer]}: A. Yes, the kite is brown and large.\\
  \text{[Model Answer]}: Yes\\
  \text{Judgment}: \textbackslash boxed\{1\}

  \vspace{1em}

  \text{[Question]}: Are the spots on a giraffe?\\
  \text{[Standard Answer]}: A. No, the spots are on a banana.\\
  \text{[Model Answer]}: no\\
  \text{Judgment}: \textbackslash boxed\{1\}

  \vspace{1em}

  \text{[Question]}: \verb|{question}|\\
  \text{[Standard Answer]}:  \verb|{answer_gt}|\\
  \text{[Model Answer]}:  \verb|{answer_pred}|\\
  \text{Judgment}:
\captionof{figure}{Answer Verification Prompt Template. The model makes a judgment based on the question, the correct answer, and the model's predicted answer.}
\label{fig:supp_prompt_sampling_val_ans}
\end{tcolorbox}

\begin{tcolorbox}[
  width=\linewidth,
  colback=box_section_back,
  colbacktitle=box_section_title_back,
  coltitle=black,
  arc=1mm,
  boxrule=0pt,
  title={\parbox{\linewidth}{\subsection*{
Reasoning Path Decomposition Prompt Template (Mapping Stage)
  }}
  },
  left=4pt,
  right=4pt,
  breakable,
  ]
  \verb|{question}|\\
  The above is a question image for mathematical geometry problems.
  
  \vspace{1em}
  
  \verb|{cot}|\\
  The above are the steps for solving mathematical geometry problems, and I guarantee that the steps and results are correct.
  
  \vspace{1em}
  
  You are an expert in structuring and breaking down information. Your task is to convert the provided solution into a structured, step-by-step thinking sequence. This sequence will serve as a student's review notes, so it is crucial to preserve the complete thought process.
  
  \vspace{1em}
  
  Please respond in JSON format. The JSON object should use sequential numbers (as strings) for keys. The value for each key must be a dictionary containing two fields:
  \begin{itemize}[left=1em]
    \item think: A string explaining the thought process for the current step.
    \item object: A string naming the main geometric figure or concept to focus on for this step.
  \end{itemize}
  
  \vspace{1em}
  
  Your response format must be:
  \begin{lstlisting}[basicstyle=\ttfamily\small,breaklines=true,aboveskip=0pt,belowskip=0pt]
{
    "1": {
    "think": "The textual thought process for this step.",
    "object": "The name of the main geometric figure."
    }
}
  \end{lstlisting}
\captionof{figure}{Decompose Reasoning Path Prompt Template. The model provides the individual steps of the reasoning path based on the image content, the question, and the overall reasoning process, while also identifying the main focus objects in the image for each step.}
\label{fig:supp_prompt_mapping_decomposition}
\end{tcolorbox}

\centering
\begin{tcolorbox}[
  width=\linewidth,
  colback=box_section_back,
  colbacktitle=box_section_title_back,
  coltitle=black,
  arc=1mm,
  boxrule=0pt,
  title={\parbox{\linewidth}{\subsection*{
Proposition-Level Reasoning Alignment Prompt Template (Mapping Stage)
  }}
  },
  left=4pt,
  right=4pt,
  breakable,
  ]
    You are an AI expert specializing in mapping reasoning processes and identifying logical errors.

    \vspace{1em}

    \begin{itemize}
        \item Task: Your task is to analyze two step-by-step procedures, \text{[wrong]} and \text{[correct]}, and perform two actions.
        \begin{itemize}[label=$\circ$]
            \item Step Mapping: For each step in \text{[wrong]}, find the step in \text{[correct]} that shares the same fundamental goal.
            \item First Error Identification: Pinpoint the very first step in the \text{[wrong]} process where a calculation or logical error occurs.
        \end{itemize}
        \item Input: Keys are step IDs and values are step descriptions.
        \begin{itemize}[label=$\circ$]
            \item \text{[wrong]}: A JSON object representing an inference process with incorrect reasoning steps leading to incorrect results.
            \item \text{[correct]}: A JSON object representing a standardized and correct inference process.
        \end{itemize}
        \item Core Criteria:
        \begin{itemize}[label=$\circ$]
            \item Mapping Criterion: The mapping must be based on semantic intent. Match steps by their underlying purpose, even if the methods or results in \text{[wrong]} are incorrect.
            \item First Error Identification Criterion: To find the ``wrong\_step'', you must trace the error back to its absolute origin. Use the following procedure:
            \begin{enumerate}
                \item Start from the last step and work backward.
                \item For each step, ask yourself: ``Is the logic or calculation within this step flawed *on its own terms*, OR is it only incorrect because it relies on a flawed result from a *previous* step?''
                \item If the step is only incorrect because it uses a previous error, then this is a **consequential error**, not the first error. Continue moving backward.
                \item The ``wrong\_step'' is the first step you encounter (moving backward) that contains a **foundational error**, a mistake in logic, a misapplied formula, or a false statement that is not dependent on any prior errors. This is the root cause.
            \end{enumerate}
        \end{itemize}
        \item Output: Your response must be in two parts: \text{[analysis]} and \text{[output]}.
        \begin{itemize}[label=$\circ$]
            \item Provide your step-by-step reasoning directly as text (not in a code block). For each step in \text{[wrong]}, identify its mapped step in \text{[correct]} and justify the mapping.
            \item A single JSON object containing:
            \begin{itemize}[label=$\circ$]
                \item All keys from \text{[wrong]}, with values being their mapped keys from \text{[correct]}.
                \item An additional key-value pair, ``wrong\_step'': ``ID'', identifying the first step in \text{[wrong]} that contains an error.
            \end{itemize}
        \end{itemize}
    \end{itemize}

    \vspace{1em}

    \text{[correct]}
    \begin{lstlisting}[basicstyle=\ttfamily\small,breaklines=true,aboveskip=0pt,belowskip=0pt]
{
    "1": "I note that triangle ABC is isosceles with AB = AC and angle BAC = 100 \degree, so angles at vertices B and C (angles ABC and ACB) must be equal.",
    "2": "To find angles ABC and ACB, I use the triangle angle sum formula: angle BAC + angle ABC + angle ACB = 180 \degree. Letting each base angle be \(x\), the equation becomes \(100 \degree + 2x = 180 \degree\). Solving gives \(x = 40 \degree\), so angles ABC and ACB each measure 40 \degree.",
    "3": "The perpendicular bisector of AB intersects AB at midpoint D (so AD = DB) and BC at E. Since it is a perpendicular bisector, the line DE forms a right angle with AB at D, making angle ADE = 90 \degree.",
    "4": "Assuming AD bisects angle BAC (dividing it into two 50 \degree angles), angle BAE is calculated by considering triangle ABE. Since angle BAC is split into 50 \degree and angle at B is 40 \degree, angle BAE combines these relationships to yield theta = 40 \degree."
}
    \end{lstlisting}

    \vspace{1em}

    \text{[wrong]}
    \begin{lstlisting}[basicstyle=\ttfamily\small,breaklines=true,aboveskip=0pt,belowskip=0pt]
{
    "1": "Analyze the Isosceles Triangle \( \triangle ABC \): Since \( AB = AC \), \( \triangle ABC \) is an isosceles triangle. The base angles \( \angle ABC \) and \( \angle ACB \) are equal. The vertex angle \( \angle BAC = 100^\circ \).",
    "2": "Using the fact that the sum of angles in any triangle is \( 180^\circ \): \( \angle ABC + \angle ACB + \angle BAC = 180^\circ \). Substituting \( \angle BAC = 100^\circ \): \( \angle ABC + \angle ACB + 100^\circ = 180^\circ \). Therefore, \( \angle ABC + \angle ACB = 80^\circ \). Since \( \angle ABC = \angle ACB \): \( 2 \angle ABC = 80^\circ \implies \angle ABC = \angle ACB = 40^\circ \).",
    "3": "Analyze the Perpendicular Bisector: The perpendicular bisector of \( AB \) intersects \( AB \) at \( D \) and \( BC \) at \( E \). By definition, \( D \) is the midpoint of \( AB \), so \( AD = DB \). The perpendicular bisector implies that \( \angle ADE = 90^\circ \).",
    "4": "Determine \( \angle BAE \): Since \( DE \) is the perpendicular bisector of \( AB \), \( E \) is equidistant from \( A \) and \( B \). Thus, \( AE = BE \) and \( \triangle ABE \) is isosceles with \( AE = BE \). In \( \triangle ABE \), since \( \angle ABE = 40^\circ \) (from earlier calculation): \( \angle BAE = \angle AEB \). Sum of angles in \( \triangle ABE \): \( \angle BAE + \angle AEB + \angle ABE = 180^\circ \). Since \( \angle BAE = \angle AEB \): \( 2 \angle BAE + 40^\circ = 180^\circ \). Solving for \( \angle BAE \): \( 2 \angle BAE = 140^\circ \implies \angle BAE = 70^\circ / 2 = 40^\circ \).",
    "5": "Thus, the measure of angle \( \angle BAE \) is \( \boxed{40^\circ} \)."
}
    \end{lstlisting}

    \vspace{1em}

    \text{[analysis]}\\
    **Step Mapping**\\
    wrong\_1 maps to correct\_1: Both steps identify triangle ABC as isosceles and state that its base angles are equal.\\
    wrong\_2 maps to correct\_2: Both steps use the triangle angle sum to calculate the base angles as 40°.\\
    wrong\_3 maps to correct\_3: Both steps define the properties of the perpendicular bisector of AB.\\
    wrong\_4 maps to correct\_4: Both steps share the ultimate goal of calculating the measure of angle BAE.\\
    wrong\_5 maps to correct\_4: This step is a concluding statement of the result calculated in the previous step, a goal which is integrated into correct\_4.\\

    \vspace{1em}

    **First Error Identification**\\
    The first foundational error occurs in \text{[wrong]} step 4.\\
    The reasoning begins correctly by stating that since point E is on the perpendicular bisector of AB, triangle ABE is isosceles with AE = BE. However, the step then makes a logical error by concluding that angle BAE must equal angle AEB. In an isosceles triangle, the angles opposite the equal sides are equal. The sides AE and BE are opposite angles ABE and BAE, respectively. Therefore, the correct relationship is angle BAE = angle ABE. The error in step 4 is a misapplication of the isosceles triangle theorem.

    \vspace{1em}

    This mistake is foundational because it is not based on any prior error. Steps 1, 2, and 3 are all correct. The subsequent calculations in step 4 are incorrect because they are built upon this initial logical flaw.

    \vspace{1em}

    \text{[output]}
    \begin{lstlisting}[basicstyle=\ttfamily\small,breaklines=true,aboveskip=0pt,belowskip=0pt]
{
    "1": "1",
    "2": "2",
    "3": "3",
    "4": "4",
    "5": "4",
    "wrong_step": "4"
}
    \end{lstlisting}

    \vspace{1em}

    \text{[correct]}:\verb|{correct}|\\
    \text{[wrong]}: \verb|{wrong}|

    \vspace{1em}

    \text{[analysis]}\\
    **Step Mapping**
\captionof{figure}{Reasoning Path Alignment Prompt Template.The model aligns the steps in the correct reasoning path with the corresponding steps in the incorrect reasoning path, providing detailed mappings for each step.}
\label{fig:supp_prompt_mapping_cru}
\end{tcolorbox}

\centering
\begin{tcolorbox}[
  width=\linewidth,
  colback=box_section_back,
  colbacktitle=box_section_title_back,
  coltitle=black,
  arc=1mm,
  boxrule=0pt,
  title={\parbox{\linewidth}{\subsection*{
Planning Pattern Generation Prompt Template (Grounding Stage)
  }}
  },
  left=4pt,
  right=4pt,
  breakable,
  ]
    \verb|{image}|\\
    **Math Question**: \verb|{question}|\\
    **Solution**: \verb|{solution}|

    \vspace{1em}

    You are an expert math assistant. Your task is to analyze the provided image, the math query and the solution to create a strategic guide for solving the problem. Based on the image and query, perform the following two steps:

    \begin{itemize}[left=1em]
        \item **Generate a Caption (Problem Setup)**:
            \begin{itemize}[label=$\circ$]
                \item A string explaining the thought process for the current step.
            \end{itemize}
        \item **Generate a Rationale (Applied Principle)**:
            \begin{itemize}[label=$\circ$]
                \item Identify the key mathematical principle (e.g., a theorem, property, or formula).
                \item Crucially, describe **how this principle should be applied to the specific elements (points, lines, angles) in the diagram** to solve the query.
                \item **This should be a direct instruction for the first logical step.**
                \item **Do not** perform the calculation or state the final numerical answer.
            \end{itemize}
    \end{itemize}

    \vspace{1em}

    **Output Format**:\\
    Strictly adhere to the following JSON format. The rationale should be a specific, actionable statement.
    \begin{lstlisting}[basicstyle=\ttfamily\small,breaklines=true,aboveskip=0pt,belowskip=0pt]
{
    "caption": "A short, descriptive sentence about the problem setup.",
    "rationale": "A statement explaining how to apply the key principle to the diagram's elements, e.g., 'Apply the Pythagorean theorem to the right triangle ABC to find the length of AC' or 'Use the property that angles on a straight line sum to 180 degrees for line segment BCD'."
}
    \end{lstlisting}
\captionof{figure}{Planning Pattern Prompt Template. The model generates a description for the image and provides the rationale necessary to solve the problem based on the image, the given question, and the solution process.}
\label{fig:supp_prompt_grounding_planning}
\end{tcolorbox}

\centering
\begin{tcolorbox}[
  width=\linewidth,
  colback=box_section_back,
  colbacktitle=box_section_title_back,
  coltitle=black,
  arc=1mm,
  boxrule=0pt,
  title={\parbox{\linewidth}{\subsection*{
Guiding Question Generation Prompt Template (Grounding Stage)
  }}
  },
  left=4pt,
  right=4pt,
  breakable,
  ]
    You are an experienced math teacher guiding students to solve a math problem \text{[question]} step by step \text{[steps]}. The problem includes an image, described in words \text{[caption]}, and follows a clear solution logic \text{[rationale]}. When a student gets stuck on a particular step, you should provide a concise first-person question [guide] based on that step to help them think of the next step. Put your guide within \textbackslash boxed\{\}.

    \vspace{1em}

    **Requirements**:
    \begin{itemize}[left=1em]
        \item The question must be based on the current step, logically clear, and encourage the student to consider the next move.
        \item Avoid directly stating the next step; prompt independent reasoning instead.
        \item Be concise, with smooth integration into the context.
    \end{itemize}

    \vspace{1em}
    \text{[question]}: A number plus half of it equals 15. What is this number?\\
    \text{[caption]}: A bar chart with a long bar on the left and a short bar on the right, with the sum pointing to 15\\
    \text{[rationale]}: Solving linear equations with one variable\\
    \text{[steps]}:
    \begin{enumerate}[left=1em]
        \item Let this number be \(x \), and half of it is \(\frac{x}{2} \).
        \item \( x + \frac{x}{2} = 15 \)
        \item Merge like terms. \(\frac{2x}{2} + \frac{x}{2} = 15 \)
        \item Solving for it gives. \(x = 10 \).
    \end{enumerate}
    \text{[confused]}: 1 to 2 \\
    \text{[guide]}: \textbackslash boxed\{How can I express "adding up equals 15" in mathematical terms?\}

    \vspace{1em}

    \text{[question]}: \verb|{question}|\\
    \text{[caption]}: \verb|{caption}|\\
    \text{[rationale]}: \verb|{rationale}|\\
    \text{[steps]}: \verb|{steps}|\\
    \text{[confused]}: \verb|{step_index}| to \verb|{step_index_next}|\\
    \text{[guide]}:
\captionof{figure}{Guiding Question Prompt Template. The model generates a guiding question for the next reasoning step based on the question, the image caption, the rationale for solving the problem, the reasoning steps, and the indices of the two adjacent steps.}
\label{fig:supp_prompt_grounding_question}
\end{tcolorbox}

\centering
\begin{tcolorbox}[
  width=\linewidth,
  colback=box_section_back,
  colbacktitle=box_section_title_back,
  coltitle=black,
  arc=1mm,
  boxrule=0pt,
  title={\parbox{\linewidth}{\subsection*{
Bounding Box Localization Prompt Template (Grounding Stage)
  }}
  },
  left=4pt,
  right=4pt,
  breakable,
  ]
    \verb|{image}|\\
    Your task is to analyze an image containing a mathematical problem and perform the following three steps:

    \vspace{1em}

    \begin{enumerate}[left=1em]
        \item **Structure Recognition**: Identify and locate the specified geometric structure \text{[structure]} in the image.
        \item **Auxiliary Recognition**: Determine which text elements present in the image are necessary for identifying the given geometric structure \text{[structure]}.
        \item **Text Detection**: Locate the bounding boxes of the text identified in step 2.
    \end{enumerate}

    \vspace{1em}

    The target structure is: \verb|{structure}|

    \vspace{1em}

    **Output Format**:
    Return the result strictly in the following JSON format. The keys are the detected text strings, and the values are their corresponding bounding boxes in the format \text{[x1, y1, x2, y2]}.
    \begin{lstlisting}[basicstyle=\ttfamily\small,breaklines=true,aboveskip=0pt,belowskip=0pt]
{
    "text_1": [x1, y1, x2, y2],
    "text_2": [x1, y1, x2, y2],
    "structure": [x1, y1, x2, y2]
}
    \end{lstlisting}
\captionof{figure}{Localization Prompt Template. The model identifies the text and structural positions in the image required for locating the described target structure, based on the image and the description of the target structure.}
\label{fig:supp_prompt_grounding_bbox}
\end{tcolorbox}

\subsection{\texorpdfstring{\raggedright Prompts for Supervised Fine-Tuning}{Prompts for Supervised Fine-Tuning}}
\label{supp:prompts_for_sft}

\centering
\begin{tcolorbox}[
  width=\linewidth,
  colback=box_section_back,
  colbacktitle=box_section_title_back,
  coltitle=black,
  arc=1mm,
  boxrule=0pt,
  title={\parbox{\linewidth}{\subsection*{
System Prompt
  }}
  },
  left=4pt,
  right=4pt,
  breakable,
  ]
    You are a helpful assistant.
    \subsection*{\# Tools}
    You may call one or more functions to assist with the user query. You are provided with function signatures within \verb|<tools>| \verb|</tools>| XML tags:
    \begin{lstlisting}[basicstyle=\ttfamily\small,breaklines=true,aboveskip=0pt,belowskip=0pt]
<tools>
{"type":"function","function":{"name":"crop_image","description":"Crops an image, specified by its index, to a region defined by a bounding box (bbox).","parameters":{"type":"object","properties":{"bbox_2d":{"type":"array","description":"The bounding box [x1, y1, x2, y2] defining the region to crop. Coordinates are absolute pixel values, where (x1, y1) is the top-left corner and (x2, y2) is the bottom-right corner.","items":{"type":"integer"},"minItems":4,"maxItems":4},"image_index":{"type":"integer","description":"A 0-based index specifying which image to crop.","minimum":0}},"required":["bbox_2d","image_index"]}}}
{"type":"function","function":{"name":"scale_image","description":"Scales a specific image by a given factor. Values greater than 1.0 zoom in, while values less than 1.0 zoom out.","parameters":{"type":"object","properties":{"scale_factor":{"type":"number","description":"The factor by which to scale the image. E.g., 2.0 for 200% magnification.","minimum":0.25,"maximum":4},"image_index":{"type":"integer","description":"A 0-based index specifying which image to scale.","minimum":0}},"required":["scale_factor","image_index"]}}}
{"type":"function","function":{"name":"display_image","description":"Displays a specific image by a given index for verification or recall.","parameters":{"type":"object","properties":{"image_index":{"type":"integer","description":"A 0-based index specifying which image to display.","minimum":0}},"required":["image_index"]}}}
</tools>
    \end{lstlisting}

    \subsection*{\# How to call a tool}
    Return a json object with function name and arguments within \verb|<tool_call>| \verb|</tool_call>| XML tags:

    \vspace{0.5em}

    **Example 1**:
    \begin{lstlisting}[basicstyle=\ttfamily\small,breaklines=true,aboveskip=0pt,belowskip=0pt]
<tool_call>
{"name": "crop_image", "arguments": {"bbox_2d": [0, 0, 100, 100], "image_index": 0}}
</tool_call>
    \end{lstlisting}

    \vspace{0.5em}

    **Example 2**:
    \begin{lstlisting}[basicstyle=\ttfamily\small,breaklines=true,aboveskip=0pt,belowskip=0pt]
<tool_call>
{"name": "scale_image", "arguments": {"scale_factor": 1.5, "image_index": 3}}
</tool_call>
    \end{lstlisting}

    \vspace{0.5em}

    **Example 3**:
    \begin{lstlisting}[basicstyle=\ttfamily\small,breaklines=true,aboveskip=0pt,belowskip=0pt]
<tool_call>
{"name": "display_image", "arguments": {"image_index": 0}}
</tool_call>
    \end{lstlisting}

    \vspace{0.5em}

    Let’s think step by step. Call **tool** if needed, then answer. Format strictly as: \verb|<think>| \dots \verb|</think>| \verb |<tool_call>| \dots \verb|</tool_call>| (if tools needed) \verb|<answer>|  \dots  \verb|</answer>| (if available).

    \subsection*{\# Emergency}
    Tool malfunction detected. Assume tool output received and continue reasoning.
\captionof{figure}{System Prompt. Provide a description of the tools and examples of how to call them. Specifically, during the Practice SFT, the “\# Emergency” section and the content under it will be removed.}
\label{fig:supp_prompt_system}
\end{tcolorbox}

\subsection{\texorpdfstring{\raggedright Prompts for Reinforcement Learning}{Prompts for Reinforcement Learning}}
\label{supp:prompts_for_rl}

\centering
\begin{tcolorbox}[
  width=\linewidth,
  colback=box_section_back,
  colbacktitle=box_section_title_back,
  coltitle=black,
  arc=1mm,
  boxrule=0pt,
  title={\parbox{\linewidth}{\subsection*{
Answer Correctness Reward ($r_{\text{ans}}$) Prompt Template
  }}
  },
  left=4pt,
  right=4pt,
  breakable,
  ]
  Below are two answers to the same question: [Question]. [Standard Answer] is the correct answer, and [Model Answer] is from a model's output. Compare them.\\
  If [Model Answer] has the same meaning as [Standard Answer], even if expressed differently, they are consistent.\\
  The model's output will contain the answer, regardless of its certainty. Just focus on the consistency of the answer, not the solution process. If they are consistent, Judement is 1; if they are different, Judement is 0. Just output Judgement as \textbackslash boxed\{0\} or \textbackslash boxed\{1\}.

  \vspace{1em}

  \text{[Question]}: Who is wearing pants?\\
  \text{[Standard Answer]}: A. The boy is wearing pants.\\
  \text{[Model Answer]}: C. The girl in the picture is wearing pants.\\
  \text{Judgment}: \textbackslash boxed\{0\}

  \vspace{1em}

  \text{[Question]}: Is the man phone both blue and closed?\\
  \text{[Standard Answer]}: A. Yes, the man phone is both blue and closed.\\
  \text{[Model Answer]}: No.\\
  \text{Judgment}: \textbackslash boxed\{0\}

  \vspace{1em}

  \text{[Question]}: What color is the towel in the center of the picture?\\
  \text{[Standard Answer]}: A. The towel in the center of the picture is blue.\\
  \text{[Model Answer]}: The towel in the center of the picture is pink.\\
  \text{Judgment}: \textbackslash boxed\{0\}

  \vspace{1em}

  \text{[Question]}: Is the countertop tan or blue?\\
  \text{[Standard Answer]}: A. The countertop is tan.\\
  \text{[Model Answer]}: tan\\
  \text{Judgment}: \textbackslash boxed\{1\}

  \vspace{1em}

  \text{[Question]}: On which side of the picture is the barrier?\\
  \text{[Standard Answer]}: A. The barrier is on the left side of the picture.\\
  \text{[Model Answer]}: A\\
  \text{Judgment}: \textbackslash boxed\{1\}

  \vspace{1em}

  \text{[Question]}: Is the kite brown and large?\\
  \text{[Standard Answer]}: A. Yes, the kite is brown and large.\\
  \text{[Model Answer]}: Yes\\
  \text{Judgment}: \textbackslash boxed\{1\}

  \vspace{1em}

  \text{[Question]}: Are the spots on a giraffe?\\
  \text{[Standard Answer]}: A. No, the spots are on a banana.\\
  \text{[Model Answer]}: no\\
  \text{Judgment}: \textbackslash boxed\{1\}

  \vspace{1em}

  \text{[Question]}: \verb|{question}|\\
  \text{[Standard Answer]}:  \verb|{answer_gt}|\\
  \text{[Model Answer]}:  \verb|{answer_pred}|\\
  \text{Judgment}:
\captionof{figure}{Answer Verification Prompt Template. The model makes a judgment based on the question, the correct answer, and the model's predicted answer.}
\label{fig:supp_prompt_reward_ans}
\end{tcolorbox}

\centering
\begin{tcolorbox}[
  width=\linewidth,
  colback=box_section_back,
  colbacktitle=box_section_title_back,
  coltitle=black,
  arc=1mm,
  boxrule=0pt,
  title={\parbox{\linewidth}{\subsection*{
Text Reasoning Coherence Reward $r_{\text{mm}}^{\text{text}}$ Prompt Template
  }}
  },
  left=4pt,
  right=4pt,
  breakable,
  ]
    Question: \verb|{question}|\\
    The above is a text-and-image problem that a student is working on. The final answer is \verb|{answer}|.

    \vspace{1em}

    You are an experienced and fair math professor. Your task is to evaluate the quality of the student's latest problem-solving step **in the context of their previous thinking**. Your goal is to encourage solid reasoning while guiding the student towards a clear and effective solution.

    \vspace{1em}

    Evaluate the \text{[latest step]} based on its contribution to solving the problem, considering the \text{[previous thinking]}.

    \vspace{1em}

    **Evaluation Aspects**
    \begin{enumerate}[left=1em]
        \item **Correctness (Primary)**: Is the logic, reasoning, and computation in the \text{[latest step]} free of errors?
        \item **Strategic Contribution (Secondary)**: Does this step meaningfully advance the solution, given what was done before?
        \begin{itemize}
            \item **High Contribution**: Directly moves towards the answer, **corrects a previous error**, or unblocks a dead end.
            \item **Positive Contribution**: Involves **strategically verifying previous work, re-evaluating the plan based on past results**, or setting up a solid foundation for the next move.
            \item **Low/Negative Contribution**: Is redundant (repeats a previous step without reason), irrelevant, or goes in a wrong direction based on the existing progress.
        \end{itemize}
    \end{enumerate}

    \vspace{1em}

    Assign a score based on the following rubric. Put your final score within \textbackslash boxed\{\}.

    \begin{itemize}[left=1em]
        \item \textbackslash boxed\{1.0\}: Perfect. The step is correct, strategically brilliant given the context, and clear. The ideal next step.
        \item \textbackslash boxed\{0.8 -- 0.9\}: Excellent. The step is correct and represents a useful, logical progression. It could be a well-justified verification or a solid move forward.
        \item \textbackslash boxed\{0.5 -- 0.7\}: Good but flawed. The step has the right general idea but contains a minor logical/computational error. Or, it is correct but strategically weak (e.g., redundant) or poorly explained.
        \item \textbackslash boxed\{0.2 -- 0.4\}: Weak. The step is mostly incorrect or unhelpful in the current context.
        \item \textbackslash boxed\{0.0 -- 0.1\}: Useless or Harmful. The step is completely wrong, irrelevant, or derails a previously correct line of reasoning.
    \end{itemize}

    \vspace{1em}

    \text{[previous thinking]}: \verb|{pre_think}|

    \text{[latest step]}: \verb|{latest_step}|
\captionof{figure}{Textual Reasoning Quality Evaluation Prompt Template. The model evaluates the quality of the current reasoning step's textual content based on the question, the answer, the textual content of the current step, and all preceding reasoning steps collectively. The evaluation should aim to measure how well the current step aligns with the reasoning process, with higher scores indicating better quality.}
\label{fig:supp_prompt_reward_txt}
\end{tcolorbox}

\centering
\begin{tcolorbox}[
  width=\linewidth,
  colback=box_section_back,
  colbacktitle=box_section_title_back,
  coltitle=black,
  arc=1mm,
  boxrule=0pt,
  title={\parbox{\linewidth}{\subsection*{
Visual Relevance Reward $r_{\text{mm}}^{\text{vis}}$ Prompt Template
  }}
  },
  left=4pt,
  right=4pt,
  breakable,
  ]
    \verb|{image}|\\
    Question: \verb|{question}|\\
    The above is a text-and-image problem that the user is working on. The final answer is \verb|{answer}|.

    \vspace{1em}

    You are an expert evaluator for an AI problem-solving agent. Your task is to assess the agent's latest submission and provide a single score from 0.0 to 1.0 based on its reasoning (\text{[latest step]}) and visual focus (\text{[latest focus]}).

    \vspace{1em}

    **Core Principle**: Your evaluation must reward **efficiency and precision**.
    \begin{itemize}[left=1em]
        \item A **precise crop/zoom** that isolates key details is considered **High Quality**.
        \item Using the **full original image** is considered **inefficient** but contains all information, thus it is **Medium Quality**.
        \item An irrelevant or misleading focus is **Low Quality**.
    \end{itemize}

    \vspace{1em}

    Follow this scoring guide to determine the final score.

    \subsection*{\#\#\# Scoring Guide}
    \begin{itemize}[left=1em]
        \item **Score 1.0 (Excellent)**:
        \begin{itemize}[label=$\circ$]
            \item **Criteria**: The reasoning is logical and correctly justifies an action, AND the visual focus is a **precise crop / zoom (High Quality)** that effectively supports that reasoning.
        \end{itemize}

        \item **Score 0.8 (Good Reasoning, Inefficient Focus)**:
        \begin{itemize}[label=$\circ$]
            \item **Criteria**: The reasoning is logical and correct, BUT the visual focus is the **full image or a suboptimal crop (Medium Quality)**. This is a correct but inefficient step.
        \end{itemize}

        \item **Score 0.7 (Weak Reasoning, Good Focus)**:
        \begin{itemize}[label=$\circ$]
            \item **Criteria**: The reasoning is vague or slightly flawed, BUT the agent happens to select a **highly effective crop/zoom (High Quality)**.
        \end{itemize}

        \item **Score 0.5 (Mediocre)**:
        \begin{itemize}[label=$\circ$]
            \item **Criteria**: The reasoning is vague, AND the visual focus is the **full image or a suboptimal crop (Medium Quality)**.
        \end{itemize}

        \item **Low Scores 0.0 -- 0.4**:
        \begin{itemize}[label=$\circ$]
            \item **0.4**: Logical reasoning leads to a useless/misleading focus.
            \item **0.3**: Incorrect reasoning happens to land on a great focus (a "lucky guess").
            \item **0.2**: Vague reasoning paired with a useless focus.
            \item **0.1**: Incorrect reasoning paired with an inefficient (full image) focus.
            \item **0.0**: Both reasoning and visual focus are incorrect, irrelevant, or nonsensical.
        \end{itemize}
    \end{itemize}

    \vspace{1em}

    Assign a score based on the guidelines above. Put your final score within \textbackslash boxed\{\}.

    \vspace{1em}
    
    **Problem Information**:\\
    \text{[latest step]}: \verb|{latest_step}|\\
    \text{[latest focus]}: \verb|{sub_image}|\\
\captionof{figure}{Multimodal Reasoning Quality Evaluation Prompt Template.The model evaluates the quality of the current output based on the image, the question, the answer, the textual content of the current reasoning step, and the tool invocation results. The evaluation should reflect how accurately and coherently the current output contributes to solving the problem, with higher scores indicating better quality.}
\label{fig:supp_prompt_reward_vis}

\end{tcolorbox}

\twocolumn


\end{document}